\documentclass[10pt,twocolumn,letterpaper]{article}
\PassOptionsToPackage{dvipsnames,table}{xcolor}

\usepackage{cvpr}
\usepackage[dvipsnames]{xcolor}

\definecolor{cvprblue}{rgb}{0.21,0.49,0.74}

\usepackage{times}
\usepackage{graphicx}
\usepackage{amsmath}
\usepackage{amssymb}
\usepackage{booktabs}
\usepackage{rotating}
\usepackage{paralist}
\usepackage{caption}
\usepackage{subcaption}
\usepackage[linesnumbered,boxed]{algorithm2e}
\usepackage{arydshln}
\usepackage{array}
\usepackage{multirow}
\usepackage{ragged2e}
\usepackage{booktabs}
\usepackage{arydshln}
\usepackage{makecell}

\usepackage[dvipsnames]{xcolor}
\usepackage[accsupp]{axessibility}
\usepackage{times}
\usepackage{array}
\usepackage{ragged2e}
\usepackage{booktabs}
\usepackage{wrapfig}

\usepackage{graphicx}
\usepackage{amsmath}
\usepackage{amssymb}
\usepackage[pagebackref=true,breaklinks=true,colorlinks,bookmarks=false,citecolor=blue,linkcolor=blue]{hyperref}
\usepackage{makecell}
\usepackage[accsupp]{axessibility}
\usepackage{color, colortbl}
\definecolor{LightCyan}{rgb}{0.88,1,1}

\hypersetup{
    colorlinks=true,
    urlcolor=magenta,
    citecolor=blue,
}

\usepackage[T1]{fontenc}
\usepackage{multirow}
\usepackage{booktabs}
\usepackage{comment}
\usepackage{color}

\usepackage{textcomp}
\usepackage{siunitx}
\usepackage{tabulary}
\usepackage{makecell}
\usepackage{multirow}
\usepackage{booktabs}
\usepackage{hyperref}

\makeatletter
\newcommand\figref{Figure~\ref}
\newcommand{\tabref}[1]{Table~\ref{#1}}
\newcolumntype{P}[1]{>{\centering\arraybackslash}p{#1}}
\newcolumntype{M}[1]{>{\centering\arraybackslash}m{#1}}
\let\ts@includegraphics\includegraphics

\usepackage{float}
\usepackage{lipsum}   
\usepackage{rotating} 
\usepackage{stfloats} 
\usepackage[export]{adjustbox} 

\makeatletter
\g@addto@macro\@maketitle{
  \begin{minipage}{\textwidth}
  \centering
  \setlength{\tabcolsep}{1pt} 

    \begin{tabular}{ccccc}
      & Blur & Haze & Low-light & Rain \\
      \multirow{2}[2]{*}[10pt]{\rotatebox[origin=c]{90}{\textbf{Input}}} & 
      \includegraphics[valign=m,width=.23\linewidth, height=1.44cm]{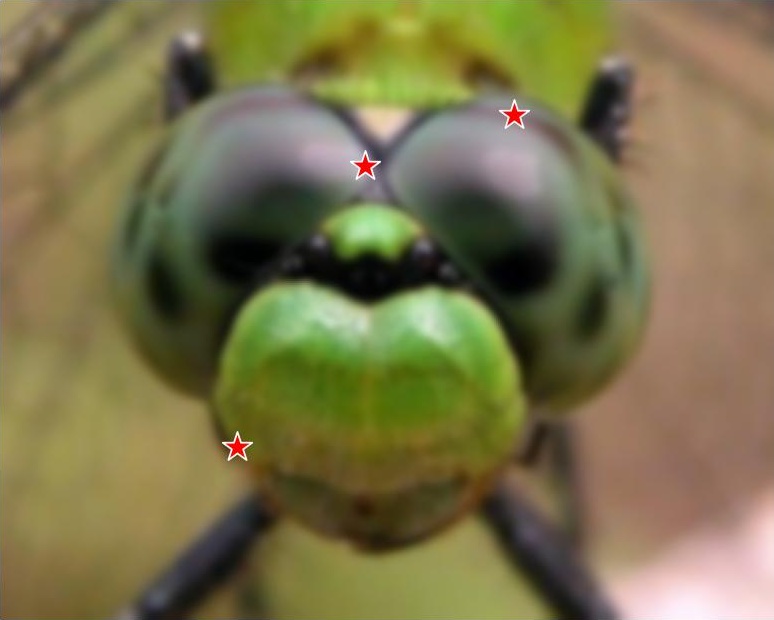} &
      \includegraphics[valign=m,width=.23\linewidth, height=1.44cm]{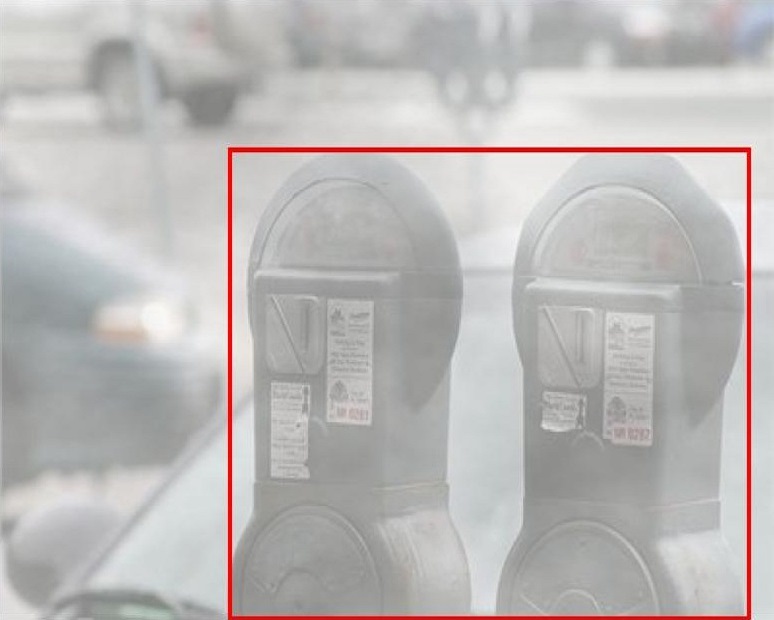} &
      \includegraphics[valign=m,width=.23\linewidth, height=1.44cm]{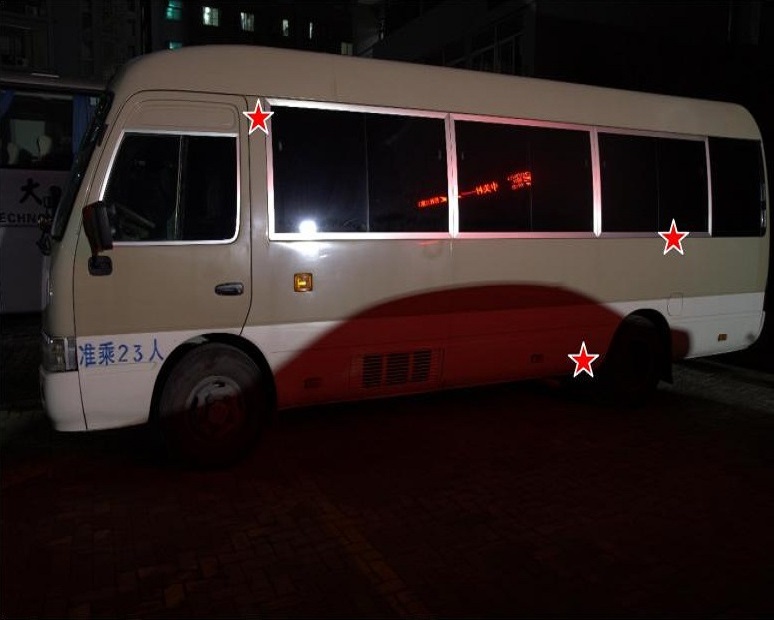} &
      \includegraphics[valign=m,width=.23\linewidth, height=1.44cm]{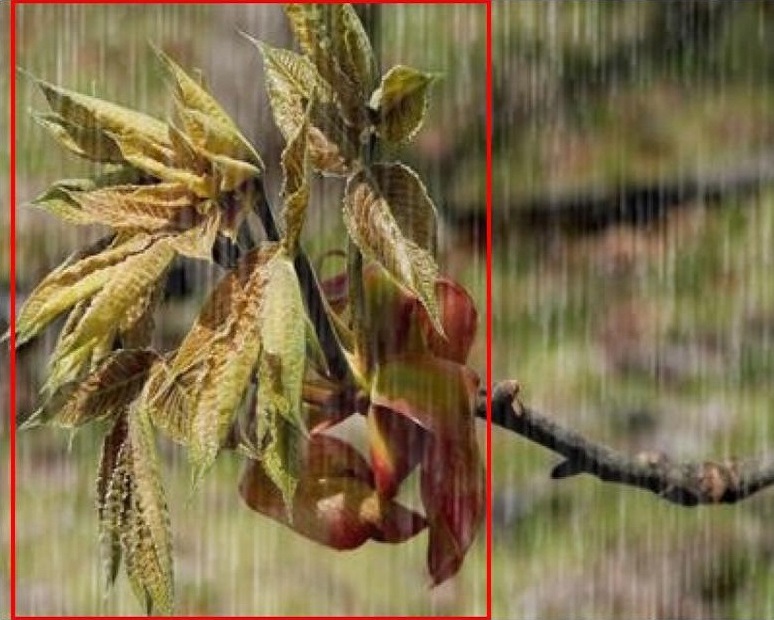} \\     
    \multicolumn{5}{c}{\vspace{-10pt}} \\
      \multirow{2}[2]{*}[10pt]{\rotatebox[origin=c]{90}{\textbf{SAM}}} & 
      \includegraphics[valign=m,width=.23\linewidth, height=1.44cm]{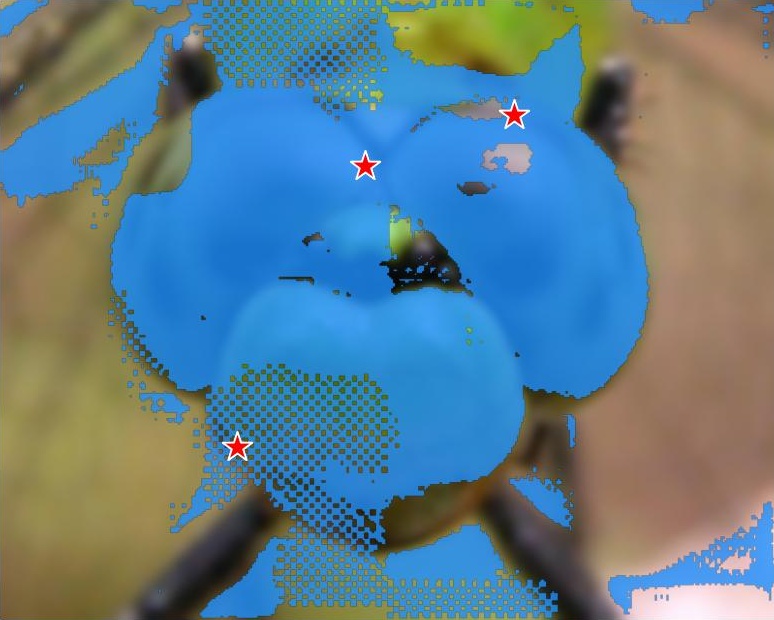} &
      \includegraphics[valign=m,width=.23\linewidth, height=1.44cm]{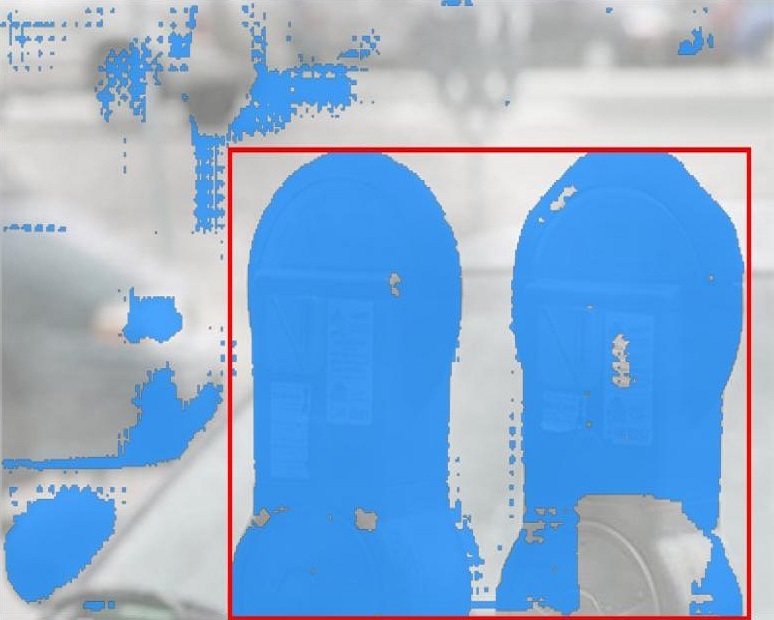} &
      \includegraphics[valign=m,width=.23\linewidth, height=1.44cm]{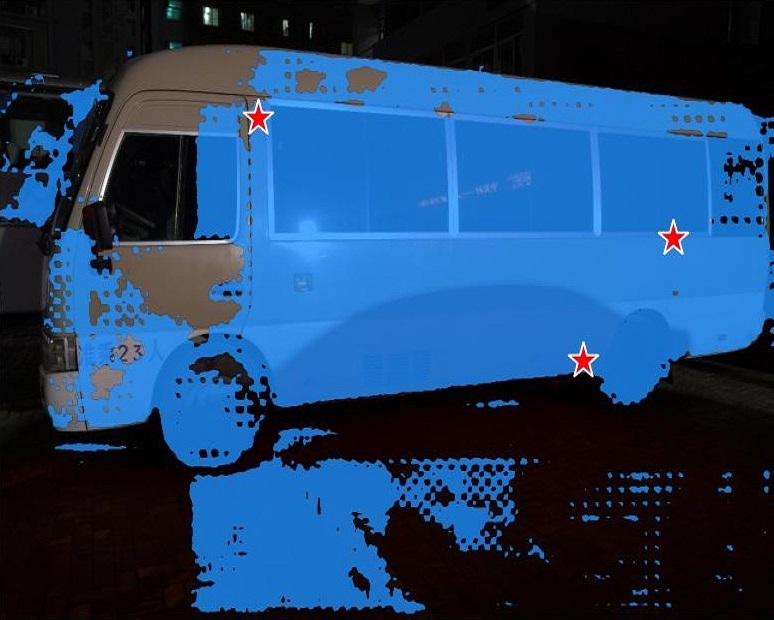} &
      \includegraphics[valign=m,width=.23\linewidth, height=1.44cm]{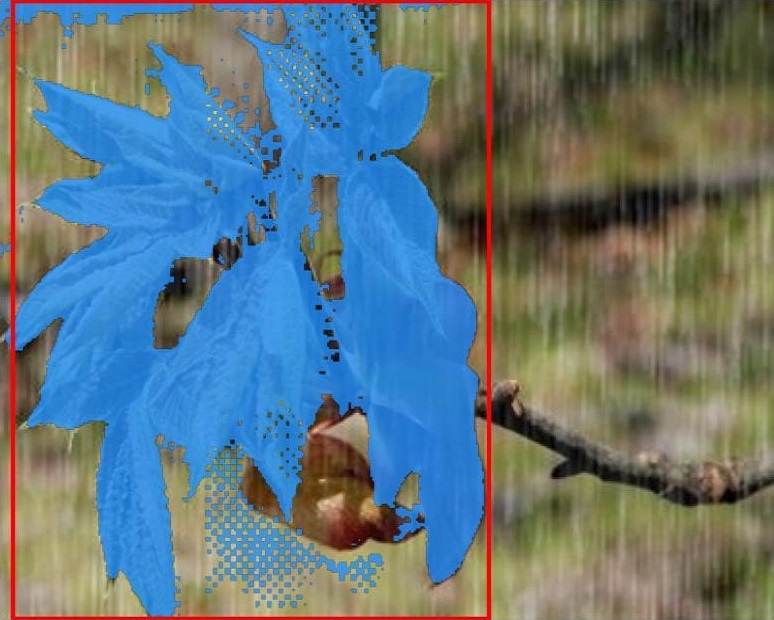} \\
      \multicolumn{5}{c}{\vspace{-10pt}} \\
      \multirow{2}[2]{*}[20pt]{\rotatebox[origin=c]{90}{\textbf{RobustSAM}}} & 
      \includegraphics[valign=m,width=.23\linewidth, height=1.44cm]{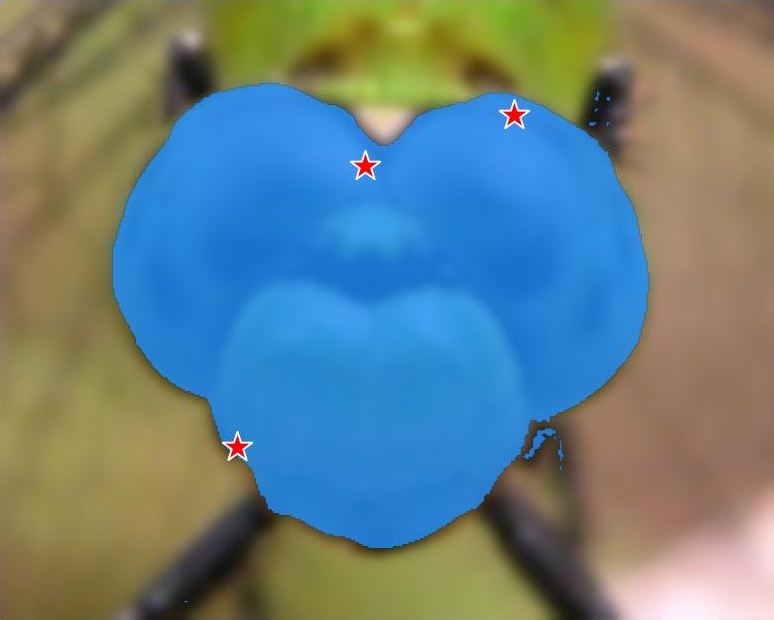} &
      \includegraphics[valign=m,width=.23\linewidth, height=1.44cm]{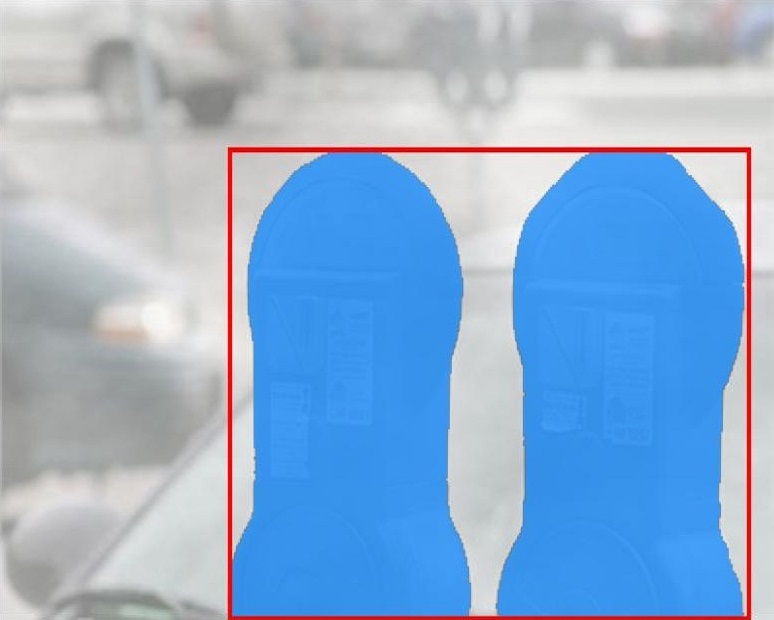} &
      \includegraphics[valign=m,width=.23\linewidth, height=1.44cm]{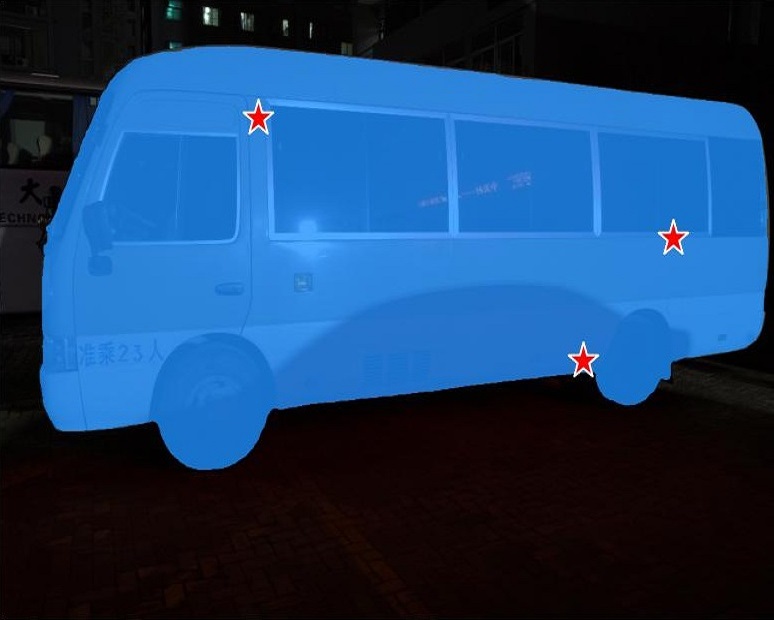} &
      \includegraphics[valign=m,width=.23\linewidth, height=1.44cm]{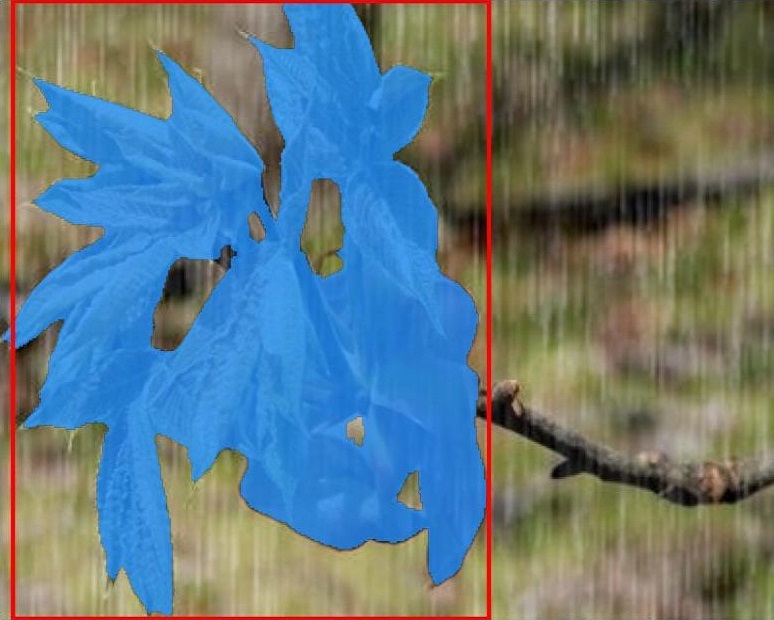} \\
    \end{tabular}
    \label{fig:teasor}
  \captionof{figure}{\textbf{Comparative results of SAM and RobustSAM under different degradations using unseen datasets:} RobustSAM outperforms with precise boundaries and intact structures, where SAM falters with errors and fragmentation. Red star points and bounding boxes are our examples' input prompts.}
  \label{fig:multiple_images}
  \end{minipage}
  \vspace{0.5cm} 
}
\makeatother

\usepackage[capitalize]{cleveref}
\crefname{section}{Sec.}{Secs.}
\Crefname{section}{Section}{Sections}
\Crefname{table}{Table}{Tables}
\crefname{table}{Tab.}{Tabs.}

\def\paperID{13741} 
\def\confName{CVPR}
\def\confYear{2024}

\begin{document}

\title{RobustSAM: Segment Anything Robustly on Degraded Images}

\author{
Wei-Ting Chen\textsuperscript{1,2†}
\quad Yu-Jiet Vong\textsuperscript{1}
\quad Sy-Yen Kuo\textsuperscript{1} 
\quad Sizhuo Ma\textsuperscript{2*}
\quad Jian Wang\textsuperscript{2*}
\\\\
\hspace{-8mm}\textsuperscript{1}National Taiwan University\quad \textsuperscript{2}Snap Inc.
}

\maketitle
\newcommand\blfootnote[1]{%
\begingroup
\renewcommand\thefootnote{}\footnote{#1}%
\addtocounter{footnote}{-1}%
\endgroup
}
\blfootnote{Project Page: \href{https://robustsam.github.io}{https://robustsam.github.io}}
\blfootnote{† Part of the work done during internship at Snap Research.}
\blfootnote{* Co-corresponding authors}

\begin{abstract}
Segment Anything Model (SAM) has emerged as a transformative approach in image segmentation, acclaimed for its robust zero-shot segmentation capabilities and flexible prompting system. Nonetheless, its performance is challenged by images with degraded quality. Addressing this limitation, we propose the Robust Segment Anything Model (RobustSAM), which enhances SAM's performance on low-quality images while preserving its promptability and zero-shot generalization. Our method leverages the pre-trained SAM model with only marginal parameter increments and computational requirements. The additional parameters of RobustSAM can be optimized within 30 hours on eight GPUs, demonstrating its feasibility and practicality for typical research laboratories. We also introduce the Robust-Seg dataset, a collection of 688K image-mask pairs with different degradations designed to train and evaluate our model optimally. Extensive experiments across various segmentation tasks and datasets confirm RobustSAM's superior performance, especially under zero-shot conditions, underscoring its potential for extensive real-world application. Additionally, our method has been shown to effectively improve the performance of SAM-based downstream tasks such as single image dehazing and deblurring.
\end{abstract}

\section{Introduction}
Accurate image segmentation is crucial to various downstream applications such as robotics, augmented/virtual reality, and content creation. 
Segment Anything Model (SAM)~\cite{kirillov2023segany} has opened a new chapter in image segmentation in the wild:
Utilizing the comprehensive SA-1B dataset, which comprises over a billion annotated masks, SAM generalizes to a huge variety of objects and can accurately segment a scene given minimalistic prompts, ranging from point annotations to bounding boxes. This innovative approach revolutionizes zero-shot segmentation by seamlessly adapting to various applications.

As SAM demonstrates its versatility across diverse segmentation tasks~\cite{ji2023segment,tang2023can,hu2023skinsam,zhang2023customized,wu2023medical,tang2023can,ma2023can}, attention has turned to its robustness and scalability when confronting complex and challenging scenarios. Specifically, enhancing its robustness on degraded images remains a frontier to be explored. A body of literature~\cite{qiao2023robustness,shan2023robustness,huang2023robustness,wang2023empirical} has pointed out that the performance of SAM decreases with imaging degradations like low lighting, noise, blur, adverse weather, and compression artifacts. These degradations significantly impact the quality of the segmentation masks that SAM generates, directly influencing downstream tasks relying on these masks. Specifically, recent image restoration works such as dehazing~\cite{jin2023let}, deblurring~\cite{li2023sam}, and super-resolution~\cite{lu2023can} have been utilizing these masks or latent features as a structure prior which generalizes to unseen scenes. However, these works assume that SAM can produce reliable and accurate masks even in degraded conditions. If the robustness of SAM is compromised in these cases, the benefit of integrating prior knowledge becomes constrained, limiting their applicability to real-world scenarios.
To address this challenge, one intuitive approach is to utilize existing image restoration techniques~\cite{patil2023multi,kulkarni2022unified,chen2022learning} to preprocess the images before feeding them into SAM.
Although these methods can improve the image quality to a degree, there is
no guarantee that the selected image restoration techniques would be able to improve image segmentation~\cite{son2020urie,
chen2022rvsl,vidal2018ug,diamond2021dirty,li2019single}. This is because most image restoration algorithms are optimized for human visual perception rather than the specific demands of segmentation models like SAM.

An alternative strategy involves directly fine-tuning SAM on degraded images. However, 
direct adjustments to the SAM decoder or integrating a new decoder module can profoundly impair the model's generalizability on zero-shot tasks. 
Furthermore, blindly fine-tuning SAM with degraded images can lead to catastrophic forgetting, where the network inadvertently loses its knowledge learned from the original, clean images~\cite{guo2019degraded,kirkpatrick2017overcoming}.
To this end, we introduce RobustSAM, which achieves robustness in handling degraded images while retaining zero-shot functionality.
Our method proposes two novel modules: the Anti-Degradation Token Generation Module and the Anti-Degradation Mask Feature Generation Module. 
Supervised by consistency losses with features extracted from paired clear images by original SAM, these modules are designed to extract degradation-invariant segmentation features.
We also fine-tuned the original output token of SAM, adapting it to our robust segmentation approach. 
By freezing the original modules from SAM during training, the proposed method enhanced its ability to process degraded images while preserving its effectiveness on zero-shot segmentation.

Moreover, the proposed additional modules in RobustSAM can be trained efficiently.
In contrast to the original SAM, which demands training on hundreds of GPUs, RobustSAM can be trained within 30 hours on eight A100s. This marks the accessibility of RobustSAM, making it ready to be integrated into various application scenarios. Extensive experiments demonstrate that our RobustSAM performs well in clear and degraded scenarios. Furthermore, we found that RobustSAM benefits SAM-based downstream tasks in degraded scenarios, such as single image dehazing and deblurring, by providing a more robust prior, thereby enhancing their effectiveness.

To enhance the capabilities and robustness of RobustSAM, we introduce the Robust-Seg dataset. Robust-Seg combines 43K meticulously annotated images from 7 existing datasets. Each image is subject to 15 different types of carefully modeled synthetic degradation, resulting in a comprehensive collection of 688K images in Robust-Seg. This extensive dataset aims to push the boundaries of image segmentation and serve as a valuable resource for future research.

To summarize, our contributions are as follows:
\begin{compactitem}
\item We propose RobustSAM, a zero-shot segmentation model built upon the Segment Anything model, with enhanced robustness against various image degradations. This enhanced robustness is shown to improve the performance of downstream applications significantly.
\item We construct the Robust-Seg dataset, a collection of 688K image-mask pairs with different degradations. We hope Robust-Seg will establish a new benchmark for segmentation models on degraded images.
\end{compactitem}

\section{Related Work}
\subsection{Segment Anything Model}
Segment Anything Model (SAM) \cite{kirillov2023segany} achieves unprecedented performance in image segmentation, advancing various subdomains in computer vision \cite{ma2023segment}.
SAM accepts intuitive prompts such as points or bounding boxes, demonstrating exceptional zero-shot transfer learning capability across diverse segmentation tasks and new image distributions. Its adaptability is proven across various domains, including medical imaging \cite{zhang2023input}, \cite{tang2023can}, \cite{han2023segment}, \cite{shen2023anything}, camouflaged object detection \cite{tang2023can}, and salient object segmentation \cite{ma2023segment}. In addition to its segmentation capability, SAM plays a foundational role in enhancing computer vision tasks, including semantic segmentation \cite{zhang2023input}, \cite{shen2023anything}, \cite{chen2023semantic}, image editing \cite{yu2023inpaint}, and video object tracking \cite{yang2023track}, \cite{cheng2023segment}. While SAM demonstrates promising capabilities, its performance is challenged by poor image quality~\cite{qiao2023robustness,shan2023robustness,huang2023robustness,wang2023empirical}, affecting segmentation and downstream task accuracy. 

\subsection{Robust Segmentation}
In the domains of autonomous driving and surveillance analysis, a multitude of studies~\cite{guo2019degraded, endo2023semantic, xia2019cooperative, rajagopalan2023improving, zhou2020multi, yang2022self} have identified a decrease in CNN-based segmentation performance when dealing with degraded images, which leads to the development of various remedial approaches. For instance, QualNet~\cite{kim2021quality} explores quality-agnostic feature extraction through a reversible encoding scheme, while URIE~\cite{son2020urie} addresses multiple image impairments, enhancing segmentation stability through classification constraints. Concurrently, FIFO~\cite{lee2022fifo} propels segmentation frameworks to learn fog-resistant features via a fog pass filter mechanism. However, these technologies primarily focus on a single type of degradation, potentially lacking robustness against multiple image degradations. Moreover, this strategy of jointly training with downstream tasks may dilute the zero-shot advantage of SAM.

\subsection{Image Restoration}
In the field of image restoration, methods targeting single types of degradation such as SRCNN~\cite{dong2014learning} pioneered the application of convolutional neural networks to the enhancement of image quality. Subsequent innovations have emerged across different domains, achieving notable successes in super-resolution (SR)~\cite{ledig2017photo,lim2017enhanced,zhang2018residual}, denoising~\cite{abdelhamed2019noise,brooks2019unprocessing,mildenhall2018burst,zhang2018ffdnet}, dehazing~\cite{chen2019pms,li2020zero,he2010single}, deraining~\cite{qian2018attentive,ren2019progressive,yang2017deep}, underwater enhancement~\cite{liu2020real,yang2021laffnet} and deblurring~\cite{gao2019dynamic,kupyn2018deblurgan,kupyn2019deblurgan}. Attempts like MPRNet~\cite{zamir2021multi}, and HINet~\cite{chen2021hinet} have been made to address multiple degradations through a single network. Recently, transformer-based methods have also gained traction in image restoration tasks~\cite{liang2021swinir,wang2022uformer,zamir2022restormer}. Nonetheless, while multi-degradation approaches such as All-in-One~\cite{li2020all}, IPT~\cite{chen2021pre}, and AirNet~\cite{li2022all} offer greater flexibility and improved performance, 
they aim to enhance visual quality for human perception, instead of improving the performance of downstream tasks such as segmentation.

\begin{figure*}[t!]
\centering \includegraphics[width=1.0\textwidth,page=1]{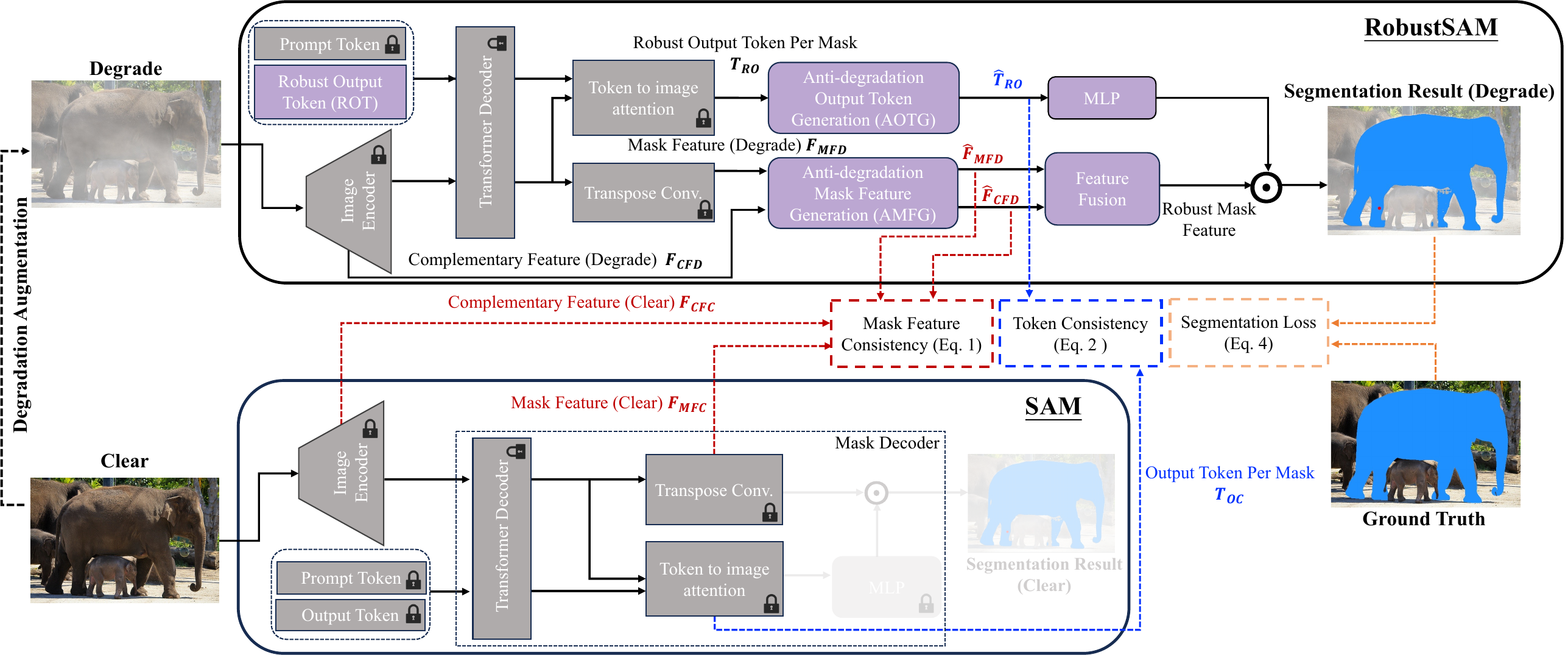}{}
\makeatother 
\caption{\textbf{Overview of our proposed RobustSAM.} RobustSAM augments the original SAM by incorporating five essential components (\textcolor{Orchid}{in purple}). \textit{During training}, clear images are fed through the original SAM modules (\textcolor{Gray}{in gray}) to produce features for clear scenes. Subsequently, degraded images, generated through augmentation of clear inputs, are processed by RobustSAM, yielding features for degraded scenarios. These are then refined via Anti-degradation modules, ensuring consistency with features from clear scenes. This methodology, supported by a segmentation loss, achieves precise segmentation outcomes in both clear and degraded image conditions. \textit{During inference}, only RobustSAM is used to predict a segmentation mask from an input image. Note: The prompt encoder is excluded for conciseness, and the padlock icons represent fixed components loaded from the original SAM model that are not updated during training.}
\label{fig:overview}
\end{figure*}

\section{Proposed Method}
\subsection{Preliminary: Segment Anything Model}
We provide a concise overview of the SAM framework~\cite{kirillov2023segany}. As shown in the lower half of~\figref{fig:overview}, SAM incorporates three key components: an image encoder, a prompt encoder, and a mask decoder. The image encoder processes input images using the Vision Transformer (ViT). The prompt encoder handles sparse prompt inputs (such as points, boxes, and text) and dense inputs (masks), transforming them into appropriate representations. The mask decoder is a modified Transformer decoder block~\cite{vaswani2017attention}. It combines image and prompt embeddings with an output token to generate mask features. This process involves prompt self-attention and bidirectional cross-attention between the prompts and image embeddings. Notably, the mask decoder uses transpose convolutions to create detailed mask features. The output token per mask, derived from token-to-image attention, is transformed by an MLP into a dynamic classifier. When multiplied with the mask features, this classifier yields the final segmentation mask.

\subsection{Robust Segment Anything Model}
We propose RobustSAM, which addresses image degradation while preserving the zero-shot learning capabilities of SAM. Diverging from standard approaches that fine-tune SAM or jointly train complex adaptation modules, RobustSAM employs a minimalistic yet deliberate enhancement. 

\subsubsection{Model Overview}
\figref{fig:overview} gives an overview of the proposed RobustSAM. The key contribution of RobustSAM is the Anti-Degradation Output Token Generation (AOTG) and Anti-Degradation Mask Feature Generation (AMFG) modules, which extract degradation-invariant information that is aligned those extracted from clear images by the original SAM. This is achieved by generating clear-degraded image pairs through 15 types of synthetic degradation augmentation. Different losses are then applied to enforce the consistency between clear and degraded features, as well as the predicted segmentation and ground truth. Notice that although supervised with synthetic degradations, RobustSAM generalizes well to real-world images, as shown in \cref{sec:exp}.
\noindent \smallskip\\
\textbf{Training.} 
To train RobustSAM, we begin by applying a degradation augmentation to a clear input image and then feeding the resulting degraded image into RobustSAM. 
Initially, the model leverages its Image Encoder to extract features from this degraded image.
Unlike the original SAM framework, we finetuned the output token, now called the Robust Output Token (ROT). 
This ROT, along with the prompt token and the features extracted by the Image Encoder, is processed through the original SAM layers to generate mask feature (degrade) $F_{MFD}$ and Robust Output Token per mask $T_{RO}$. 

The AOTG block processes the $T_{RO}$ to extract information resilient to degradation, transforming it into $\hat{T}_{RO}$. Simultaneously, the AMFG block refines the mask and complementary features from the Image Encoder's early and final layers  ($F_{MFD}$ and $F_{CFD}$), removing degradation-related information to produce refined features ($\hat{F}_{MFD}$ and $\hat{F}_{CFD}$). Following the architecture proposed in~\cite{sam_hq}, a Feature Fusion block combines these refined features into our final robust mask feature for improving the segmentation quality. 

In parallel, the original clear image is processed by standard SAM to extract clear versions of the complementary feature ($F_{CFC}$), mask feature ($F_{MFC}$), and output token ($T_{OC}$). Consistency losses between these clear features and refined features from RobustSAM ensure alignment with undegraded image outputs. The segmentation results from the degraded input are then compared against the ground truth using a segmentation loss function.

In our degradation augmentation approach, we include 15 types of degradations and an identity mapping. This ensures that clear images retain their quality, avoiding performance drops in non-degraded scenarios.
\noindent \smallskip\\
\textbf{Inference.} During inference, only the RobustSAM (\figref{fig:overview}, top half) is used to generate the segmentation mask.

\smallskip
In the following, we give a detailed discussion on the proposed Anti-Degradation Output Token Generation and Anti-Degradation Mask Feature Generation modules.

\begin{figure*}[t!]
\centering \includegraphics[width=1.0\textwidth,page=1]{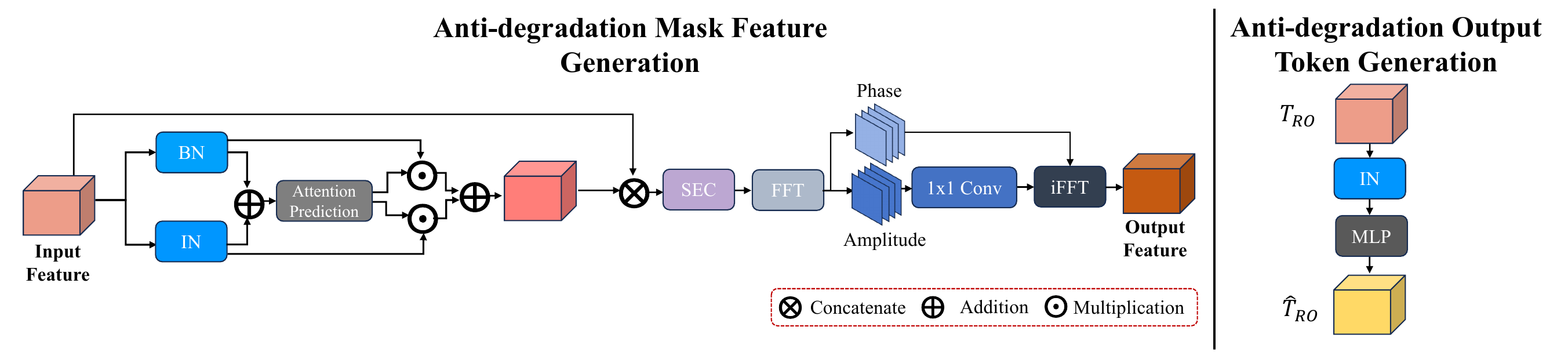}{}
\makeatother 
\caption{\textbf{Overview of the proposed Anti-degradation Mask Feature Generation (AMFG) and Anti-degradation Output Token Generation (AOTG).} SEC denotes Squeeze-and-Excitation Channel attention.}
\label{fig:style}
\end{figure*}

\subsubsection{Anti-Degradation Mask Feature Generation}
As shown in~\figref{fig:style}, the input features are first processed by Instance Normalization (IN). Inspired by previous work~\cite{son2020urie,ulyanov2016instance,huang2017arbitrary}, the purpose of IN is to standardize the variations associated with image degradation. Intuitively, this removes the \emph{style} attributes while preserving the core content. This step is essential to mitigate the influence of individual image distortions, ensuring the content's stability under diverse degradation conditions. Parallel to this, inspired by~\cite{son2020urie}, we include another branch that applies Batch Normalization (BN). BN is crucial as it addresses the potential loss of detail resulting from the IN process, as indicated by~\cite{nam2018batch,son2020urie}. 

We then merge the features generated by BN and IN individually. An attention mechanism scrutinizes the merged features to generate attention maps, which dynamically weigh the importance of each feature type, thus synthesizing a feature set that encapsulates the advantages of both normalization techniques~\cite{son2020urie}. To compensate for any semantic information that may have been lost, this enhanced feature set is concatenated with the original input features along the channel dimension. Additionally, we integrate channel attention, akin to the squeeze-and-excitation approach (SEC)~\cite{li2018recurrent,hu2018squeeze}, to refine the feature integration adaptively.

Inspired by~\cite{huang2021rda, yang2020fda, xu2021fourier, yu2022frequency, yang2023visual}, we introduced the Fourier Degradation Suppression module to enhance the integrated features by transforming them from the spatial to the frequency domain using the Fourier transform. This technique leverages the amplitude components to capture \emph{style} information about image degradation. By applying a 1×1 convolution, we focus on isolating and removing degradation elements. Meanwhile, the phase components are preserved to maintain the structural integrity. Following this, an inverse Fourier transform brings the refined features back to the spatial domain. 
This process treats degradations as image styles and generates degradation-invariant features for robust segmentation.
This module is applied to two features generated by preceding modules: the complementary feature (degrade) $F_{CFD}$ and the mask feature (degrade) $F_{MFD}$.
To ensure that these refined features maintain consistency with the corresponding features extracted by the SAM model (\textit{i.e.}, $F_{CFC}$ and $F_{MFC}$) when using clear images as input, we employ the Mask Feature Consistency Loss (\( \mathcal{L}_{\text{MFC}} \)).
\begin{equation}
\mathcal{L}_{\text{MFC}} = \lVert \hat{F}_{CFD} - F_{CFC} \rVert_2 + \lVert \hat{F}_{MFD} - F_{MFC} \rVert_2
\end{equation}
By minimizing each part of \( \mathcal{L}_{\text{MFC}} \), we ensure that the refined features remain consistent with those extracted under clear image conditions, thus guaranteeing the robustness and consistency of the features across different degradations.

\subsubsection{Anti-Degradation Output Token Generation}
The Anti-Degradation Output Token Generation module is dedicated to refining Robust Output Token per mask ($T_{RO}$) to remove degradation-related information. Unlike conventional mask features, $T_{RO}$ primarily functions to ensure the clarity of classification boundaries, thus containing less texture information. 
Therefore, we found that using a lightweight module to filter information sensitive to degradation is sufficient.
As depicted on the right side of Figure \ref{fig:style}, this module utilizes multiple layers of Instance Normalization followed by a single MLP layer.
This strategy aims to maintain computational efficiency while ensuring that the model can recover robust mask information from inputs affected by degradation. The refined token $\hat{T}_{RO}$ is then compared with the output token $T_{OC}$ extracted under clear input conditions by the original SAM to calculate Token Consistency Loss $\mathcal{L}_{\text{TC}}$, 
\begin{equation}
\mathcal{L}_{\text{TC}} = \lVert \hat{T}_{RO} - T_{OC} \rVert_2
\end{equation}
This loss ensures that the refined token remains consistent with those extracted under clear image conditions. After processing through the MLP, the output is combined with the Robust Mask Feature to generate the final mask.

\subsubsection{Overall Loss}
The overall loss function integrates the Mask Feature Consistency Loss (\( \mathcal{L}_{\text{MFC}} \)), Token Consistency Loss (\( \mathcal{L}_{\text{TC}} \)), and the Segmentation Loss (\( \mathcal{L}_{\text{Seg}} \)) to form a comprehensive penalty for the model. The overall loss is expressed as:
\begin{equation}
\mathcal{L}_{\text{Overall}} = \mathcal{L}_{\text{MFC}} + \lambda_{1}\mathcal{L}_{\text{TC}} + \lambda_{2}\mathcal{L}_{\text{Seg}}
\end{equation}
Here, \( \mathcal{L}_{\text{Seg}} \) is a combined segmentation loss that incorporates both Dice~\cite{lin2017focal} and Focal losses~\cite{sudre2017generalised}:
\begin{equation}
\mathcal{L}_{\text{Seg}} = \mathcal{L}_{\text{Dice}}(P, G) + \lambda_{3} \mathcal{L}_{\text{Focal}}(P, G)
\end{equation}
where \( P \) is the predicted mask, \( G \) is the ground truth mask, and $\lambda_{1}$--$\lambda_{3}$ are hyperparameters for weighting different losses. This composite loss function is designed to ensure enhancement in segmentation quality while bolstering the robustness of the model against degradation.

\section{Implementation Details}
\label{sec:implementation}
\subsection{Dataset}
To train and evaluate RobustSAM, we constructed a comprehensive \textbf{Robust-Seg} dataset, featuring 688,000 image-mask pairs. This dataset is composed of images from several existing datasets, specifically LVIS~\cite{gupta2019lvis}, ThinObject-5k~\cite{liew2021deep}, MSRA10K~\cite{ChengPAMI}, NDD20~\cite{trotter2020ndd20}, STREETS~\cite{snyder2019streets}, FSS-1000~\cite{FSS1000}, and COCO~\cite{lin2014microsoft}.
In this dataset, we incorporate original clear images and versions augmented with 15 types of synthetic degradations, including blur, noise, low light, adverse weather conditions, and so on. This approach ensures the model is extensively trained and robust to various image qualities.

During training, we utilize the entire training sets (and their augmentations) of MSRA10K, ThinObject-5k, and LVIS. The test sets (and their augmentations) of MSRA10k and LVIS are used to validate the segmentation accuracy of the model. To challenge the \emph{zero-shot} generalization of the model, we test it against the full range of images (and their augmentations) from the NDD20, STREETS, FSS-1000, and COCO datasets.

In addition, we conduct extensive testing using the complete BDD-100k~\cite{yu2018bdd100k} and LIS~\cite{Hong2021Crafting,2023lis} datasets, which include a variety of real-world degradations such as low-light, blur, rain, and snow. This approach ensures a thorough evaluation of RobustSAM's performance in realistic scenarios and its robustness to adverse environmental conditions typically encountered in real-world applications.

\subsection{Training Detail}
During the training phase of RobustSAM, we keep the pre-trained SAM parameters frozen, focusing only on optimizing the proposed modules for robustness. We train RobustSAM using point-based prompts.

RobustSAM significantly enhances segmentation quality and is designed for fast and efficient training.
With a learning rate of 0.0005 for 40 epochs, the training is completed in 30 hours for 130,000 iterations on 8 Nvidia A100 GPUs. \tabref{tab:comp_compute} details the training and inference performance of RobustSAM compared to SAM. RobustSAM not only delivers improved segmentation outcomes but does so with remarkable training efficiency compared to SAM.

\begin{table}[t!]
\small
\begin{center}
\scalebox{0.75}{
\begin{tabular}{lcccccc}
\toprule
& \multicolumn{4}{c}{Training} & \multicolumn{2}{c}{Inference} \\
\cmidrule(lr){2-5} \cmidrule(lr){6-7}
& \makecell{Learnable \\ Params} & \# GPU & \makecell{Batch \\ Size} & \makecell{Time \\ (h)} & FPS & Mem. \\ 
\midrule
SAM       & 1250 MB & 256 & 256 & N/A  & 2.90 & 9.63G \\
RobustSAM & 403 MB  & 8   & 8   & 30   & 2.80 & 10.08G \\
\bottomrule
\end{tabular}
}
\end{center}
\vspace{-0.5cm}
\caption{\textbf{Comparative computational requirements for SAM and RobustSAM.}}
\label{tab:comp_compute}
\end{table}

\subsection{Evaluation Protocol}
To evaluate the performance of RobustSAM, we employ several metrics: Intersection over Union (IoU), Dice Coefficient (Dice), Pixel Accuracy (PA), and Average Precision (AP) at various threshold levels.

\begin{figure*}[t!]
  \centering
  \setlength{\tabcolsep}{1pt}
  \begin{tabular}{cccccc}
    Input & SAM & HQ-SAM & AirNet+SAM & URIE+SAM & RobustSAM \\

    \includegraphics[width=.16\textwidth, height=2.cm]{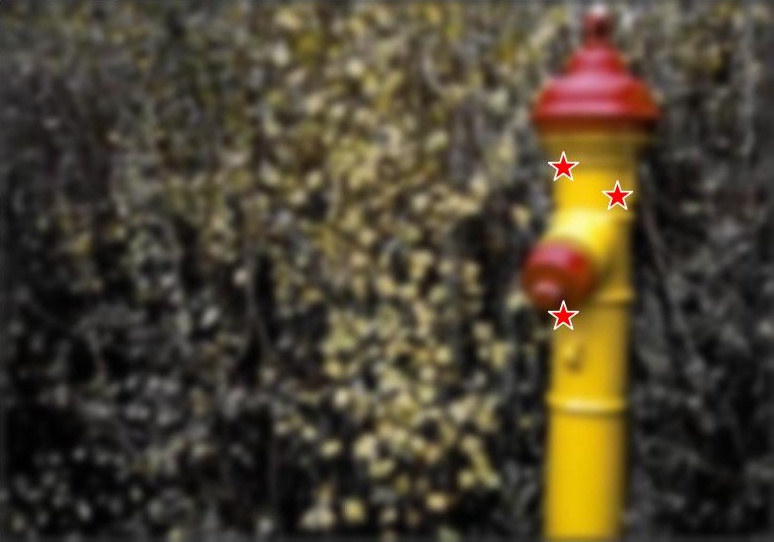} &
    \includegraphics[width=.16\textwidth, height=2.cm]{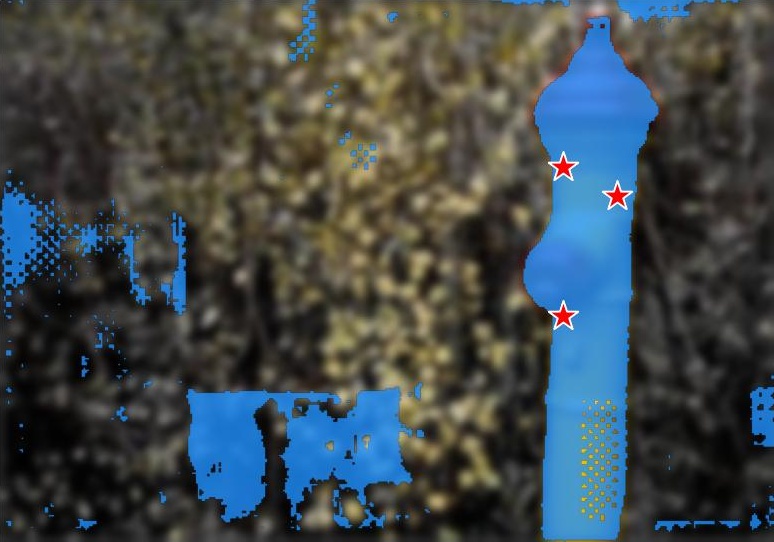} & 
    \includegraphics[width=.16\textwidth, height=2.cm]{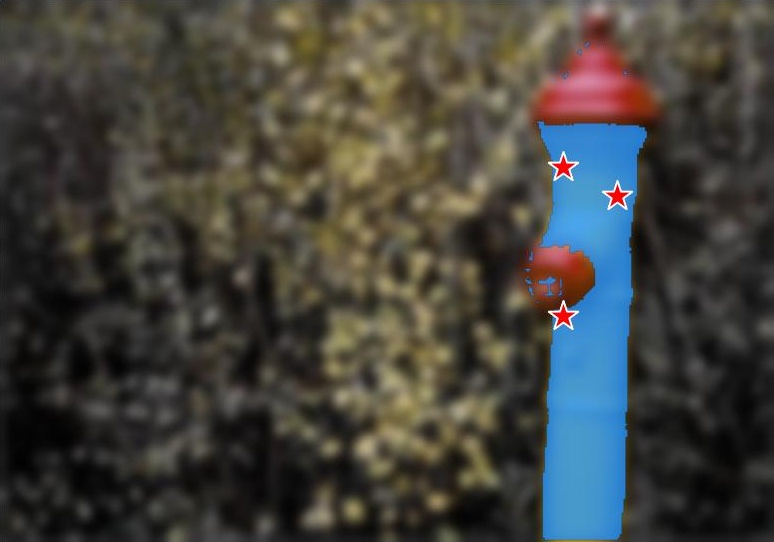} &
    \includegraphics[width=.16\textwidth, height=2.cm]{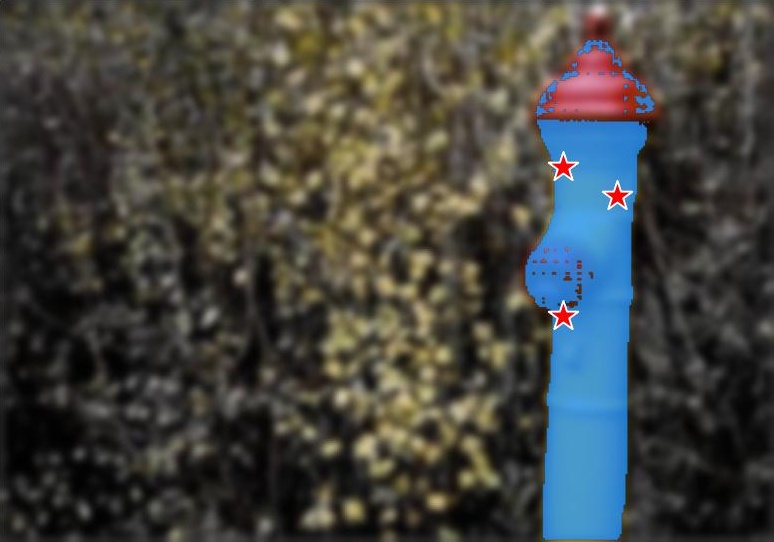} & 
    \includegraphics[width=.16\textwidth, height=2.cm]{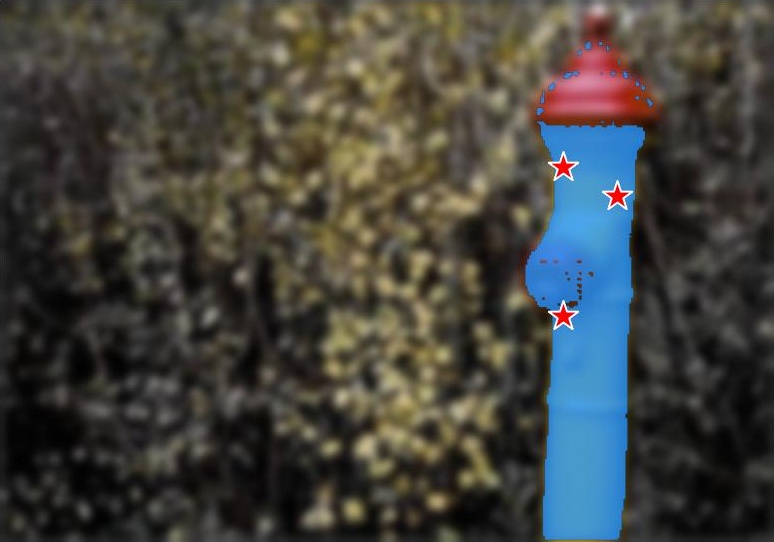} & 
    \includegraphics[width=.16\textwidth, height=2.cm]{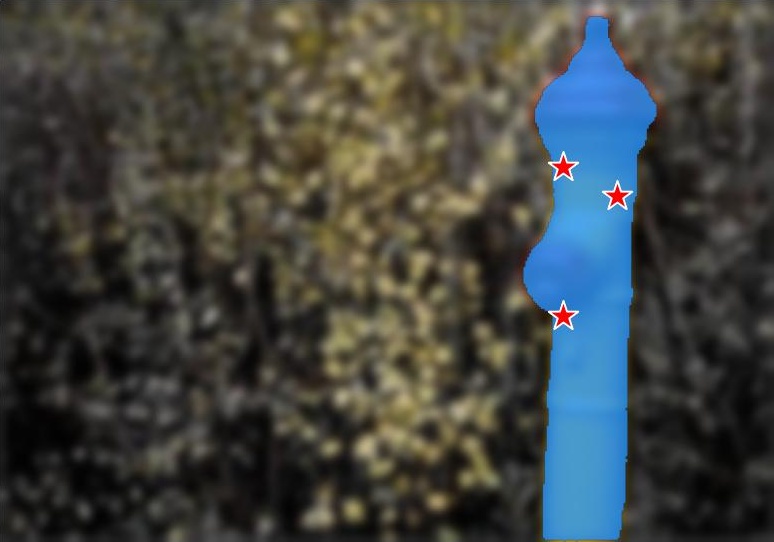} \\
    \multicolumn{6}{c}{\vspace{-14pt}} \\

    \includegraphics[width=.16\textwidth, height=2.cm]{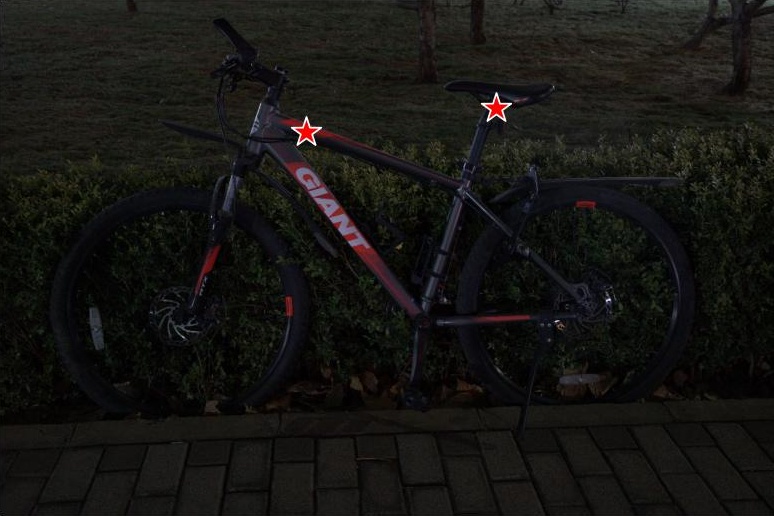} & 
    \includegraphics[width=.16\textwidth, height=2.cm]{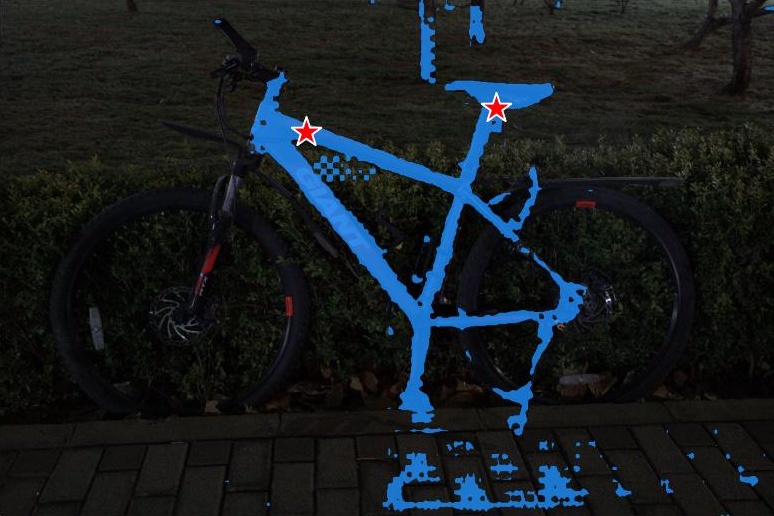} & 
    \includegraphics[width=.16\textwidth, height=2.cm]{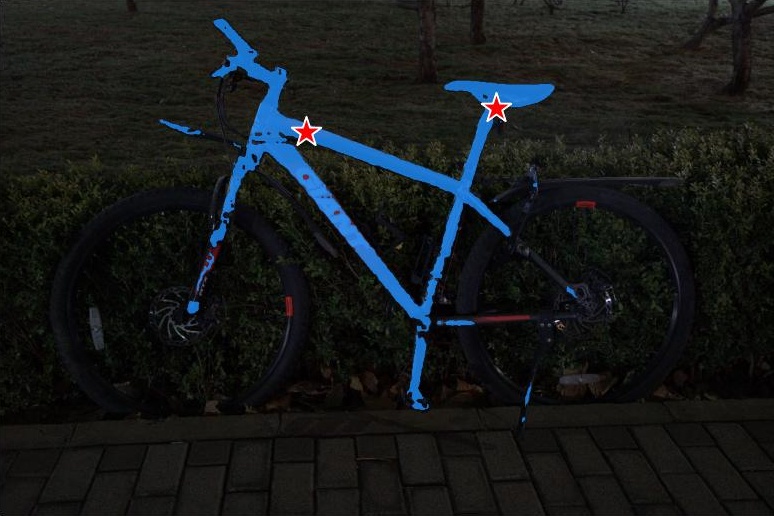} & 
    \includegraphics[width=.16\textwidth, height=2.cm]{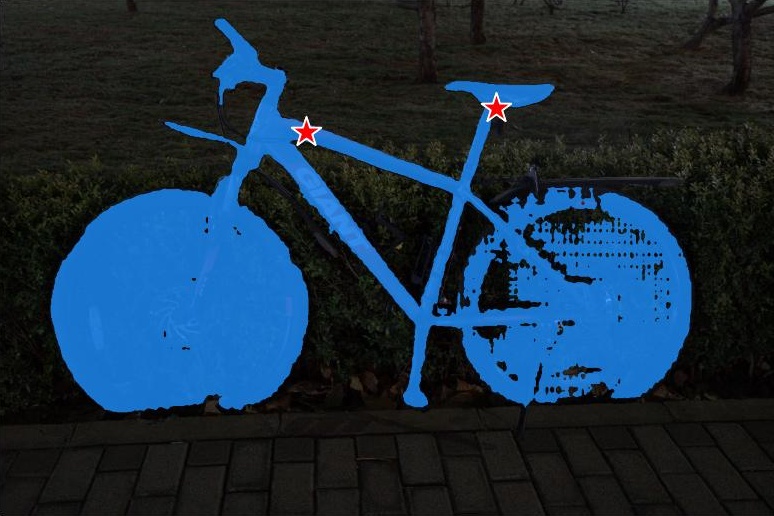} &
    \includegraphics[width=.16\textwidth, height=2.cm]{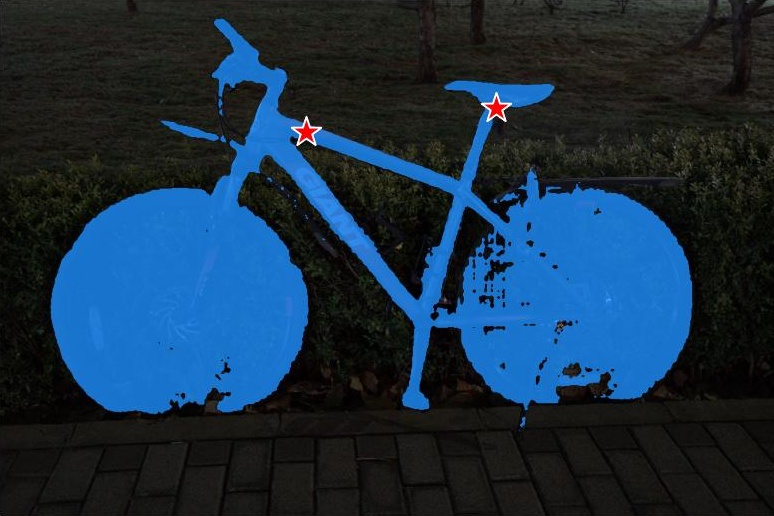} & 
    \includegraphics[width=.16\textwidth, height=2.cm]{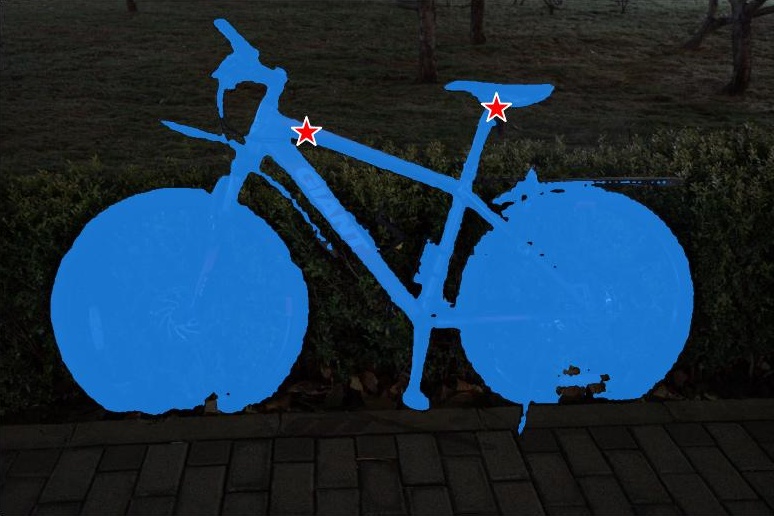} \\

    \multicolumn{6}{c}{\vspace{-14pt}} \\
    \includegraphics[width=.16\textwidth, height=2.1cm]{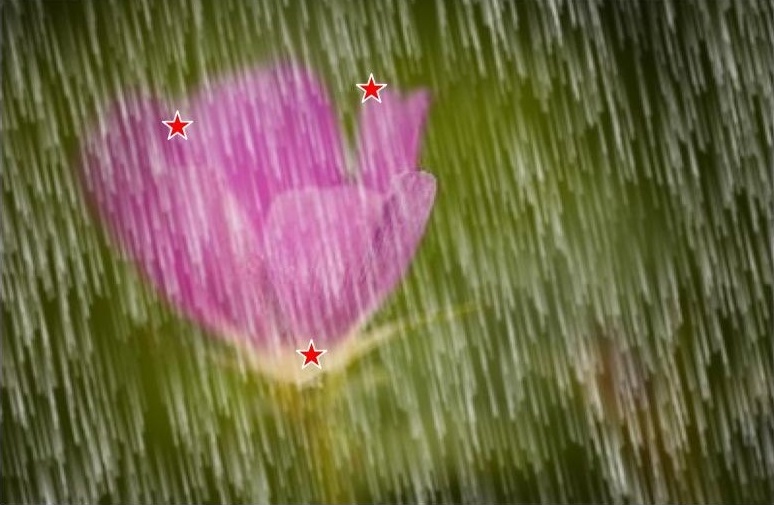} &
    \includegraphics[width=.16\textwidth, height=2.1cm]{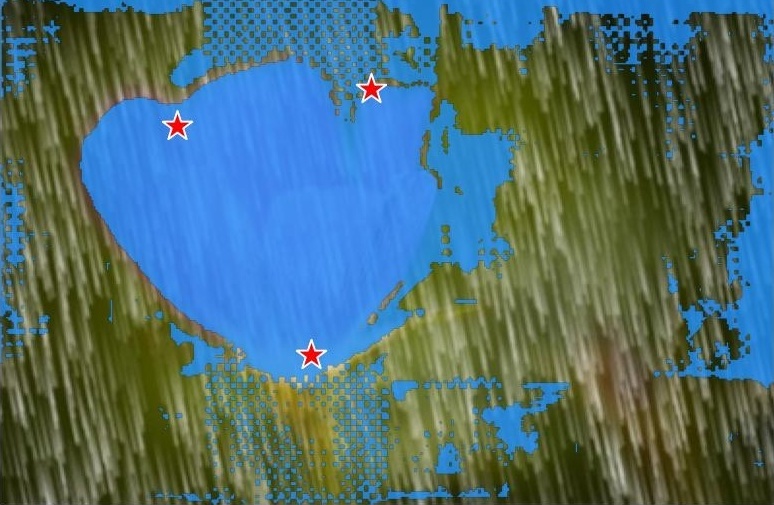} &
    \includegraphics[width=.16\textwidth, height=2.1cm]{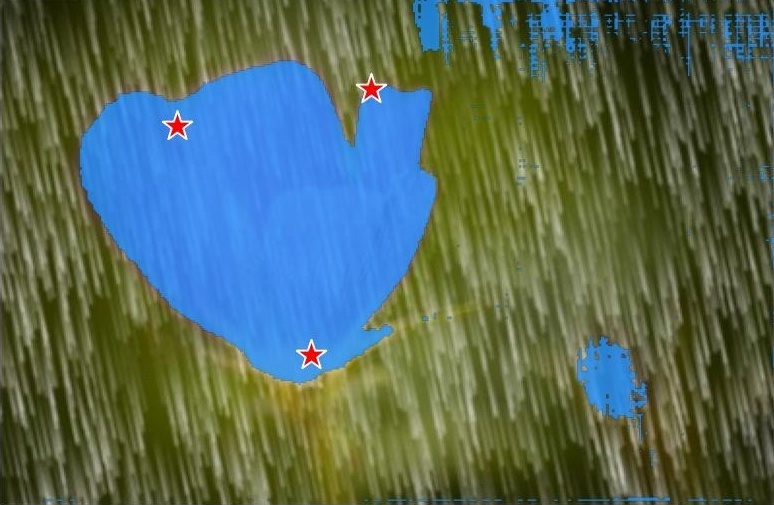} & 
    \includegraphics[width=.16\textwidth, height=2.1cm]{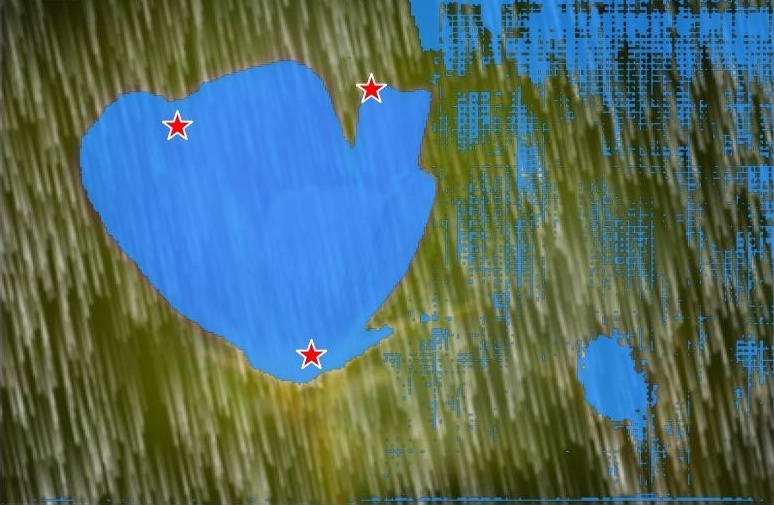} &
    \includegraphics[width=.16\textwidth, height=2.1cm]{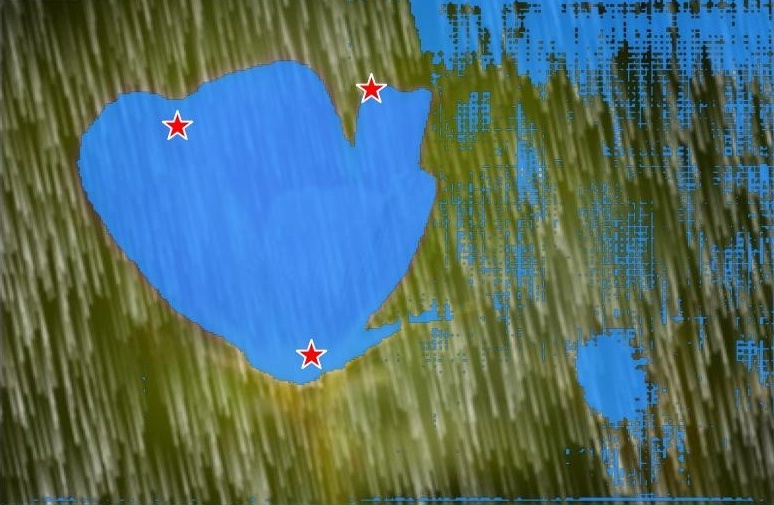} & 
    \includegraphics[width=.16\textwidth, height=2.1cm]{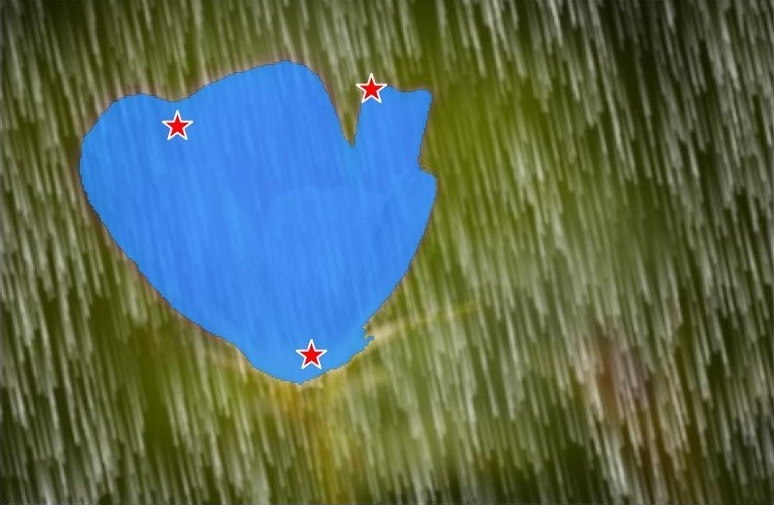} 
\\    
        \multicolumn{6}{c}{\vspace{-14pt}} \\
    \includegraphics[width=.16\textwidth, height=2.5cm, clip, trim={0cm 0.7cm 0cm 0.7cm}]{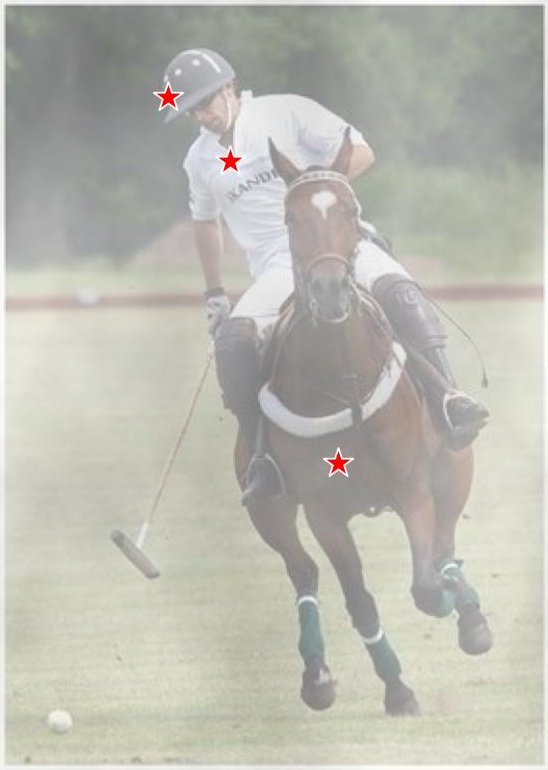} &
    \includegraphics[width=.16\textwidth, height=2.5cm, clip, trim={0cm 0.7cm 0cm 0.7cm}]{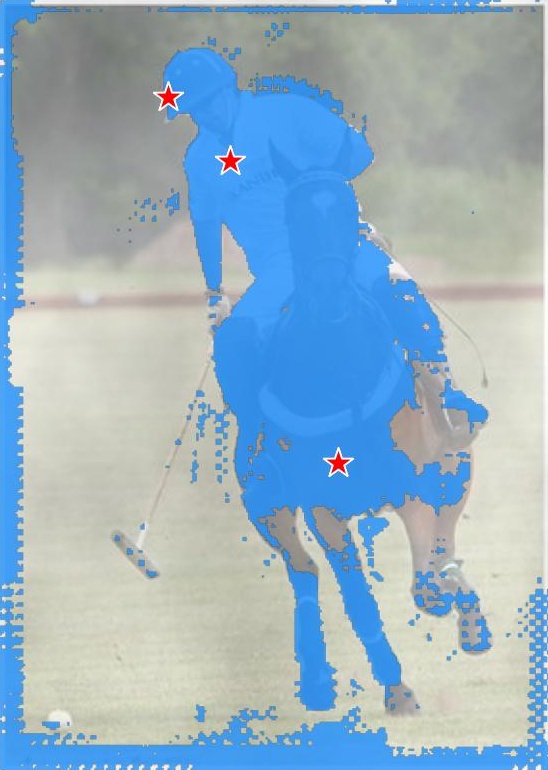} &
    \includegraphics[width=.16\textwidth, height=2.5cm, clip, trim={0cm 0.7cm 0cm 0.7cm}]{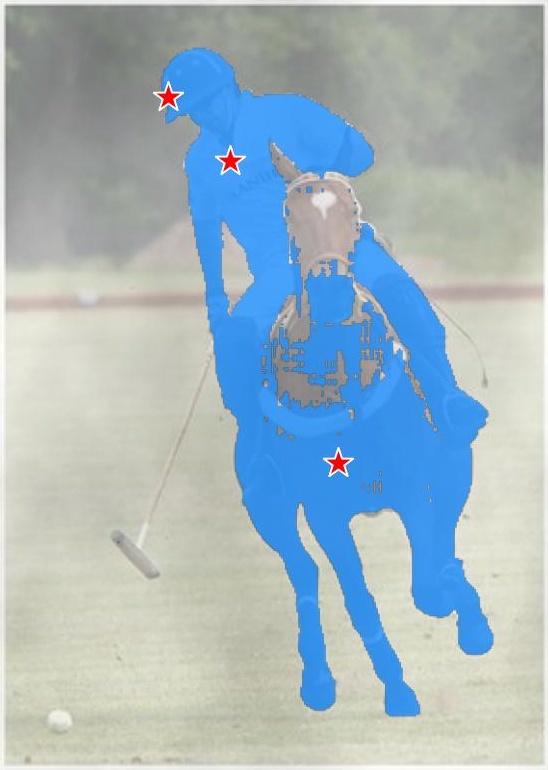} &
    \includegraphics[width=.16\textwidth, height=2.5cm, clip, trim={0cm 0.5cm 0cm 0.5cm}]{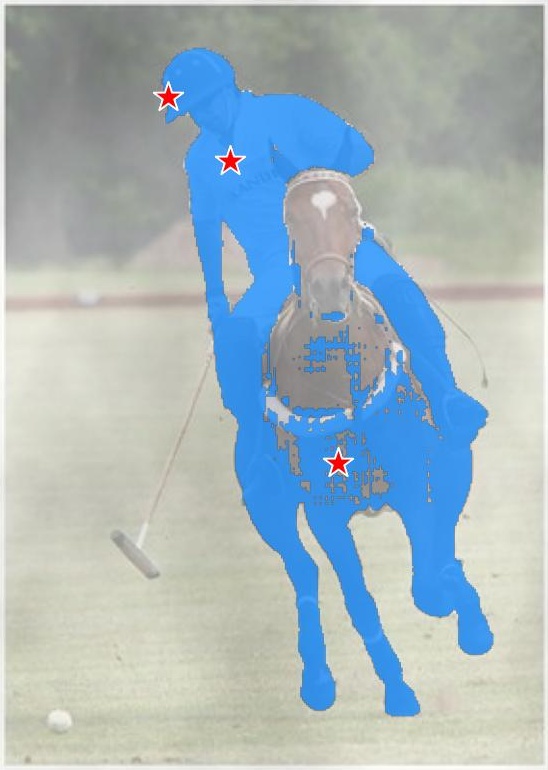} &
    \includegraphics[width=.16\textwidth, height=2.5cm, clip, trim={0cm 0.5cm 0cm 0.5cm}]{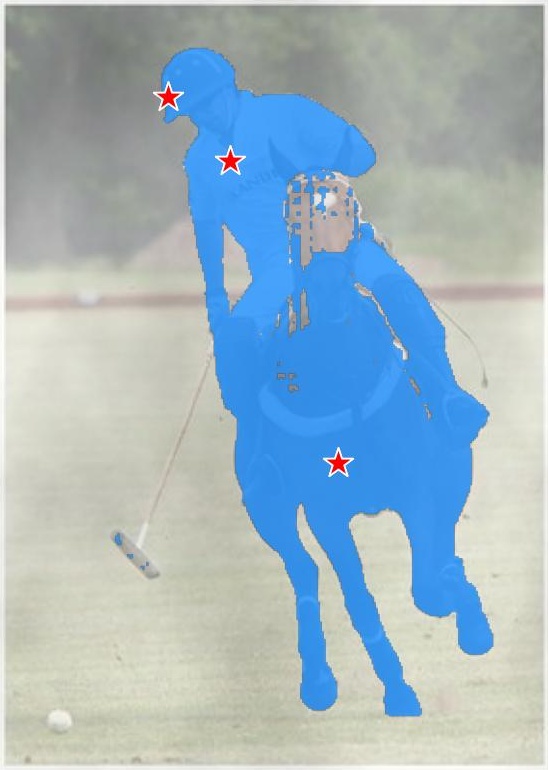} &
    \includegraphics[width=.16\textwidth, height=2.5cm, clip, trim={0cm 0.7cm 0cm 0.7cm}]{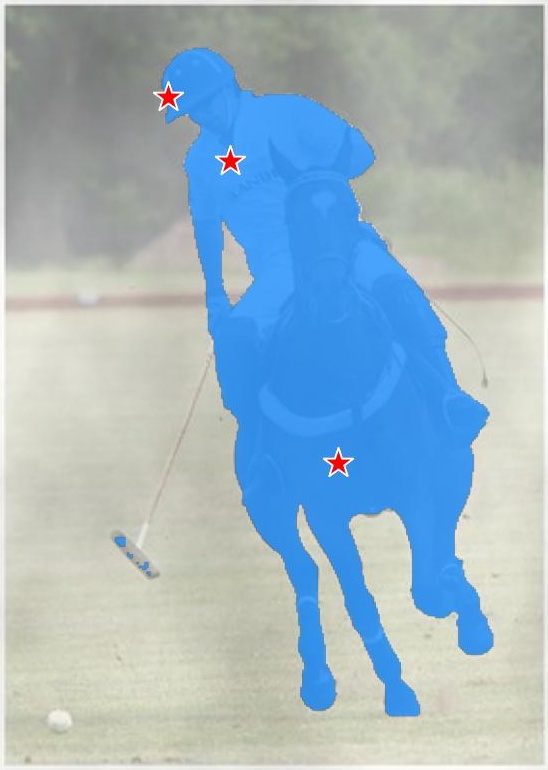}     
    \\    
  \end{tabular}
\caption{\textbf{Qualitative Analysis of Segmentation:} A visual comparison on unseen datasets highlighting the performance improvements of the RobustSAM over existing strategies.}
  \label{fig:qualitative_ori}
\end{figure*}

\section{Experimental Results}\label{sec:exp}
In this section, we present the evaluation results of the proposed RobustSAM.

\begin{table}[t!]
\centering
\scalebox{0.68}{
\begin{tabular}{ccccccc}
\toprule
\multirow{2}{*}{Method} & \multicolumn{2}{c}{Degrade} & \multicolumn{2}{c}{Clear} & \multicolumn{2}{c}{Average}  \\ 
\cmidrule(lr){2-3} \cmidrule(lr){4-5} \cmidrule(lr){6-7}
                        & IoU & PA & IoU & PA & IoU & PA \\
\midrule

SAM                 & 0.8194 & 0.9108 & 0.8402 & 0.9235 & 0.8207 & 0.9116 \\
HQ-SAM                 & \underline{0.8358} & \underline{0.9202} & \underline{0.8604} & \underline{0.9328} & \underline{0.8373} & \underline{0.9210} \\
AirNet+SAM                 & 0.8157 & 0.9193 & 0.8236 & 0.9294 & 0.8162 & 0.9199 \\
URIE+SAM                 & 0.8217 & 0.9125 & 0.8450 & 0.9245 & 0.8231 & 0.9132 \\
\rowcolor{LightCyan}
\textbf{RobustSAM}                 & \textbf{0.8609} & \textbf{0.9640} & \textbf{0.8726} & \textbf{0.9649} & \textbf{0.8616} & \textbf{0.9641} \\
\bottomrule
\end{tabular}}
\caption{\textbf{Performance Comparison on test set of MSRA10k~\cite{ChengPAMI} datasets (seen datasets with synthetic degradations) in Robust-Seg dataset using point prompts for ``Degrade'', ``Clear'', and ``Average'' scenarios.} ``Degrade'' refers to the set of images subjected to 15 different types of degradation, ``Clear'' refers to the original, undegraded images, and ``Average'' represents the weighted sum average of the ``Degrade'' and ``Clear'' scenarios. The words with \textbf{boldface} indicate the best results, and those with \underline{underline} indicate the second-best results.}
\label{tab:quantitative_syn}
\end{table}

\begin{table}[t!]
\centering
\scalebox{0.68}{
\begin{tabular}{ccccccc}
\toprule
\multirow{2}{*}{Method} & \multicolumn{2}{c}{Degrade} & \multicolumn{2}{c}{Clear} & \multicolumn{2}{c}{Average}  \\ 
\cmidrule(lr){2-3} \cmidrule(lr){4-5} \cmidrule(lr){6-7}
                        & IoU & PA & IoU & PA & IoU & PA \\
\midrule

SAM                 & 0.7341 & 0.9181 & 0.7415 & 0.9282 & 0.7346 & 0.9187 \\
HQ-SAM                 &\underline{0.7405} & \underline{0.9242} & \underline{0.7502} & \underline{0.9319} & \underline{0.7411} & \underline{0.9246} \\
AirNet+SAM                 & 0.7352 & 0.9198 & 0.7419 & 0.9293 & 0.7356 & 0.9204 \\
URIE+SAM                 & 0.7336 & 0.9182 & 0.7406 & 0.9277 & 0.7340 & 0.9188 \\
\rowcolor{LightCyan}
\textbf{RobustSAM}                 & \textbf{0.7506} & \textbf{0.9327} & \textbf{0.7592} & \textbf{0.9339} & \textbf{0.7511} & \textbf{0.9328} \\
\bottomrule
\end{tabular}}
\caption{\textbf{Performance comparison on the test set of LVIS~\cite{gupta2019lvis} dataset (a seen dataset with synthetic degradations) in Robust-Seg dataset using Bounding Box prompts.}}
\label{tab:seen_dataset}
\end{table}

\begin{table}[t!]
\centering
\scalebox{0.72}{
\begin{tabular}{ccccccc}
\toprule
\multirow{2}{*}{Method} & \multicolumn{2}{c}{Degrade} & \multicolumn{2}{c}{Clear} & \multicolumn{2}{c}{Average}  \\ 
\cmidrule(lr){2-3} \cmidrule(lr){4-5} \cmidrule(lr){6-7}
                        & IoU & PA & IoU & PA & IoU & PA \\
\midrule

SAM                 & 0.7981 & 0.9555  & 0.8295  & 0.9707  & 0.8000  & 0.9565  \\
HQ-SAM                 & \underline{0.8079}  & 0.9617  & \underline{0.8448}  & \underline{0.9756}  & \underline{0.8102}  & 0.9625 \\
AirNet+SAM                 & 0.7988 & \underline{0.9629}  & 0.8312 & 0.9752  & 0.8008  & \underline{0.9637} \\
URIE+SAM                 & 0.7904  & 0.9593  & 0.8288  & 0.9740  & 0.7928  & 0.9602  \\
\rowcolor{LightCyan}
\textbf{RobustSAM}                 & \textbf{0.8195}  & \textbf{0.9778}  & \textbf{0.8529}  & \textbf{0.9817} & \textbf{0.8216}  & \textbf{0.9780}  \\
\bottomrule
\end{tabular}}
\caption{\textbf{Zero-shot segmentation comparison on the whole NDD20~\cite{trotter2020ndd20}, STREETS~\cite{snyder2019streets}, and FSS-1000~\cite{FSS1000} (unseen datasets with synthetic degradations) in Robust-Seg dataset using point prompts.}}
\label{tab:eval_rest_ori}
\end{table}

\begin{table}[t!]
\centering
\scalebox{0.75}{
\begin{tabular}{ccccc}
\toprule
\multirow{2}{*}{Method} & \multicolumn{4}{c}{Performance Metrics} \\
\cmidrule(lr){2-5}
                        & AP & $\text{AP}_\text{S}$ & $\text{AP}_\text{M}$ & $\text{AP}_\text{L}$ \\
\midrule
SAM                    & 0.5002 & 0.3168 & 0.4292 & 0.5243 \\
HQ-SAM                 & \underline{0.5052} & 0.2920 & 0.4267 & \underline{0.5517} \\
AirNet+SAM             & 0.4926 & 0.3068 & 0.4263 & 0.5187 \\
URIE+SAM               & 0.4980 & \underline{0.3186} & \underline{0.4319} & 0.5215 \\
\rowcolor{LightCyan}
\textbf{RobustSAM}     & \textbf{0.5130} & \textbf{0.3192} & \textbf{0.4416} & \textbf{0.5518} \\
\bottomrule
\end{tabular}}
\caption{\textbf{Zero-shot segmentation comparison on the whole COCO~\cite{lin2014microsoft} (unseen datasets with synthetic degradations) in Robust-Seg dataset using Bounding Box prompts.}}
\label{tab:eval_coco_ori}
\end{table}

\begin{table}[t!]
\centering
\scalebox{0.75}{
\begin{tabular}{cccccc}
\toprule
\multirow{2}{*}{Method} & \multicolumn{2}{c}{Point} & \multicolumn{2}{c}{Bounding Box}  \\ 
\cmidrule(lr){2-3} \cmidrule(lr){4-5}
                        & IoU & Dice & IoU & Dice \\
\midrule
\midrule

SAM                 & 0.3056 & 0.3837 & 0.8808 & 0.9171 \\
HQ-SAM              & 0.2943 & 0.3712 & \underline{0.8877} & \underline{0.9245} \\
AirNet+SAM          & \underline{0.3245} & \underline{0.4550} & 0.8776 & 0.9129 \\
URIE+SAM            & 0.3042 & 0.3828 & 0.8799 & 0.9165 \\
\rowcolor{LightCyan}
\textbf{RobustSAM}  & \textbf{0.3717} & \textbf{0.8926} & \textbf{0.8958} & \textbf{0.9416} \\
\bottomrule
\end{tabular}}
\caption{\textbf{Zero-shot segmentation comparison on the whole BDD-100k~\cite{yu2018bdd100k} and LIS~\cite{Hong2021Crafting,2023lis} datasets  (unseen datasets with real-world degradations) using point prompts.}}
\label{tab:eval_real_ori}
\end{table}

\subsection{Performance Evaluation}
In the landscape of image segmentation under challenging conditions, our RobustSAM framework is compared with several existing methods to establish its efficacy. We benchmark against the foundational SAM model and a strategic two-stage methodology where images are first passed through universal image restoration techniques to refine input quality, subsequently followed by SAM-driven segmentation. To that end, we incorporate AirNet~\cite{li2022all}, a state-of-the-art general visual quality enhancement method tailored for \textit{unknown} degradation.
Furthermore, we integrate URIE~\cite{son2020urie}, an image restoration approach optimized to set the stage for more effective segmentation. Additionally, we compare with HQ-SAM~\cite{sam_hq}, which is a high-quality iteration of the original SAM. 
\noindent \smallskip\\
\textbf{Comparison on Seen Datasets.} We evaluate the performance of our proposed methods within the framework of Robust-Seg on seen datasets: LVIS~\cite{gupta2019lvis} and MSRA10K~\cite{ChengPAMI}. Specifically, we assess the methods on the test sets of these datasets. The results show that our approach yields superior performance, effectively handling the varied challenges posed by these diverse scenes. Additionally, RobustSAM demonstrates its strength in significantly improving segmentation in degraded scenarios while maintaining or enhancing performance in clear scenes. Detailed results are presented in \tabref{tab:quantitative_syn} and \tabref{tab:seen_dataset}, demonstrating our method's efficacy across different segmentation scenarios.
\noindent \smallskip\\
\textbf{Zero-shot Segmentation Comparison.} 
We first assessed our methods on the entire NDD20~\cite{trotter2020ndd20}, STREETS~\cite{snyder2019streets}, and FSS-1000~\cite{FSS1000} and COCO~\cite{lin2014microsoft} datasets, all of which are synthesized degradations designed to challenge segmentation models. These test sets' results, which highlight our approaches' adaptability and accuracy, are detailed in \tabref{tab:eval_rest_ori} and \tabref{tab:eval_coco_ori}. Moreover, to validate our methods' capabilities against real-world degradations, we expanded our evaluation to include the entire BDD-100k~\cite{yu2018bdd100k} and LIS datasets~\cite{Hong2021Crafting,2023lis}. These datasets are particularly challenging due to their wide variety of real-world degradations, ranging from weather conditions to noise and variable lighting. The results of these assessments are provided in~\tabref{tab:eval_real_ori}. In addition, we present the qualitative comparison of segmentation in \figref{fig:qualitative_ori}. These results indicate that RobustSAM possesses significant robustness in zero-shot segmentation, maintaining high performance across different degradations.

\begin{figure}[t!]
  \centering
  \setlength{\tabcolsep}{1pt} 
  \scalebox{0.8}{
    \begin{tabular}{cccc}

      & Dehazing &\hspace{0.1cm} Deblurring & \\

      \rotatebox[origin=c]{90}{Input} & 
      \includegraphics[valign=m,width=.5\linewidth, height=1.3cm]{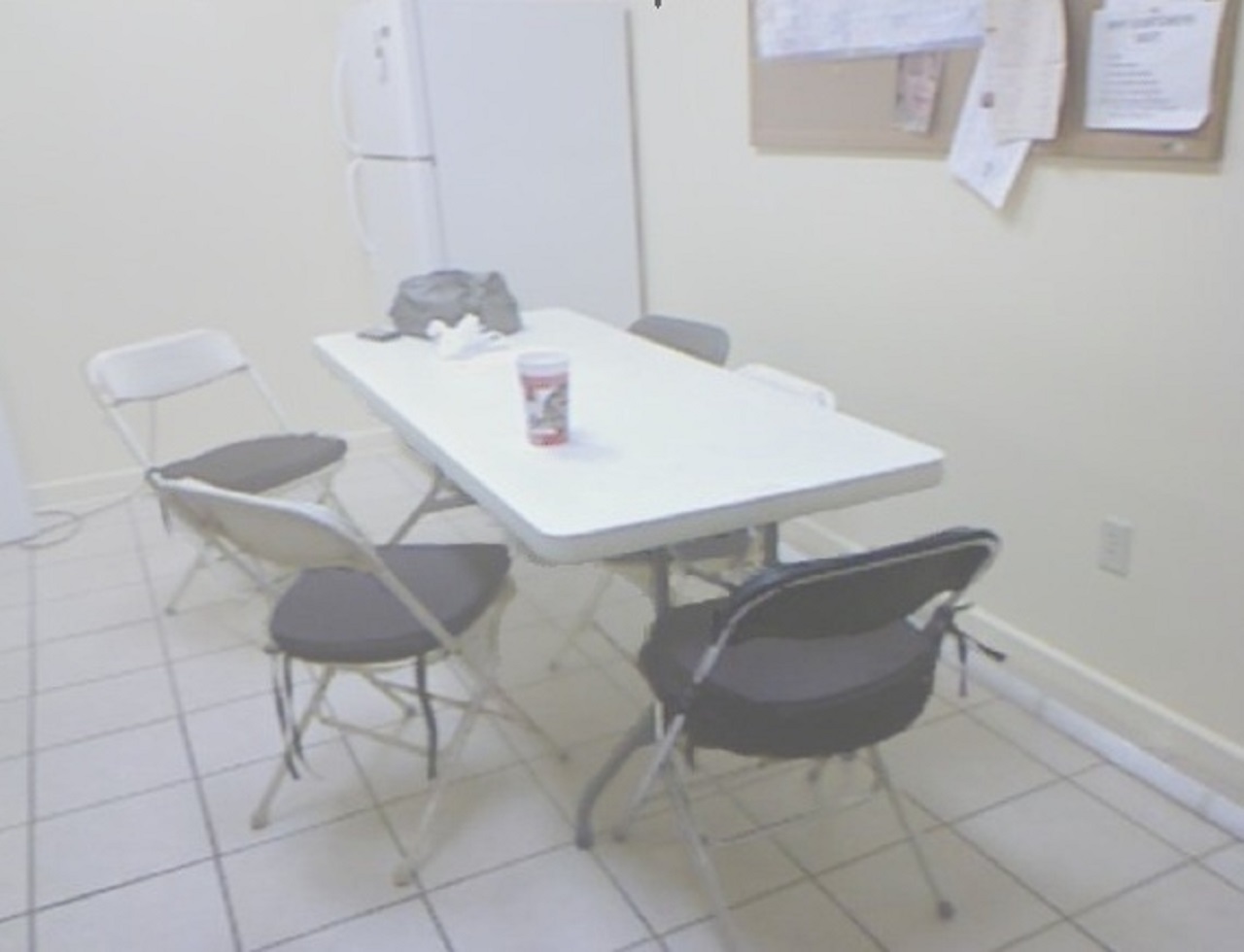} &
      \hspace{0.1cm}
      \includegraphics[valign=m,width=.5\linewidth, height=1.3cm]{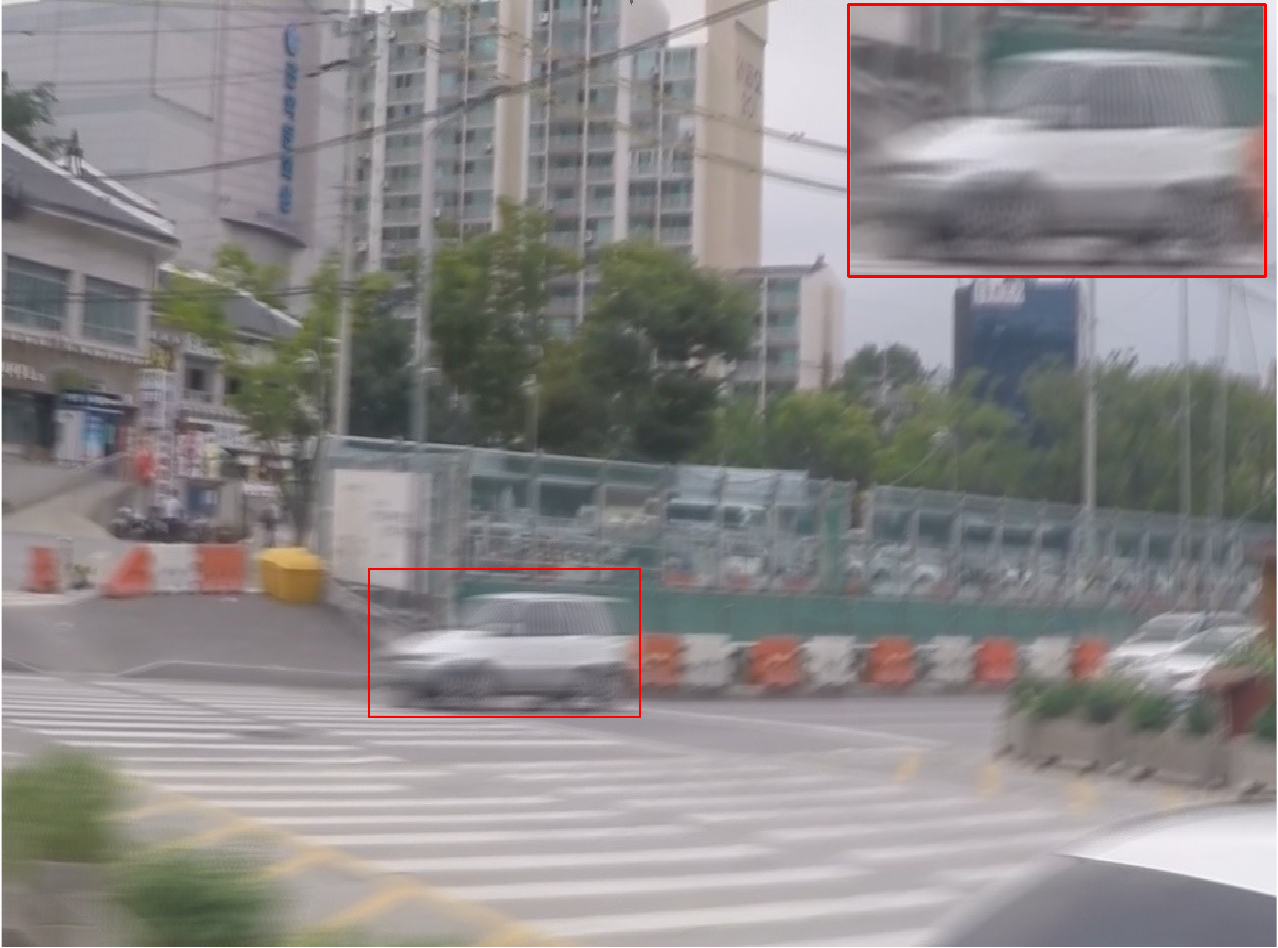} & \\
      \multicolumn{4}{c}{\vspace{-10pt}} \\ 
      
      \rotatebox[origin=c]{90}{SAM} &
      \includegraphics[valign=m,width=.5\linewidth, height=1.3cm]{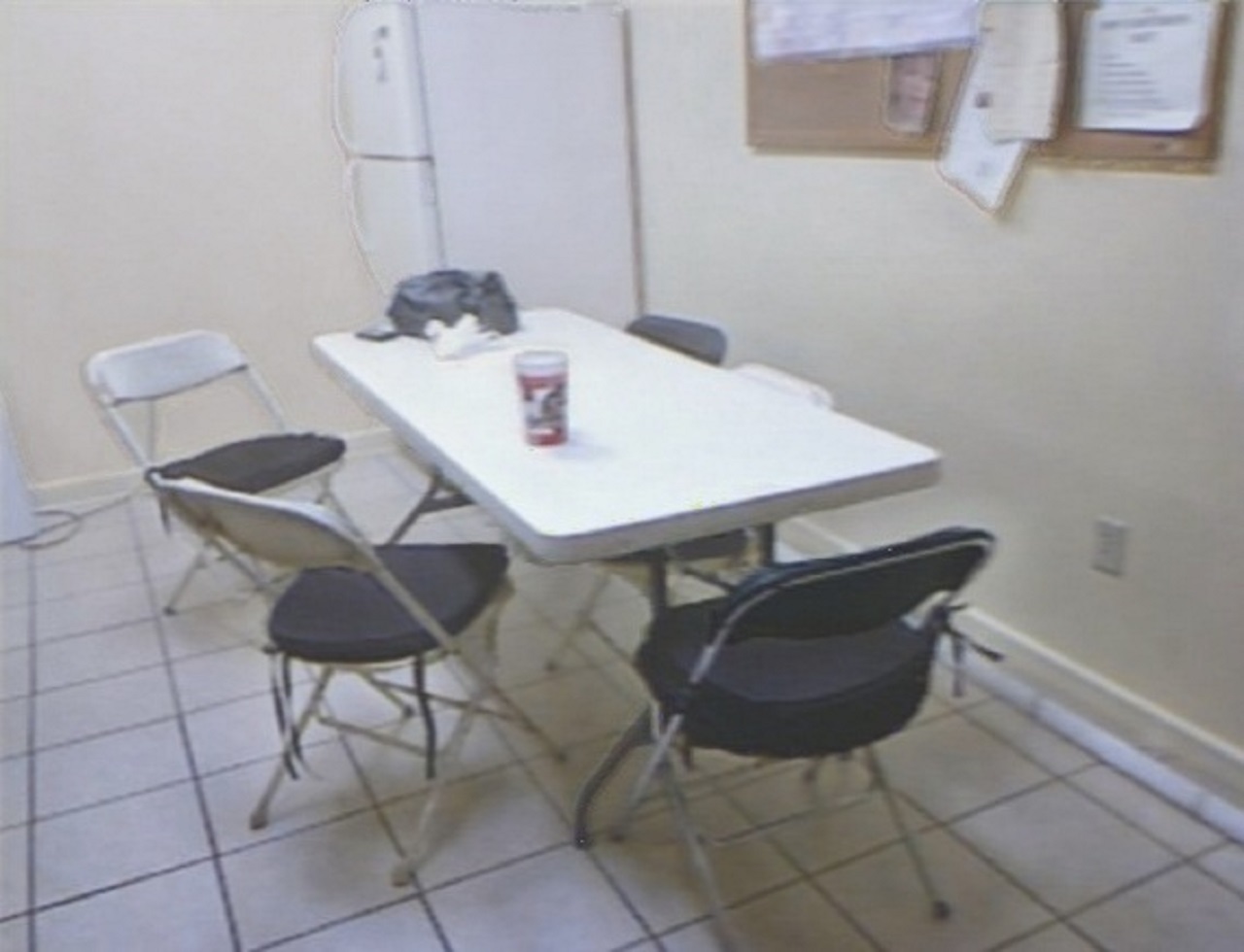} &
      \hspace{0.1cm}
      \includegraphics[valign=m,width=.5\linewidth, height=1.3cm]{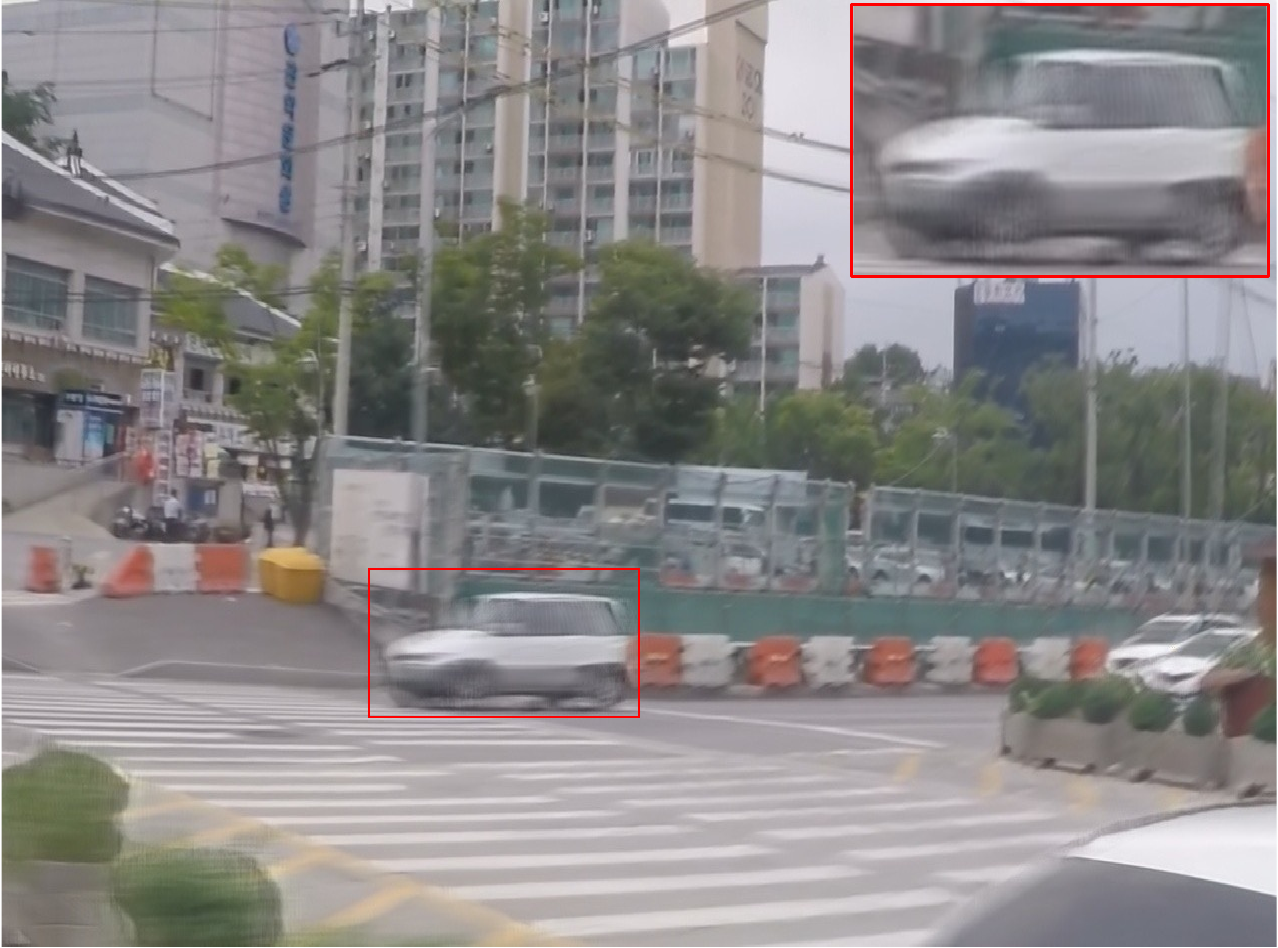} & \\
      \multicolumn{4}{c}{\vspace{-10pt}} \\ 
      
      \rotatebox[origin=c]{90}{RobustSAM} &
      \includegraphics[valign=m,width=.5\linewidth, height=1.3cm]{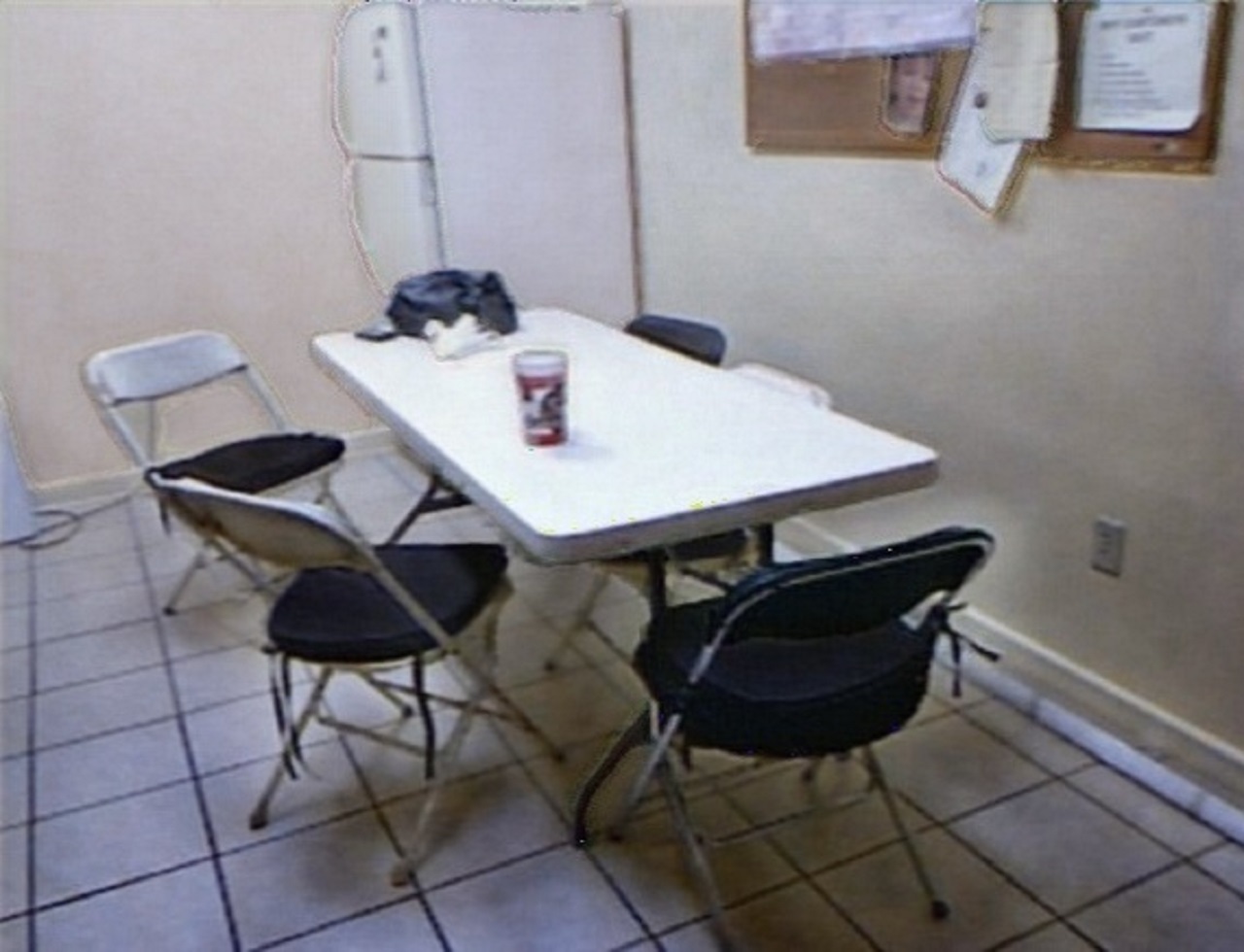} &
      \hspace{0.1cm}
      \includegraphics[valign=m,width=.5\linewidth, height=1.3cm]{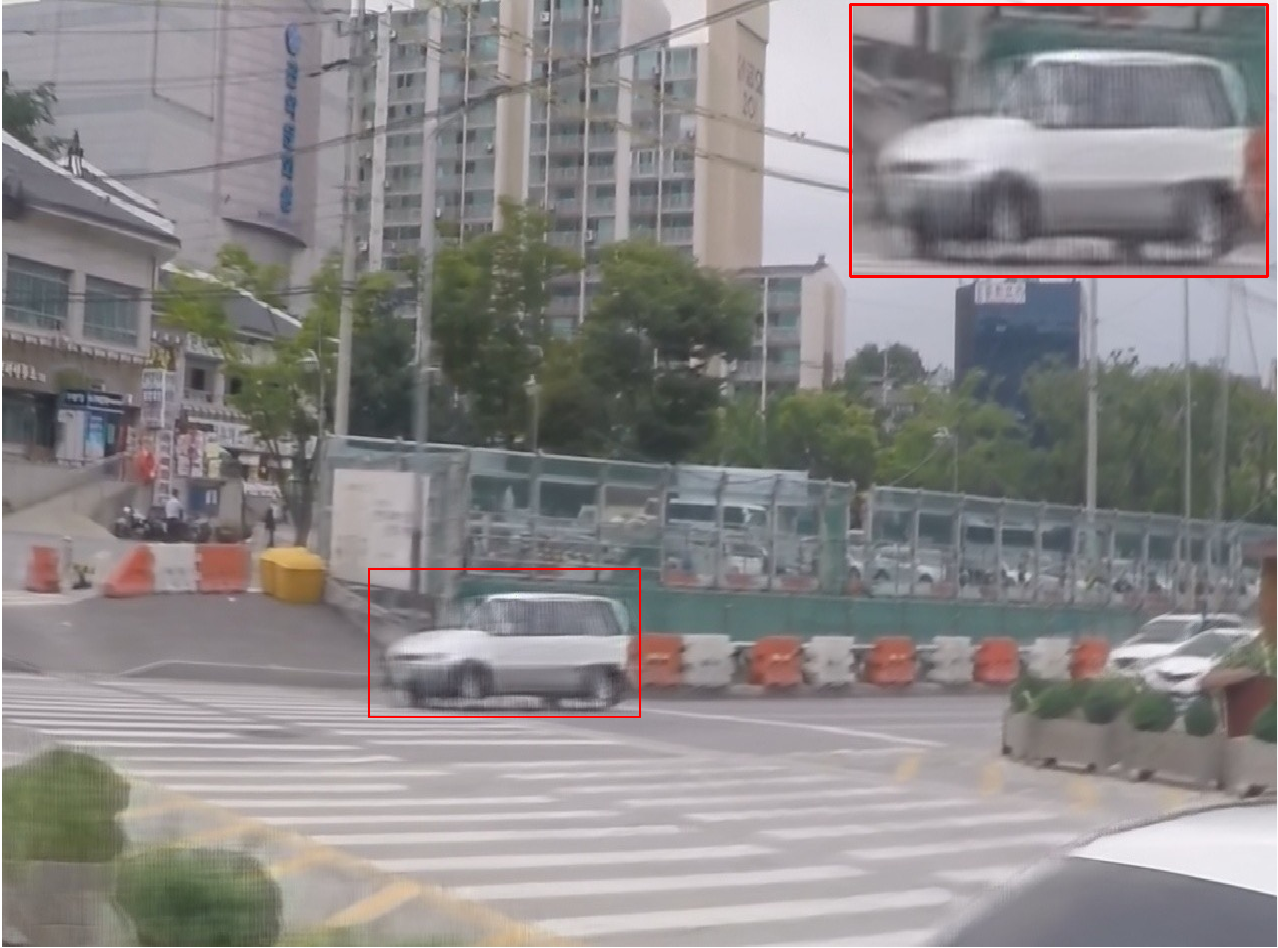} & \\
      \multicolumn{4}{c}{\vspace{-10pt}} \\ 
      
      \rotatebox[origin=c]{90}{Ground Truth} &
      \includegraphics[valign=m,width=.5\linewidth, height=1.3cm]{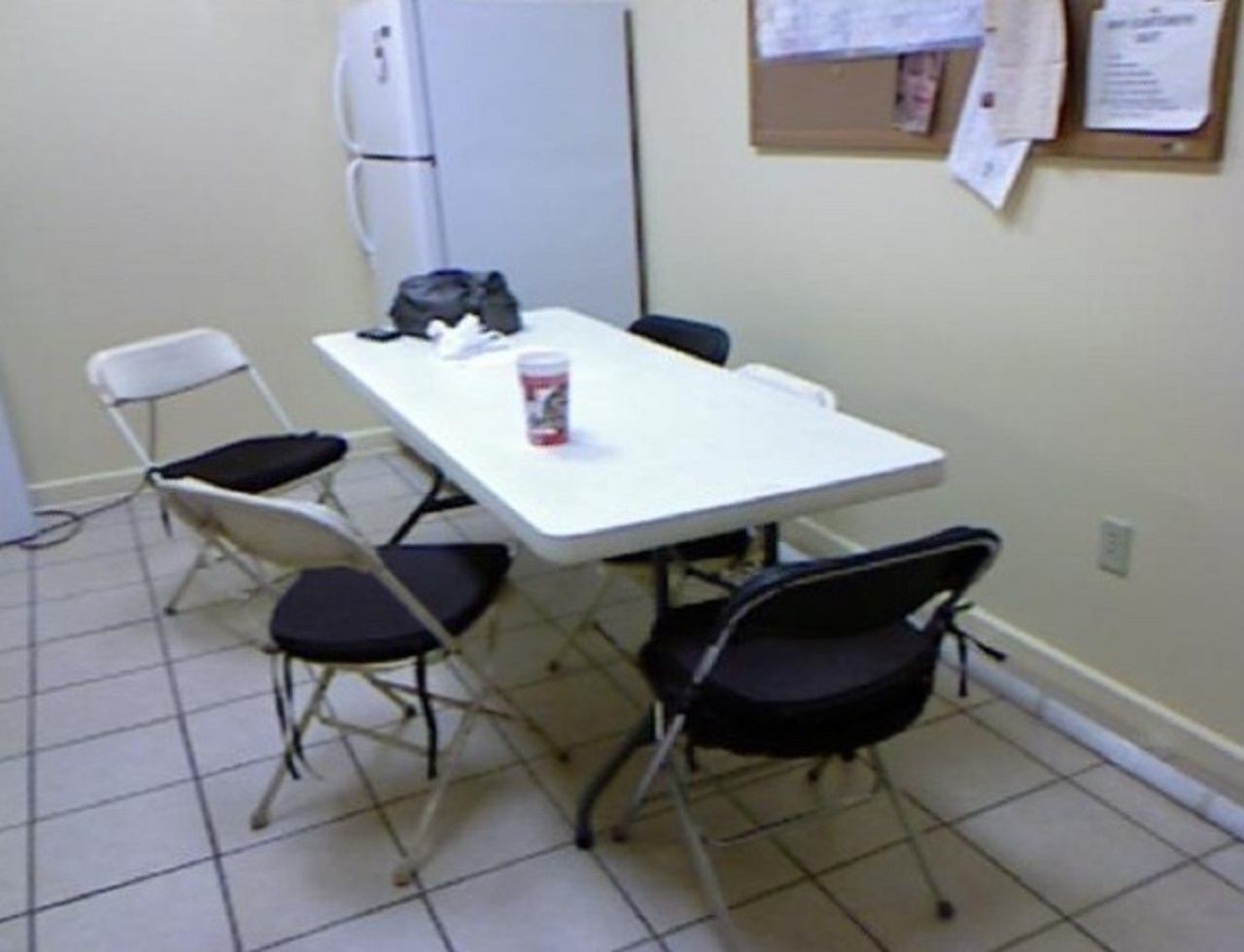} &
      \hspace{0.1cm}
      \includegraphics[valign=m,width=.5\linewidth, height=1.3cm]{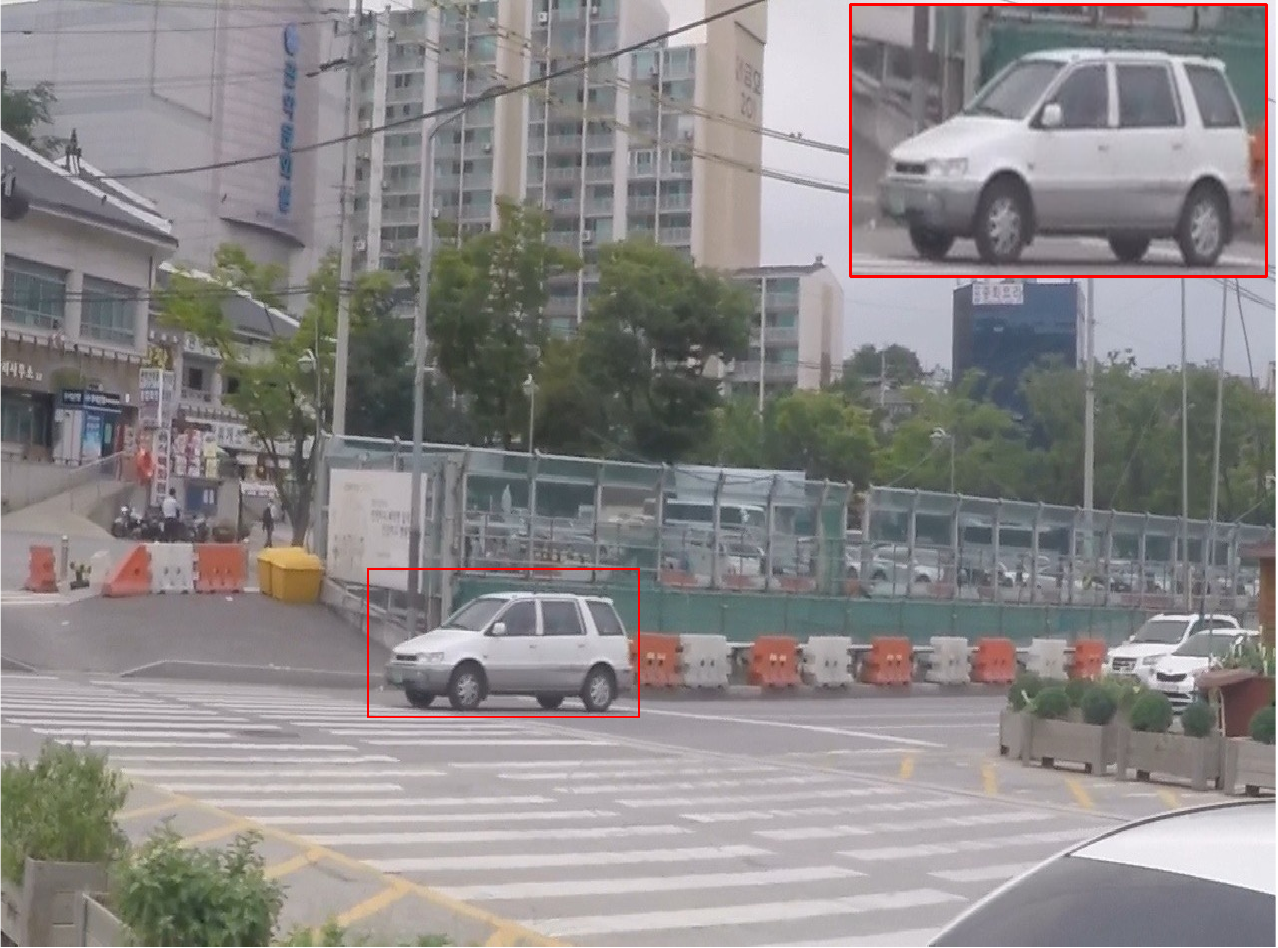} & \\
      \multicolumn{4}{c}{\vspace{-10pt}} \\
    \end{tabular}}
  \caption{\textbf{Enhancing SAM-based applications:} A qualitative demonstration of RobustSAM's superiority in refining SAM-based single image dehazing and deblurring.}
  \label{fig:qualitative_downstream}
\end{figure}


\begin{table}[t!]
\centering
\scalebox{0.72}{
\begin{tabular}{lcc}
\toprule
\multirow{2}{*}{\centering Module} & \multicolumn{2}{c}{Metric}  \\
\cmidrule(lr){2-3}
                              & IoU & PA \\
\midrule\midrule
\multicolumn{3}{c}{Baseline} \\
\hdashline
SAM                                           & 0.3056    & 0.8911   \\
SAM-Finetune                                  & 0.1871    & 0.7691   \\
SAM-Finetune Decoder                           & 0.2476    & 0.8691   \\
SAM-Finetune Output Token                      & 0.3194    & 0.9036   \\
\midrule
\multicolumn{3}{c}{RobustSAM} \\
\hdashline
w AMFG                                         & 0.3455    & 0.9059   \\
w AMFG-F                                       & 0.3535    & 0.9120   \\
w AMFG-F+AOTG                                  & 0.3651    & 0.9193   \\
\rowcolor{LightCyan}
w AMFG-F+AOTG+ROT (\textbf{ALL})                        & \textbf{0.3717}    & \textbf{0.9226}   \\
\bottomrule
\end{tabular}}
\caption{\textbf{Efficacy of Proposed Modules:} An evaluation on the BDD-100k~\cite{yu2018bdd100k} and LIS~\cite{Hong2021Crafting,2023lis} datasets reveals that each of the proposed modules enhances the performance of RobustSAM. (We use point prompts in this comparison.)}
\label{tab:ablation_ori}
\end{table}

\begin{table}[t!]
\centering
\scalebox{0.75}{
\begin{tabular}{ccccc}
\toprule
\multirow{2}{*}{Prior} & \multicolumn{2}{c}{Dehazing} & \multicolumn{2}{c}{Deblurring}  \\
\cmidrule(lr){2-3} \cmidrule(lr){4-5}
                       & PSNR & SSIM & PSNR & SSIM \\
\midrule
\midrule
SAM                    & 21.677 & 0.8451 & 27.491 & 0.9066 \\
\rowcolor{LightCyan}
\textbf{RobustSAM}     & \textbf{23.159} & \textbf{0.8685} & \textbf{29.351} & \textbf{0.9229} \\
\bottomrule
\end{tabular}}
\caption{\textbf{Quantitative evaluation for Dehazing~\cite{jin2023let} and Deblurring~\cite{li2023sam}  tasks using different priors.}}
\label{tab:downstream}
\end{table}
\subsection{Ablation Study}
To further understand the impact of our contributions, we conducted an ablation study. All experiments were performed on the BDD-100k~\cite{yu2018bdd100k} and LIS~\cite{Hong2021Crafting,2023lis} datasets.
\noindent \smallskip\\
\textbf{Fine-tune SAM?} 
We fine-tuned the SAM model in various configurations: fine-tuning the entire model, the decoder, and the output token. The results are presented in \tabref{tab:ablation_ori}. It was observed that fine-tuning the entire SAM model or its decoder drastically reduces its zero-shot capabilities, leading to a significant drop in performance. Fine-tuning only output token resulted in performance improvements; however, they were still inferior compared to RobustSAM.
\noindent \smallskip\\
\textbf{Effectiveness of Proposed Modules.} Furthermore, we validated the effectiveness of each proposed module, including the Anti-degradation Mask Feature Generation Module (AMFG), ADM with Fourier Degradation Suppression module (AMFG-F), Anti-degradation Output Token Generation (AOTG), and Robust Output Token (ROT). The findings presented in \tabref{tab:ablation_ori} show that each introduced module positively influences RobustSAM's overall performance, with the AMFG module demonstrating the most significant enhancement.

\subsection{Improving SAM-prior Tasks}
To validate whether our RobustSAM can enhance the performance of applications based on SAM priors under degraded image conditions, we selected single image dehazing~\cite{jin2023let} and single image deblurring~\cite{li2023sam} as test cases. Following the original papers' settings for these tasks, we utilized SAM and RobustSAM as their priors and evaluated their performance on SOTS dataset~\cite{li2017reside} for dehazing and GoPro dataset~\cite{nah2017deep} for deblurring. The findings, presented in~\tabref{tab:downstream} and \figref{fig:qualitative_downstream}, demonstrate that employing RobustSAM yields superior performance on downstream tasks. This enhancement can be attributed to RobustSAM's improved segmentation accuracy on degraded images, providing a more robust prior for these tasks.

\section{Conclusion}
This paper introduces RobustSAM, which excels in segmenting images under diverse degradations. The model's strength is rooted in its components—particularly the Anti-degradation Mask Feature Generation Module, Anti-degradation Output Token Generation, and Robust Output Token modules. To verify the effectiveness of RobustSAM, we proposed a large-scale dataset called Robust-Seg. Furthermore, we prove RobustSAM's superiority extends to improving SAM-based tasks such as dehazing and deblurring, confirming its value as a dependable tool for image processing under degraded conditions. Its performance sets a new standard for robustness in zero-shot segmentation, offering a promising direction for future research.

\appendix

\begin{center}
\textbf{\large Supplemental Materials}
\end{center}
\renewcommand{\theequation}{S.\arabic{equation}}
\renewcommand\thefigure{S.\arabic{figure}}
\renewcommand\thetable{S.\arabic{table}}
\setcounter{equation}{0}
\setcounter{figure}{0}
\setcounter{table}{0}

\section{More Experiments}
\subsection{Ablation Study}
We conducted comprehensive ablation studies on the BDD-100k~\cite{yu2018bdd100k} and LIS~\cite{Hong2021Crafting,2023lis} datasets, along with the subset of the COCO~\cite{lin2014microsoft} dataset incorporated within the Robust-Seg dataset collection. It is important to highlight that these datasets are unseen (zero-shot) and with real-world degradations in our evaluations. The results, as detailed in~\tabref{tab:ablation}, underscore that every component integrated into RobustSAM contributes positively to its overall performance. This consistent enhancement across various datasets demonstrates the robustness and adaptability of RobustSAM, particularly in zero-shot settings.

\subsection{Different Backbones in RobustSAM}
\begin{figure}[t!]
\centering \includegraphics[width=0.49\textwidth,page=1]{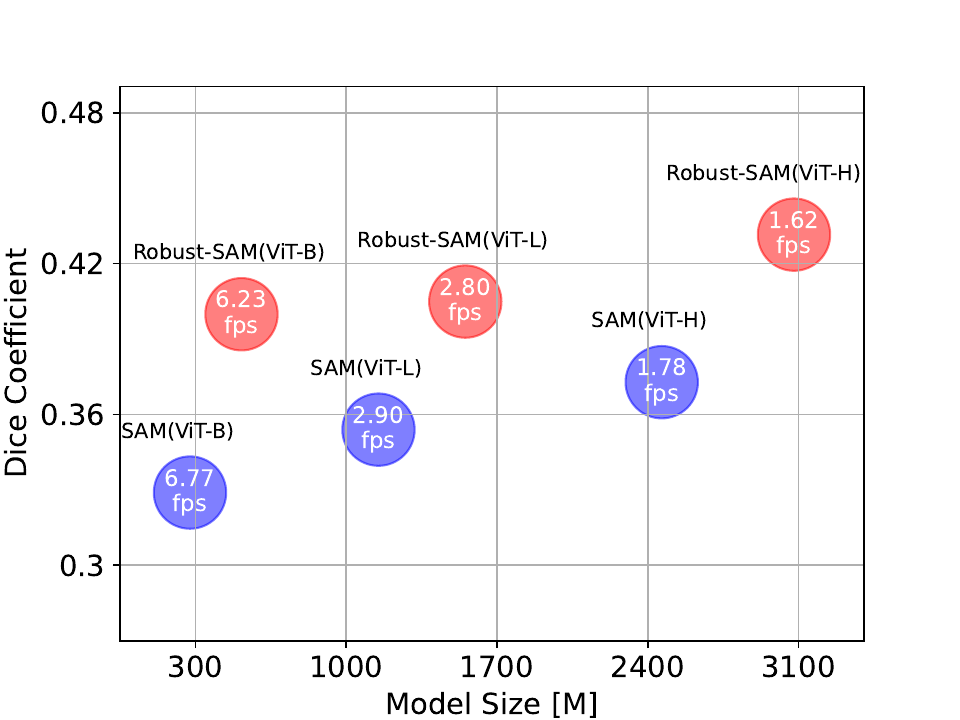}{}
\makeatother 
\caption{\textbf{Comparison of performance, speed, and model size among various SAM and RobustSAM variants.} The suffixes -B, -L, and -H correspond to ViT-B (Base), ViT-L (Large), and ViT-H (Huge) versions, respectively, representing different scales and complexities of the Vision Transformer architecture.}  
\label{fig:com_vit}
\end{figure}

In~\tabref{tab:backbone}, we showcase a thorough comparison of SAM and RobustSAM across various Vision Transformer (ViT)~\cite{dosovitskiy2020image} backbones, including ViT-B, ViT-L, and ViT-H. This comparison encompasses the numerical results on the combined BDD-100k~\cite{yu2018bdd100k} and LIS~\cite{Hong2021Crafting,2023lis} datasets and extends to the COCO~\cite{lin2014microsoft} dataset. The data clearly illustrate that RobustSAM consistently outperforms SAM across these diverse backbones. Furthermore, in \figref{fig:com_vit}, we provide a comprehensive comparison of performance, inference speed, and model size among various SAM and RobustSAM variants, offering additional insights into the efficiency and scalability of these models. Notably, our models effectively enhance performance in degraded scenarios with only a marginal increase in computational burden.

\begin{figure*}[t!]
  \centering
  \setlength{\tabcolsep}{2pt} %
  \begin{tabular}{cc}

    \includegraphics[width=0.45\linewidth]{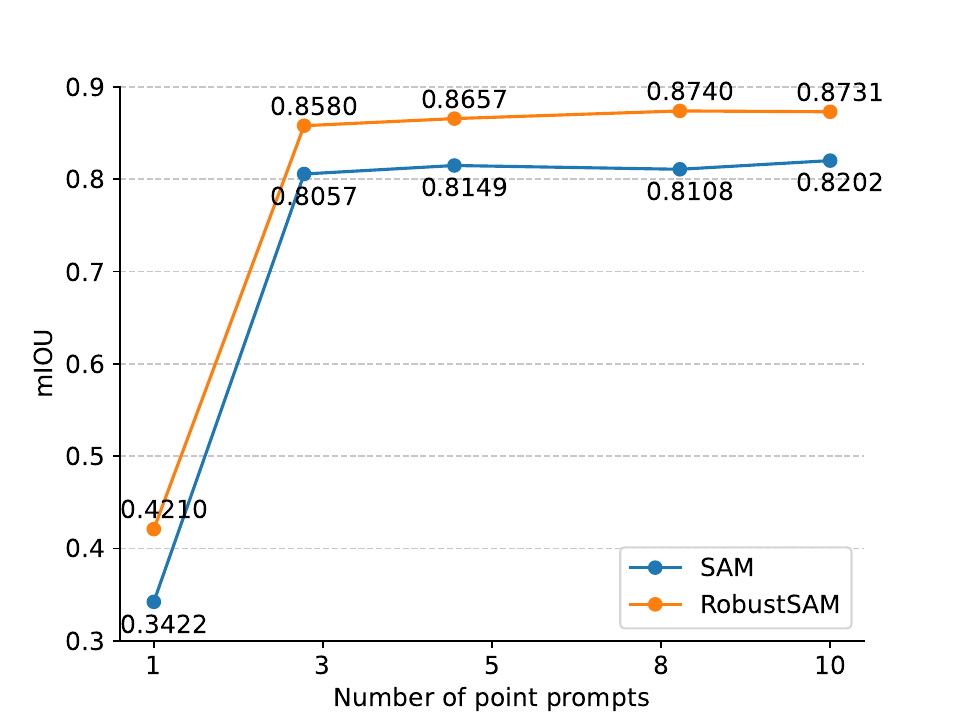} &
    \includegraphics[width=0.45\linewidth]{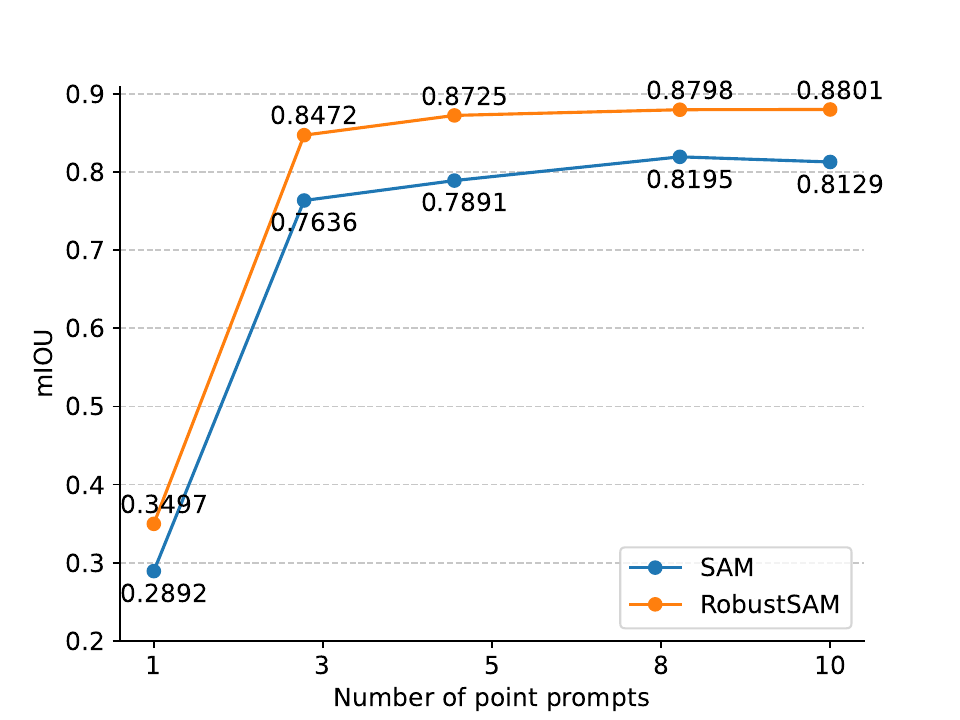} \\
    mIoU on BDD-100k dataset & mIoU on LIS dataset \\
  \end{tabular}
\caption{\textbf{Comparative analysis of interactive segmentation performance using different numbers of input points on the BDD-100k~\cite{yu2018bdd100k} and LIS~\cite{Hong2021Crafting,2023lis} datasets in a zero-shot setting.} RobustSAM consistently surpasses SAM across diverse point quantities, exhibiting a more pronounced enhancement, especially in scenarios with reduced prompt ambiguity on the BDD-100k dataset.}
\label{fig:varing_points}
\end{figure*}

\subsection{Comparison of Varying Number of Point Prompts}
To examine the interactive segmentation performance of RobustSAM using point prompts, we have conducted a comprehensive comparison in~\figref{fig:varing_points}. This comparison assesses RobustSAM against SAM with a range of input point numbers on the BDD-100k~\cite{yu2018bdd100k} and LIS~\cite{Hong2021Crafting,2023lis} datasets in a zero-shot learning context. RobustSAM consistently achieves superior performance throughout these datasets compared to SAM, irrespective of the number of input points used.

\begin{figure}[t!]
  \centering
  \setlength{\tabcolsep}{2pt} %
  {\footnotesize %
  \begin{tabular}{ccc}
    Input  & \begin{tabular}{@{}c@{}}Attn. Map of \\ SAM’s Output Token\end{tabular} & \begin{tabular}{@{}c@{}}Attn. Map of \\RobustSAM’s \\ Robust Output Token\end{tabular} \\
    \includegraphics[width=0.3\columnwidth]{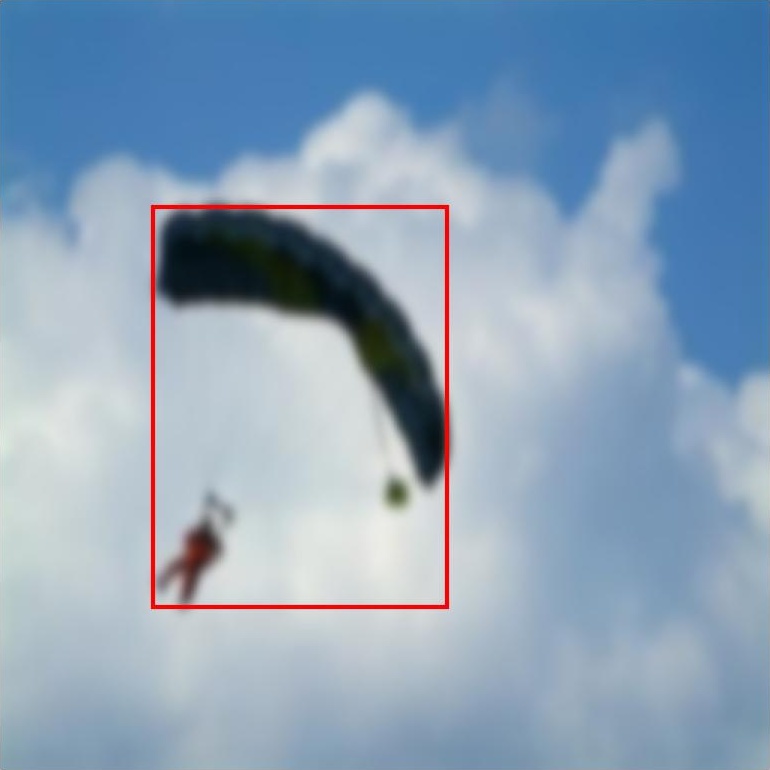} &
    \includegraphics[width=0.3\columnwidth]{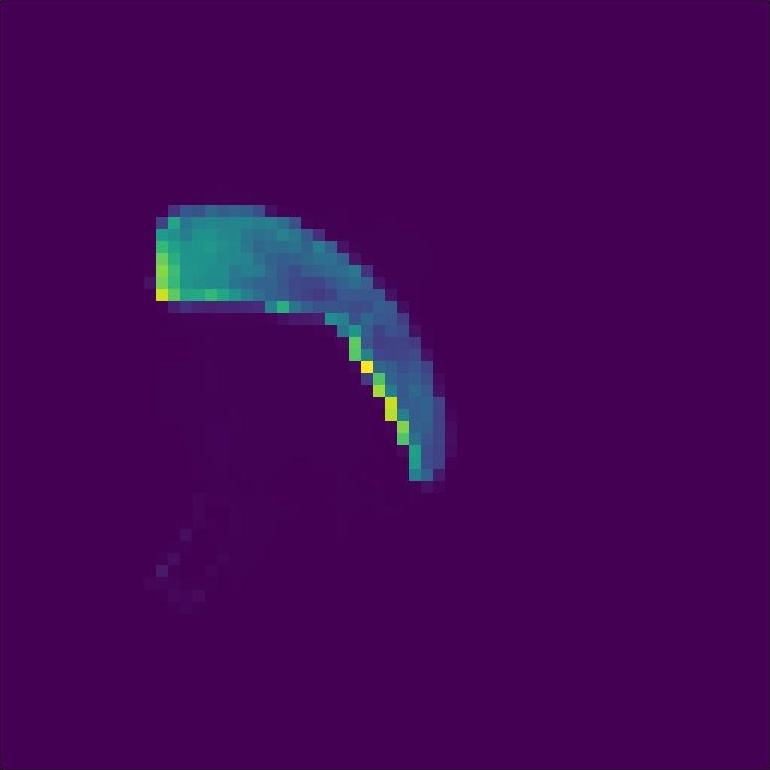} &
    \includegraphics[width=0.3\columnwidth]{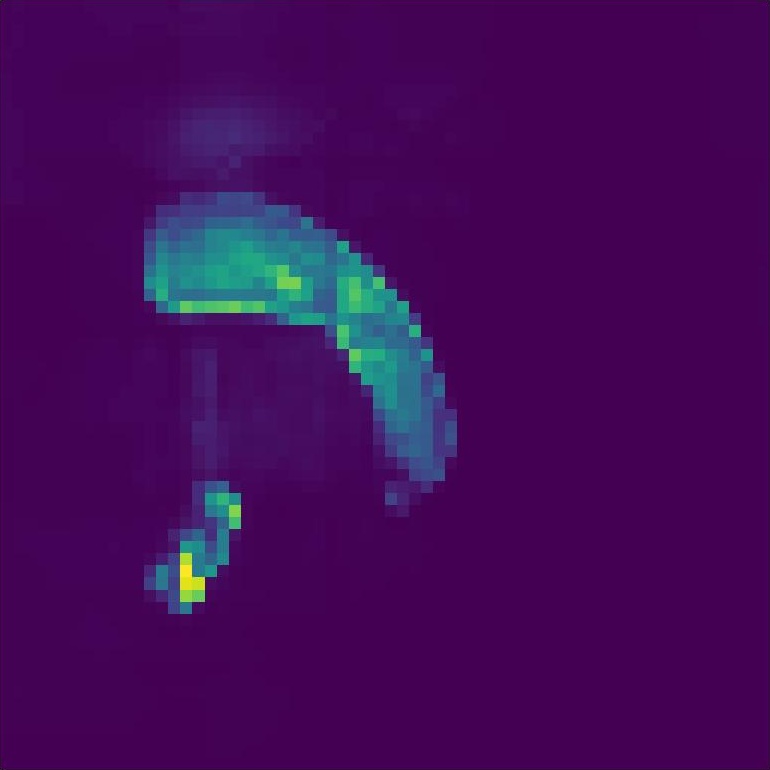} \\        
    \includegraphics[width=0.3\columnwidth]{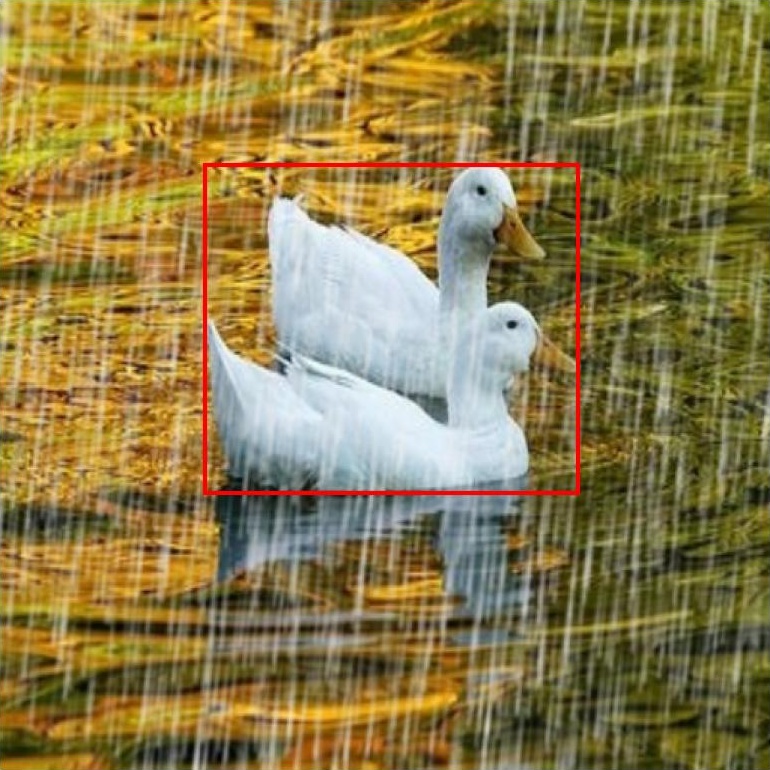} &
    \includegraphics[width=0.3\columnwidth]{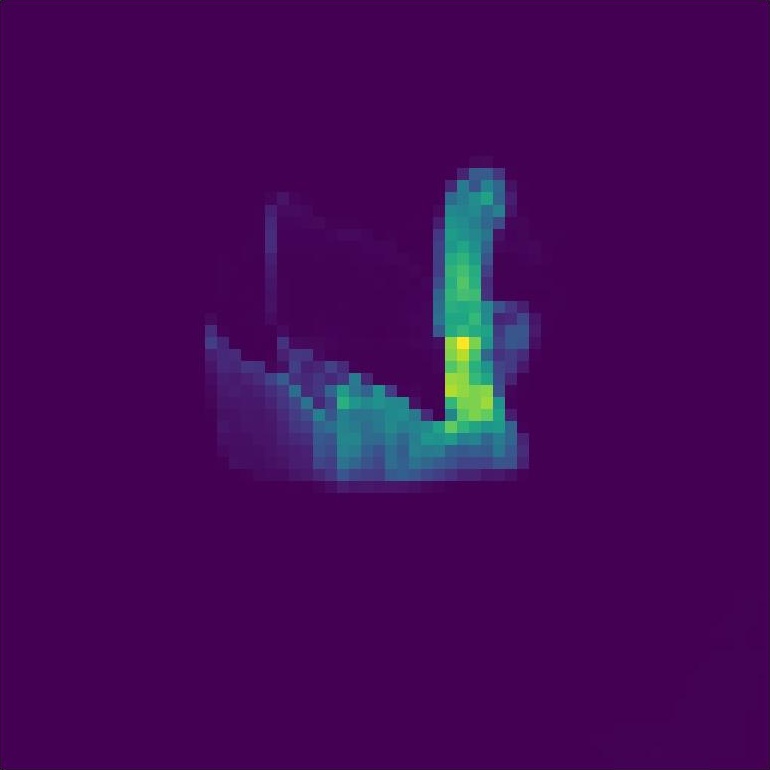} &
    \includegraphics[width=0.3\columnwidth]{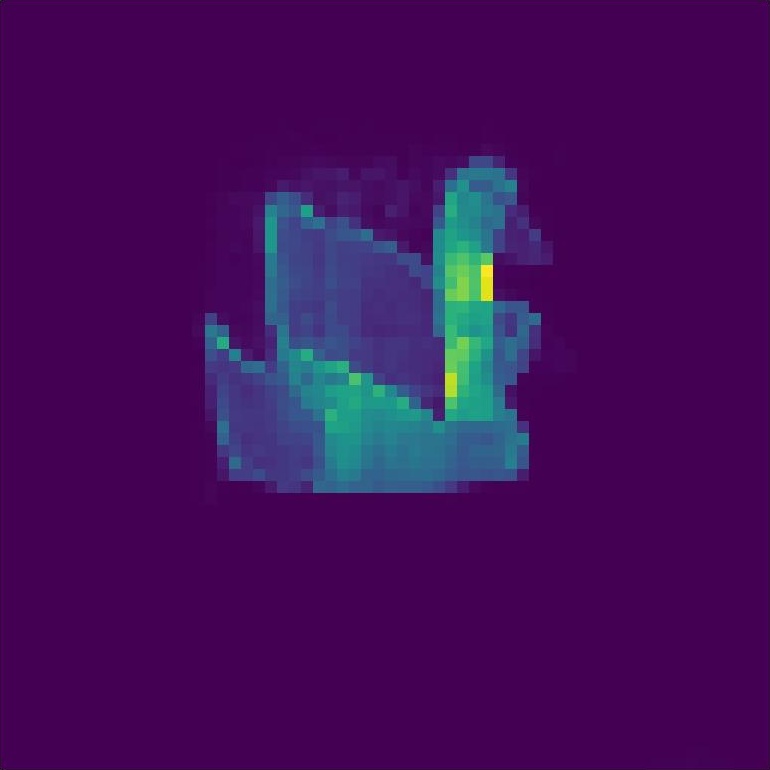} \\    
    \includegraphics[width=0.3\columnwidth]{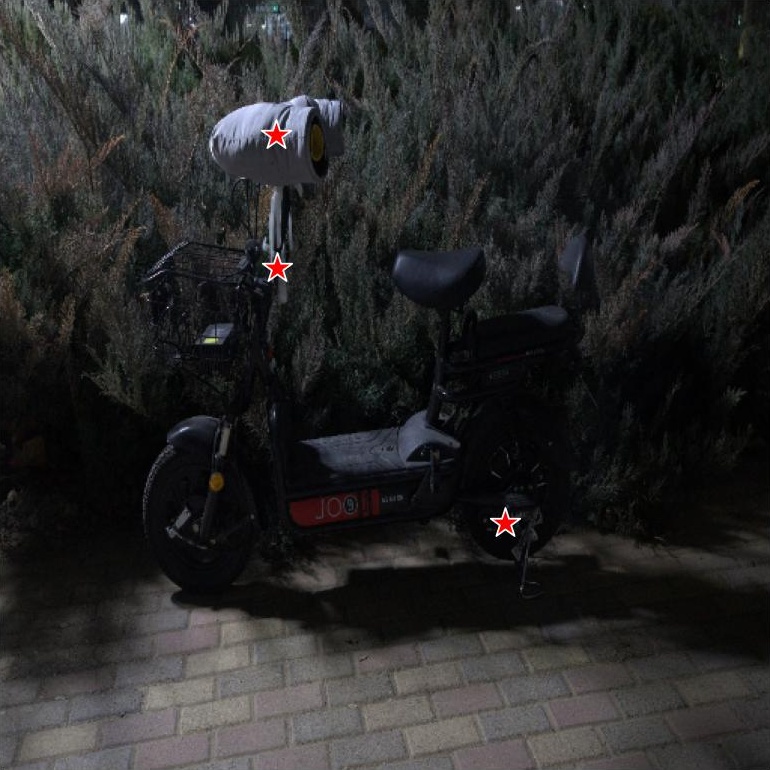} &
    \includegraphics[width=0.3\columnwidth]{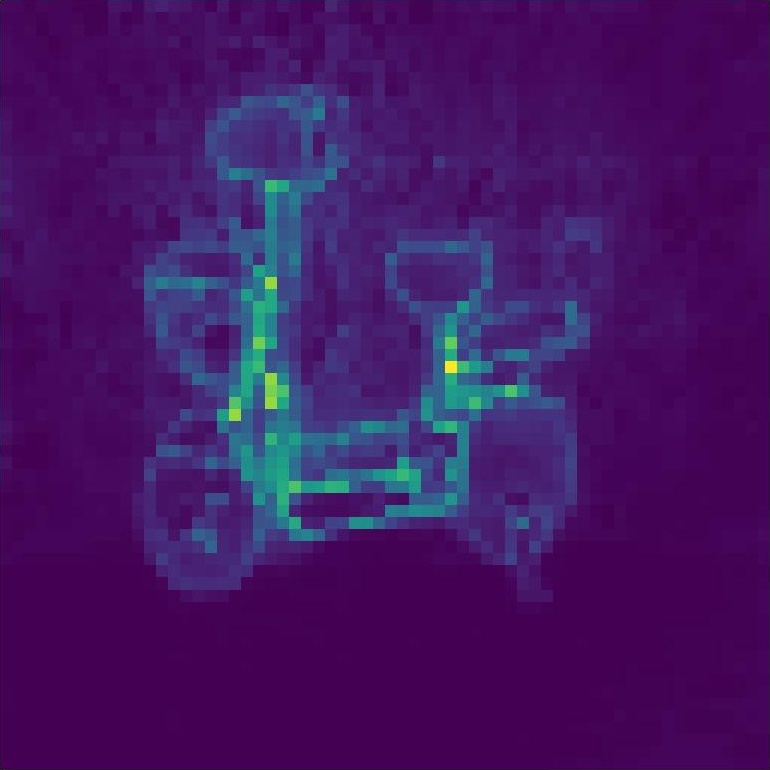} &
    \includegraphics[width=0.3\columnwidth]{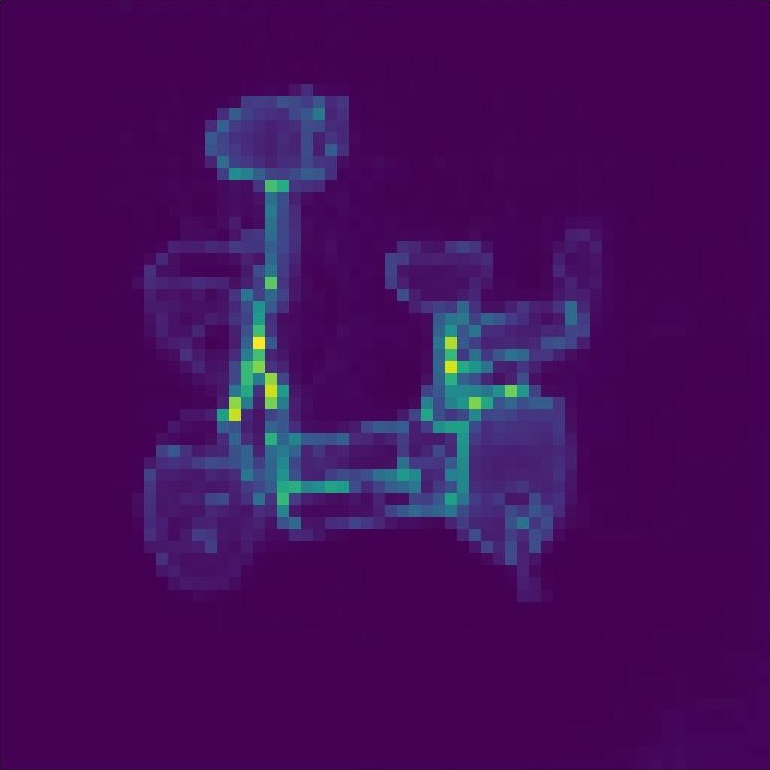} \\    
  \end{tabular}
  }
\caption{\textbf{Comparison of cross-attention in the final decoder layer between SAM's original output token and the enhanced Robust Output Token.} The Robust Output Token distinctly exhibits a precise focus on accurately identifying object boundaries and contents. It further demonstrates attention to the boundary and thin structural regions, which are typically overlooked by the original output token.}
  \label{fig:token}
\end{figure}

\begin{table*}[t!]
\centering
\scalebox{0.82}{
\begin{tabular}{lccccccc}
\toprule
\multirow{2}{*}{Module} & \multicolumn{2}{c}{BDD-100k+LIS} & \multicolumn{4}{c}{COCO} \\
\cmidrule(lr){2-3} \cmidrule(lr){4-7}
                              & IoU & PA & AP & $\text{AP}_\text{S}$ & $\text{AP}_\text{M}$ & $\text{AP}_\text{L}$ \\
\midrule\midrule
\multicolumn{7}{c}{Baseline} \\
\hdashline
SAM                                           & 0.3056    & 0.8911   & 0.5002 & 0.3168 & 0.4292 & 0.5243 \\
SAM-Finetune                                  & 0.1871    & 0.7691   & 0.1321    & 0.0384    & 0.1211    & 0.1731    \\
SAM-Finetune Decoder                          & 0.2476    & 0.8691   & 0.1457 & 0.0391 & 0.1322 & 0.1868 \\
SAM-Finetune Output Token                     & 0.3194    & 0.9036   & 0.4853 & 0.3011 & 0.4312 & 0.5266 \\
\midrule
\multicolumn{7}{c}{RobustSAM} \\
\hdashline
w AMFG                                        & 0.3455    & 0.9059   & 0.5021 & 0.3147 & 0.4295 & 0.5273 \\
w AMFG-F                                      & 0.3535    & 0.9120   & 0.5045 & 0.3150 & 0.4336 & 0.5370 \\
w AMFG-F+AOTG                                 & 0.3651    & 0.9193   & 0.5075 & 0.3161 & 0.4349 & 0.5381 \\
\rowcolor{LightCyan}
w AMFG-F+AOTG+ROT (\textbf{ALL})               & \textbf{0.3717}    & \textbf{0.9226}   & \textbf{0.5130} & \textbf{0.3192} & \textbf{0.4416} & \textbf{0.5518} \\
\bottomrule
\end{tabular}}
\caption{\textbf{Efficacy of Proposed Modules:} An evaluation of the BDD-100k~\cite{yu2018bdd100k}, LIS~\cite{Hong2021Crafting,2023lis}, and COCO~\cite{lin2014microsoft} datasets reveals that each of the proposed modules enhances the performance of RobustSAM. (We use point prompts for BDD-100k+LIS and bounding box prompts for COCO in this comparison.)}
\label{tab:ablation}
\end{table*}

\begin{table*}[t!]
\centering
\scalebox{0.85}{
\begin{tabular}{lcccccc}
\toprule
\multirow{2}{*}{Model}          & \multicolumn{2}{c}{BDD-100k+LIS} & \multicolumn{4}{c}{COCO} \\
\cmidrule(lr){2-3} \cmidrule(lr){4-7}
               & IoU   & PA    & AP & $\text{AP}_\text{S}$ & $\text{AP}_\text{M}$ & $\text{AP}_\text{L}$  \\
\midrule\midrule
SAM-B          & 0.3003 & 0.8826 & 0.4589 & 0.2958 & 0.3840 & 0.4752 \\
\rowcolor{LightCyan}
RobustSAM-B    & \textbf{0.3317} & \textbf{0.8972} & \textbf{0.4710} & \textbf{0.2961} & \textbf{0.4175} & \textbf{0.5268} \\
SAM-L          & 0.3056 & 0.8911 & 0.5002 & 0.3168 & 0.4292 & 0.5243 \\
\rowcolor{LightCyan}
RobustSAM-L    & \textbf{0.3717} & \textbf{0.9226} & \textbf{0.5130} & \textbf{0.3192} & \textbf{0.4416} & \textbf{0.5518} \\
SAM-H          & 0.3384 & 0.9305 & 0.5087 & 0.3184 & 0.4430 & 0.5255 \\
\rowcolor{LightCyan}
RobustSAM-H    & \textbf{0.3813} & \textbf{0.9367} & \textbf{0.5167} & \textbf{0.3188} & \textbf{0.4455} & \textbf{0.5697} \\
\bottomrule
\end{tabular}}
\caption{\textbf{Performance comparison between SAM and RobustSAM across different Vision Transformer (ViT) backbones.}}
\label{tab:backbone}

\end{table*}

\subsection{Token Visualization}
In~\figref{fig:token}, we provide an illustrative comparison of cross-attention in the last token-to-image layer of the mask decoder between SAM's original output token and the enhanced Robust Output Token of RobustSAM. This comparison underscores the ability of the Robust Output Token to achieve more focused and precise attention. It excels in accurately identifying object boundaries and contents, an aspect often overlooked by SAM's original output token in degraded scenes. This focused attention is particularly evident in its handling of the object's boundaries, demonstrating RobustSAM's enhanced capability to discern and highlight details and structures, crucial for effective segmentation in challenging imaging conditions.

\subsection{Visualization of Feature Representation}
\begin{figure}[t!]
  \centering\vspace{-2ex}
  \setlength{\tabcolsep}{0.1pt}\scriptsize
    \includegraphics[width=1\linewidth]{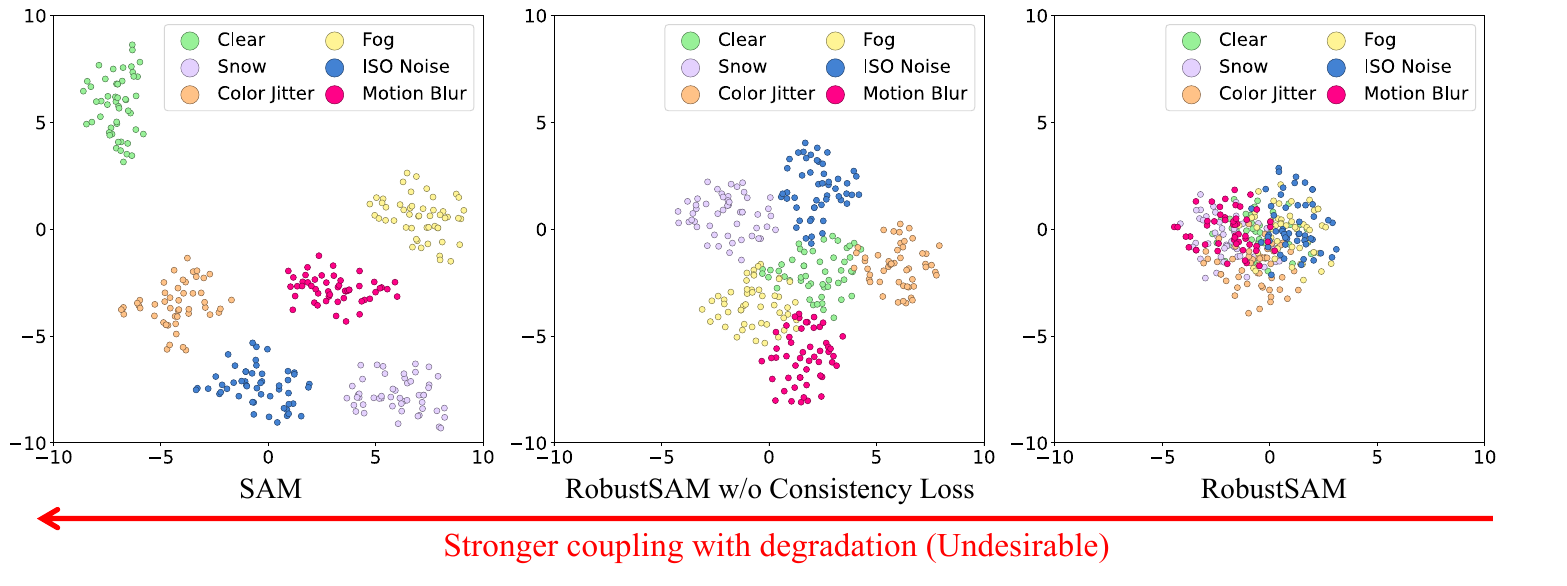}
    \caption{\textbf{Visualization of feature representation based on different baselines.}}
    \label{fig:tsne_sam}

\end{figure}

We conducted an experiment by randomly sampling 50 images for each of the six degradations from our dataset. We ran SAM, RobustSAM w/o consistency loss and RobustSAM to extract the mask features and performed t-SNE analysis. The results shown in \figref{fig:tsne_sam} indicate that in the original SAM, features from the same degradation type tended to cluster together. However, with RobustSAM's feature suppression mechanism, features from different degradations significantly overlap each other, suggesting the minimal influence of degradation on feature extraction. Moreover, when consistency loss is not used, the clustering due to same degradation is more evident compared to RobustSAM, demonstrating the effectiveness of the consistency loss. 

\begin{table}[t!]
\centering
\scalebox{0.8}{
\begin{tabular}{ccc} 
\toprule
SA-1B     & SAM  & RobustSAM  \\ 
\hline
IoU/PA &     0.6917/0.8719
     &   \textbf{0.7065}/\textbf{0.8726} \\
\bottomrule
\end{tabular}}
\caption{\textbf{Zero-shot segmentation comparison on a subset of SA-1B dataset~\cite{kirillov2023segany}.}}
\label{tab:sa1b}
\end{table}

\begin{table}[t!]
\centering
\scalebox{0.85}{
\begin{tabular}{cccccc}
\toprule
\multirow{2}{*}{Method} & \multicolumn{2}{c}{Point} & \multicolumn{2}{c}{Bounding Box}  \\ 
\cmidrule(lr){2-3} \cmidrule(lr){4-5}
                        & IoU & Dice & IoU & Dice \\
\midrule
\midrule

SAM                 & 0.3056 & 0.3837 & 0.8808 & 0.9171 \\
HQ-SAM              & 0.2943 & 0.3712 & 0.8877 & 0.9245 \\
HQ-SAM-F              & 0.2951 & 0.3720 & 0.8906 & \underline{0.9406} \\
AirNet+SAM          & 0.3245 & 0.4550 & 0.8776 & 0.9129 \\
AirNet-F+SAM          & 0.3251 & \underline{0.4583} & 0.8781 & 0.9133 \\
MPR-Net+SAM         & 0.3271 & 0.4079 & 0.8758 & 0.9174 \\
SwinIR+SAM          & 0.3294 & 0.4117 & 0.8677 & 0.9241 \\
SwinIR-F+SAM          & \underline{0.3313} & 0.4120 & 0.8683 & 0.9253 \\
MW-Net+SAM          & 0.3195 & 0.4000 &0.8898 & 0.9379 \\
MW-Net-F+SAM          & 0.3199 & 0.4006 & \underline{0.8908} & 0.9387 \\
URIE+SAM            & 0.3042 & 0.3828 & 0.8799 & 0.9165 \\
\rowcolor{LightCyan}
\textbf{RobustSAM}  & \textbf{0.3717} & \textbf{0.8926} & \textbf{0.8958} & \textbf{0.9416} \\
\bottomrule
\end{tabular}}
\caption{\textbf{Zero-Shot Segmentation Comparison:} This figure presents a comparison of segmentation performance on the entire BDD-100k~\cite{yu2018bdd100k} and LIS~\cite{Hong2021Crafting,2023lis} datasets, which are unseen datasets with real-world degradations. We utilized point and bounding box prompts for segmentation. The notation `+SAM'' indicates the process of first restoring the image, followed by applying SAM for segmentation, consistent with the methodology described in Tables \ref{tab:eval_rest} and \ref{tab:eval_coco}.}
\label{tab:eval_real}
\end{table}

\begin{table}[t!]
\centering
\scalebox{0.74}{
\begin{tabular}{ccccccc}
\toprule
\multirow{2}{*}{Method} & \multicolumn{2}{c}{Degrade} & \multicolumn{2}{c}{Clear} & \multicolumn{2}{c}{Average}  \\ 
\cmidrule(lr){2-3} \cmidrule(lr){4-5} \cmidrule(lr){6-7}
                        & IoU & PA & IoU & PA & IoU & PA \\
\midrule

SAM                 & 0.7981 & 0.9555  & 0.8295  & 0.9707  & 0.8000  & 0.9565  \\
HQ-SAM                 & 0.8079  & 0.9617  & 0.8448  & 0.9756  & 0.8102  & 0.9625 \\
HQ-SAM-F                 & \underline{0.8082}  & 0.9620  & \underline{0.8452}  & \underline{0.9760}  & \underline{0.8106}  & 0.9630 \\
AirNet+SAM                 & 0.7988 & 0.9629  & 0.8312 & 0.9752  & 0.8008  & 0.9637 \\
AirNet-F+SAM                 & 0.7992 & \underline{0.9635}  & 0.8316 & \underline{0.9760}  & 0.8083  & \underline{0.9640} \\
MPR-Net+SAM                 & 0.7969  & 0.9585  & 0.8227 & 0.9712  & 0.7985  & 0.9593  \\
SwinIR+SAM                 & 0.7951  & 0.9543  & 0.8210 & 0.9701  & 0.7971  & 0.9580  \\
SwinIR-F+SAM                 & 0.7956  & 0.9553  & 0.8255 & 0.9713  & 0.7983  & 0.9592  \\
MW-Net+SAM                 & 0.7713  & 0.9432  & 0.8183 & 0.9692  & 0.7813  & 0.9491  \\
MW-Net-F+SAM                & 0.7722  & 0.9440  & 0.8192 & 0.9701  & 0.7830  & 0.9501  \\
URIE+SAM                 & 0.7904  & 0.9593  & 0.8288  & 0.9740  & 0.7928  & 0.9602  \\
\rowcolor{LightCyan}
\textbf{RobustSAM}                 & \textbf{0.8195}  & \textbf{0.9778}  & \textbf{0.8529}  & \textbf{0.9817} & \textbf{0.8216}  & \textbf{0.9780}  \\
\bottomrule
\end{tabular}}
\caption{\textbf{Zero-shot segmentation comparison on the whole NDD20~\cite{trotter2020ndd20}, STREETS~\cite{snyder2019streets}, and FSS-1000~\cite{FSS1000} (unseen datasets with synthetic degradations) in Robust-Seg dataset using point prompts.}}
\label{tab:eval_rest}
\end{table}

\begin{table}[t!]
\centering
\scalebox{0.8}{
\begin{tabular}{ccccc}
\toprule
\multirow{2}{*}{Method} & \multicolumn{4}{c}{Performance Metrics} \\
\cmidrule(lr){2-5}
                        & AP & $\text{AP}_\text{S}$ & $\text{AP}_\text{M}$ & $\text{AP}_\text{L}$ \\
\midrule
SAM                    & 0.5002 & 0.3168 & 0.4292 & 0.5243 \\
HQ-SAM                 & 0.5052 & 0.2920 & 0.4267 & \underline{0.5517} \\
HQ-SAM-F                 & \underline{0.5063} & 0.2925 & 0.4272 & \textbf{0.5518} \\
AirNet+SAM             & 0.4926 & 0.3068 & 0.4263 & 0.5187 \\
AirNet-F+SAM             & 0.4933 & 0.3075 & 0.4272 & 0.5203 \\
MPR-Net+SAM            & 0.4986 & 0.3133 & 0.4301 & 0.5227 \\
SwinIR+SAM            & 0.4911 & 0.3027 & 0.4211 & 0.5195 \\
SwinIR-F+SAM            & 0.4923 & 0.3038 & 0.4219 & 0.5201 \\
MW-Net+SAM            & 0.5027 & 0.3161 & 0.4354 & 0.5290 \\
MW-Net-F+SAM            & 0.5033 & 0.3165 & \underline{0.4362} & 0.5294 \\
URIE+SAM               & 0.4980 & \underline{0.3186} & 0.4319 & 0.5215 \\
\rowcolor{LightCyan}
\textbf{RobustSAM}     & \textbf{0.5130} & \textbf{0.3192} & \textbf{0.4416} & \textbf{0.5518} \\
\bottomrule
\end{tabular}}
\caption{\textbf{Zero-shot segmentation comparison on the whole COCO~\cite{lin2014microsoft} (unseen datasets with synthetic degradations) in Robust-Seg dataset using Bounding Box prompts.}}
\label{tab:eval_coco}
\end{table}

\begin{figure*}[t!]
  \centering
  \setlength{\tabcolsep}{1pt} 
  \begin{tabular}{cccccc}
    Input & SAM & HQ-SAM & AirNet+SAM & URIE+SAM & RobustSAM \\

    \includegraphics[width=.16\textwidth, height=2.cm]{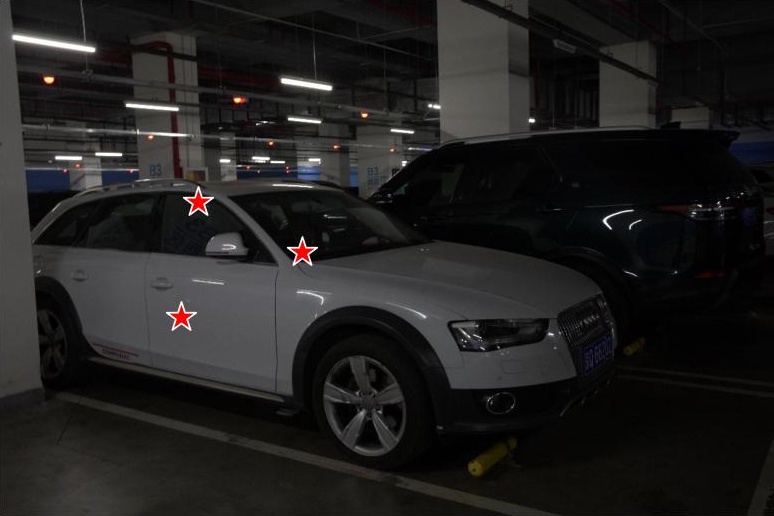} & 
    \includegraphics[width=.16\textwidth, height=2.cm]{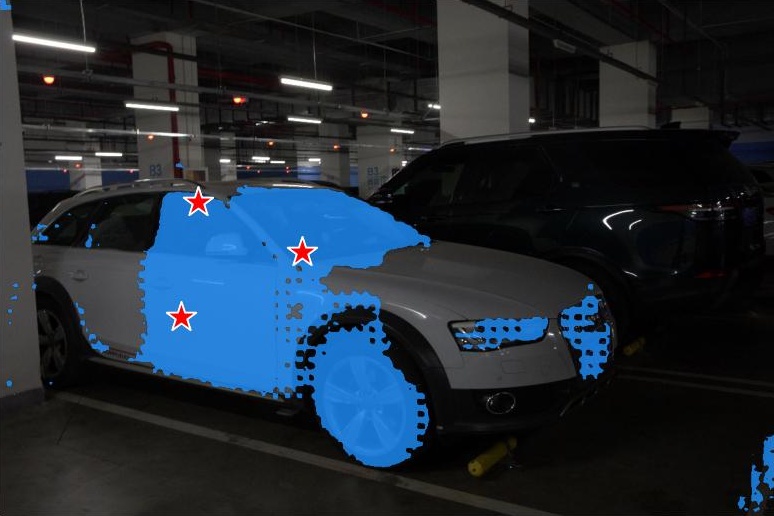} & 
    \includegraphics[width=.16\textwidth, height=2.cm]{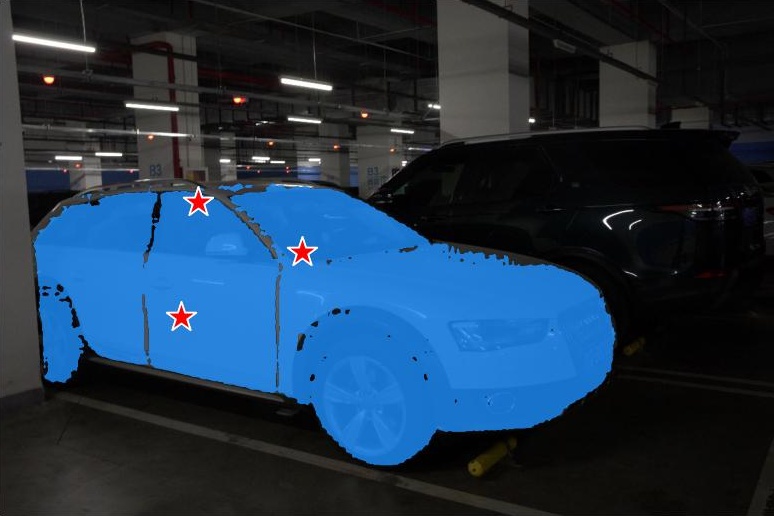} & 
    \includegraphics[width=.16\textwidth, height=2.cm]{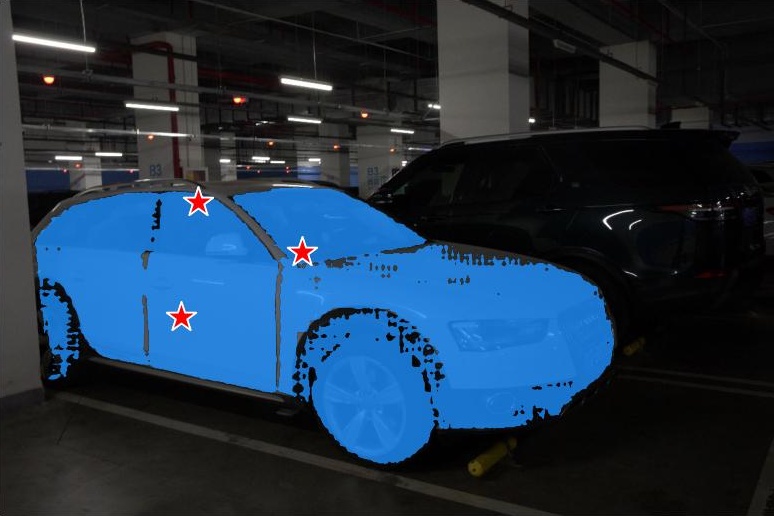} &
    \includegraphics[width=.16\textwidth, height=2.cm]{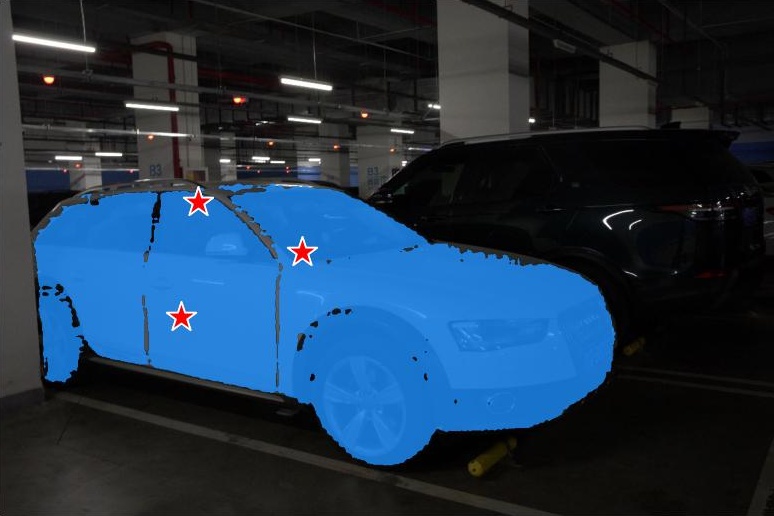} & 
    \includegraphics[width=.16\textwidth, height=2.cm]{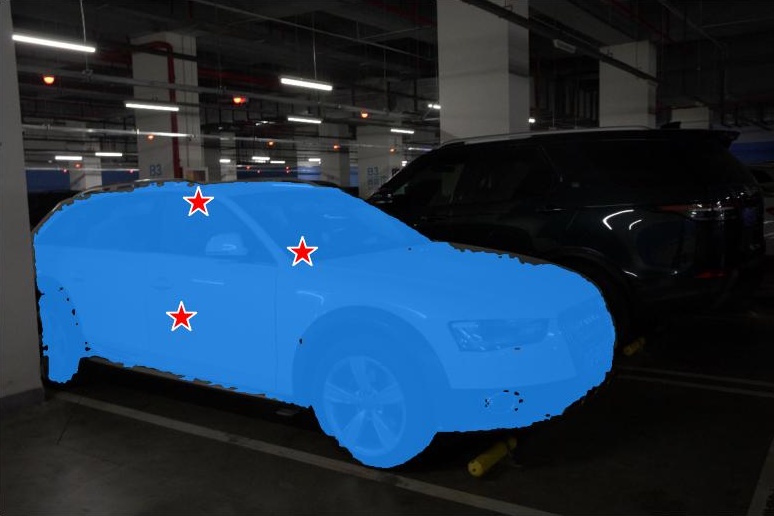} \\
        \multicolumn{6}{c}{\vspace{-14pt}} \\
    \includegraphics[width=.16\textwidth, height=2.cm]{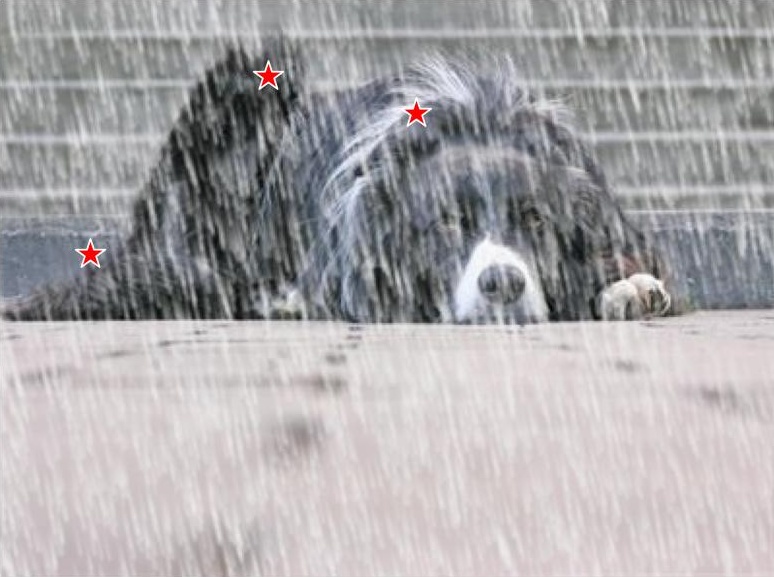} &
    \includegraphics[width=.16\textwidth, height=2.cm]{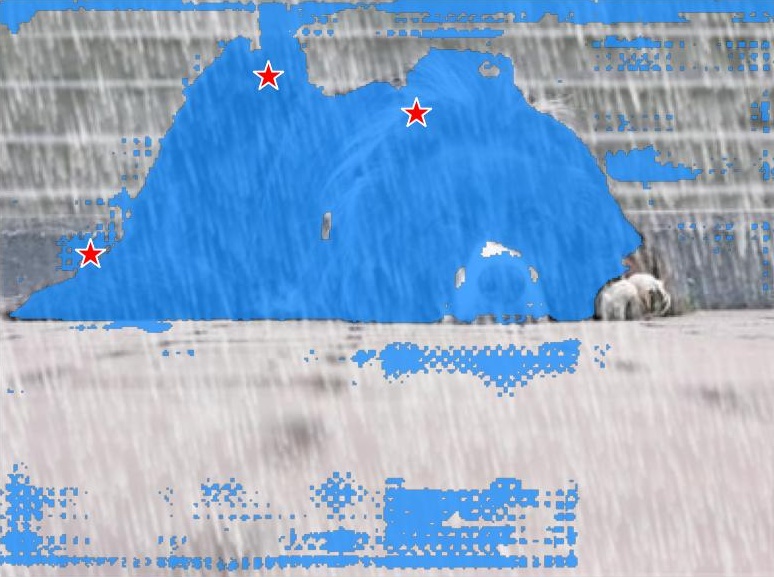} &
    \includegraphics[width=.16\textwidth, height=2.cm]{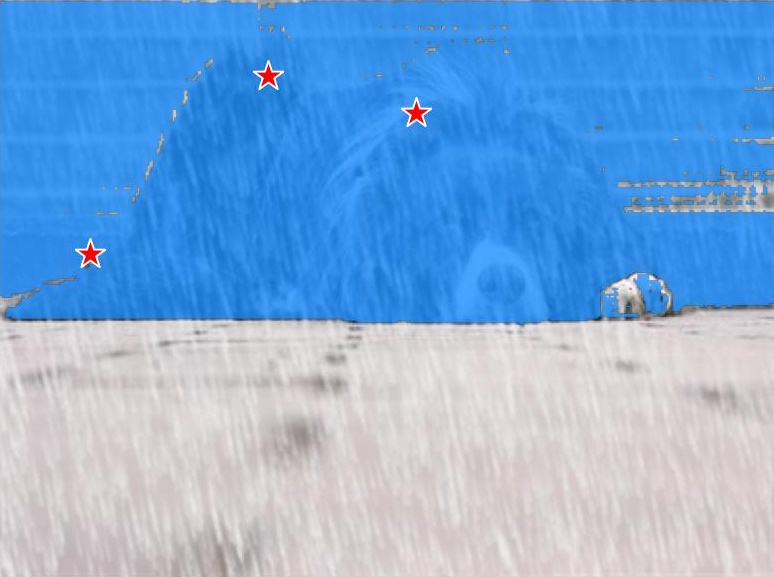} & 
    \includegraphics[width=.16\textwidth, height=2.cm]{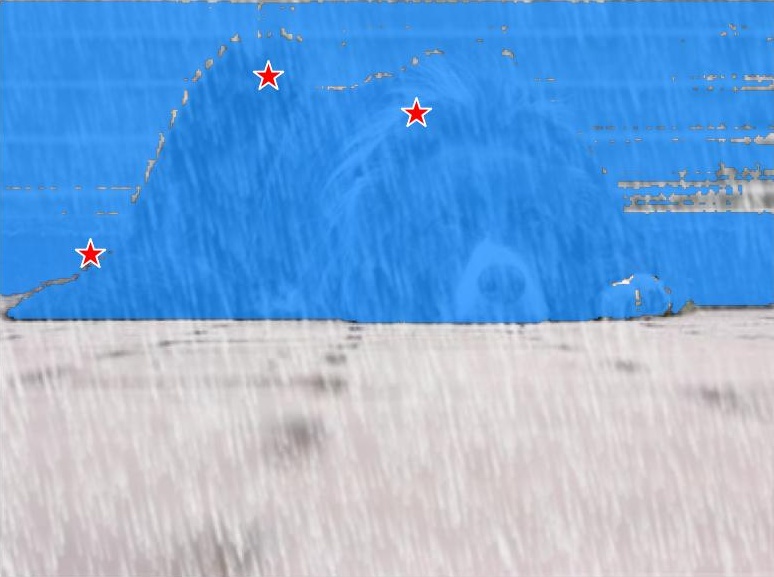} &
    \includegraphics[width=.16\textwidth, height=2.cm]{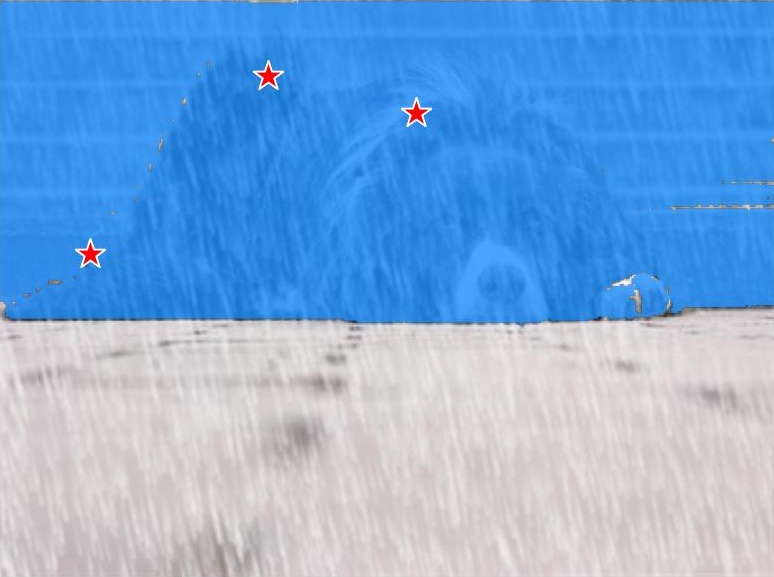} & 
    \includegraphics[width=.16\textwidth, height=2.cm]{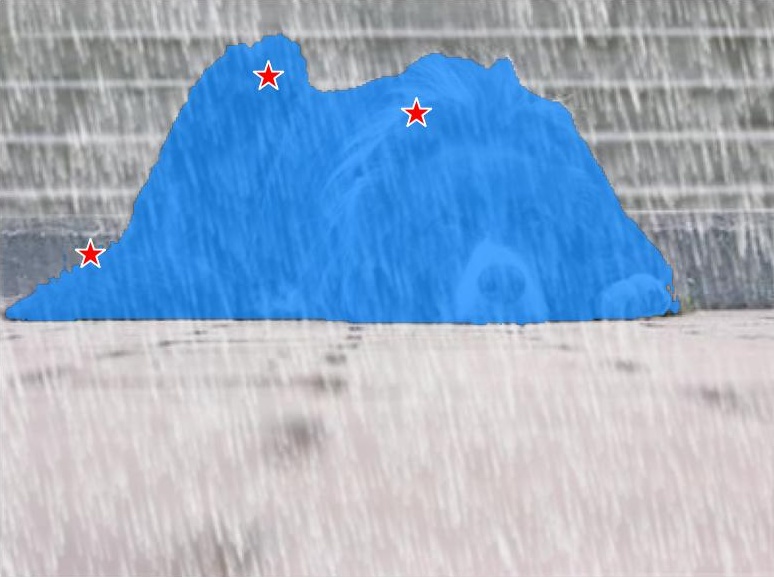} 
\\    
    \multicolumn{6}{c}{\vspace{-14pt}} \\

    \includegraphics[width=.16\textwidth, height=2.cm]{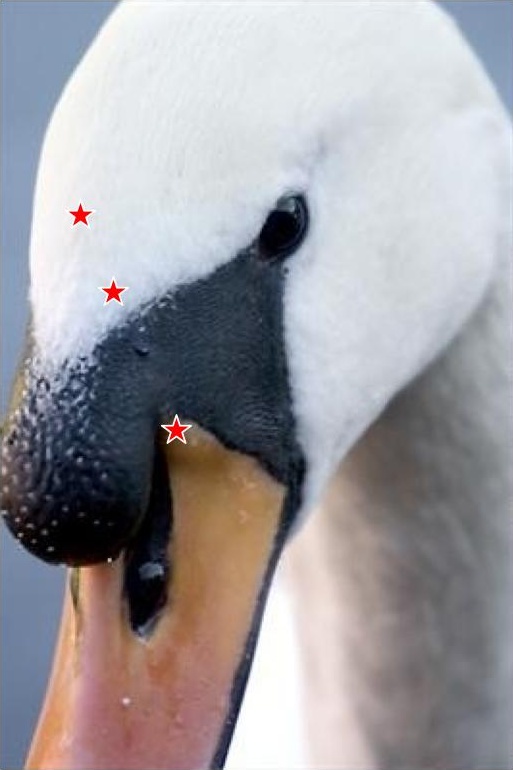} &
    \includegraphics[width=.16\textwidth, height=2.cm]{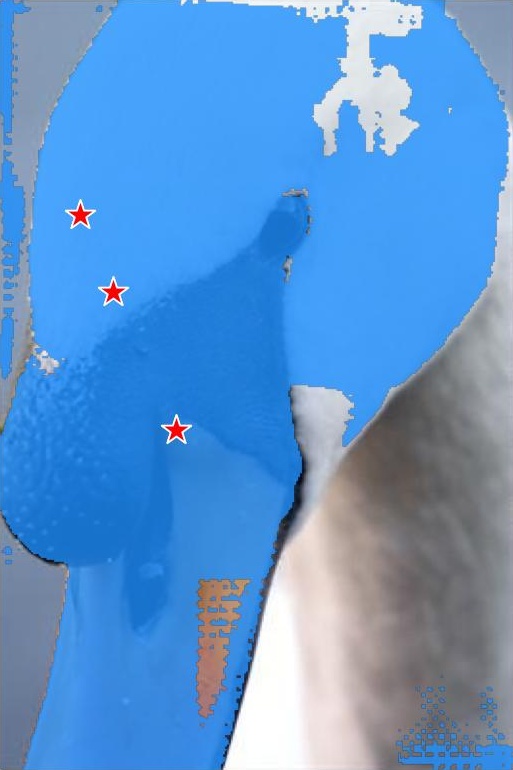} &
    \includegraphics[width=.16\textwidth, height=2.cm]{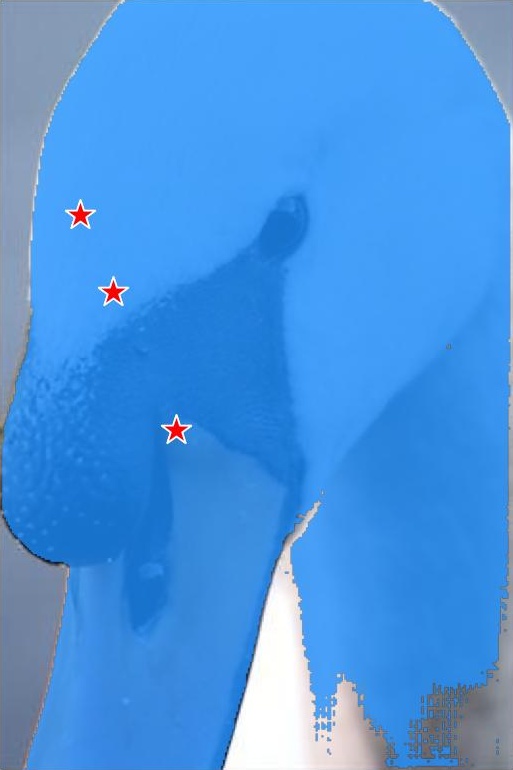} & 
    \includegraphics[width=.16\textwidth, height=2.cm]{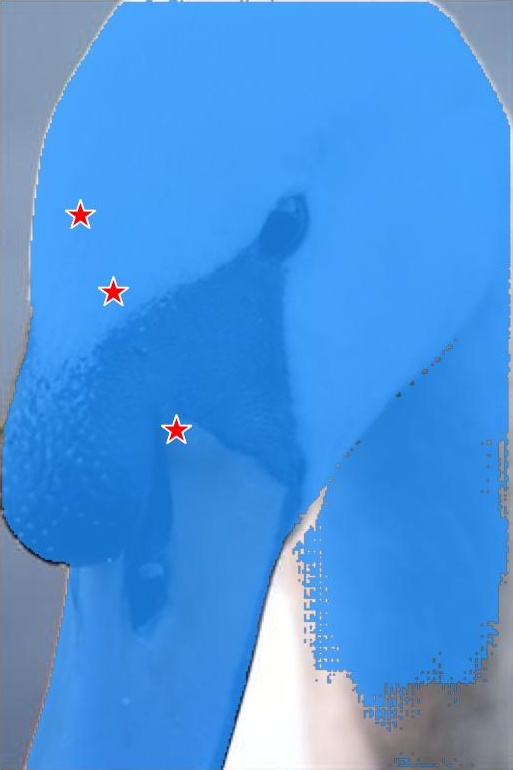} &
    \includegraphics[width=.16\textwidth, height=2.cm]{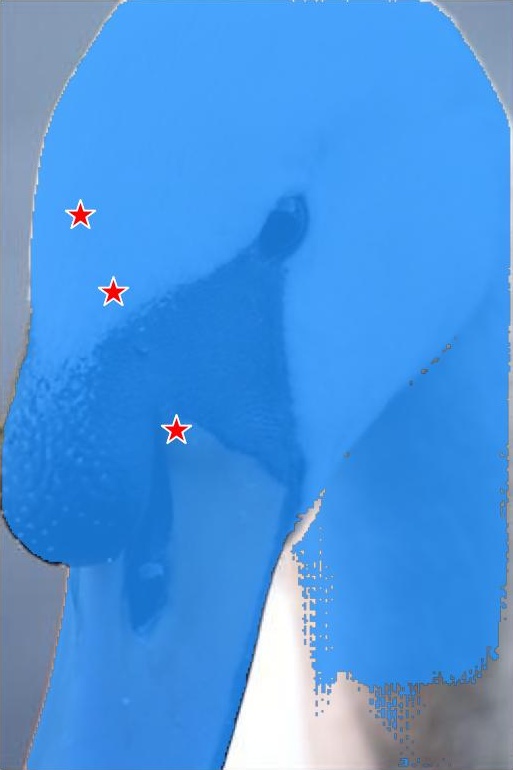} & 
    \includegraphics[width=.16\textwidth, height=2.cm]{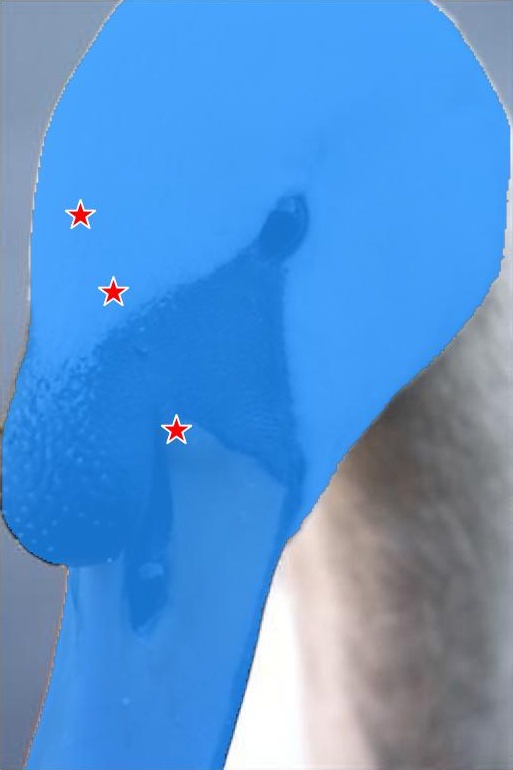} 
\\    

    \multicolumn{6}{c}{\vspace{-14pt}} \\
    \includegraphics[width=.16\textwidth, height=2.7cm]{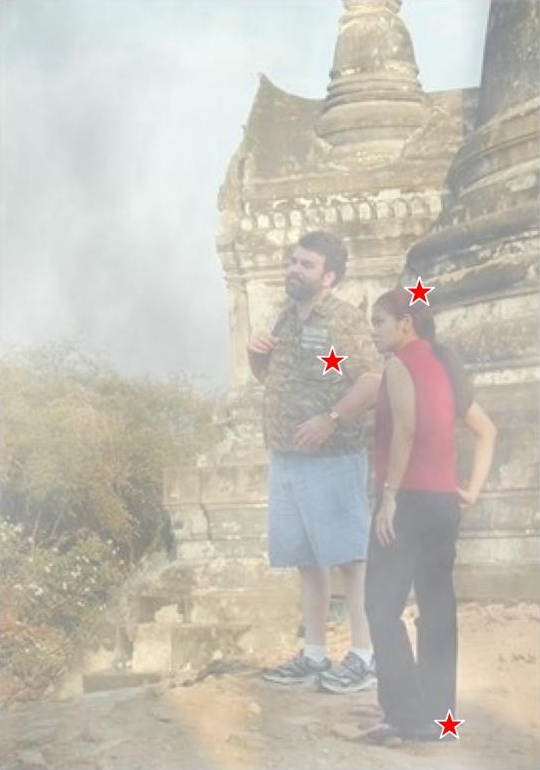} &
    \includegraphics[width=.16\textwidth, height=2.7cm]{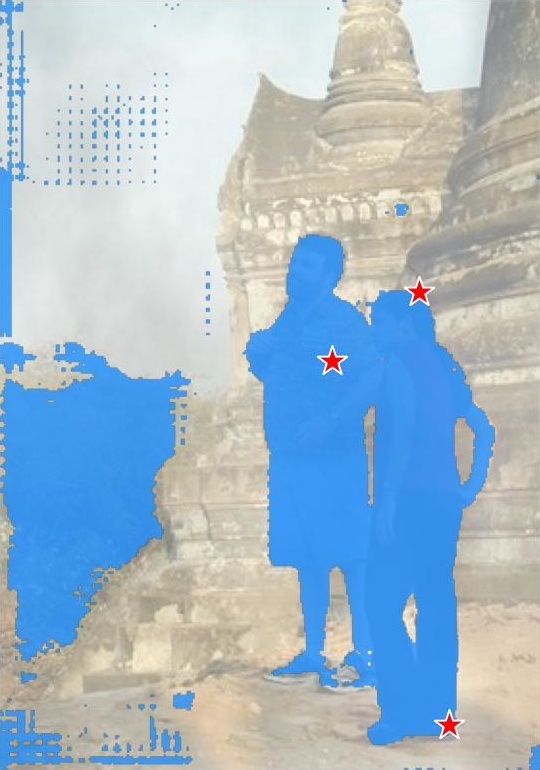} & 
    \includegraphics[width=.16\textwidth, height=2.7cm]{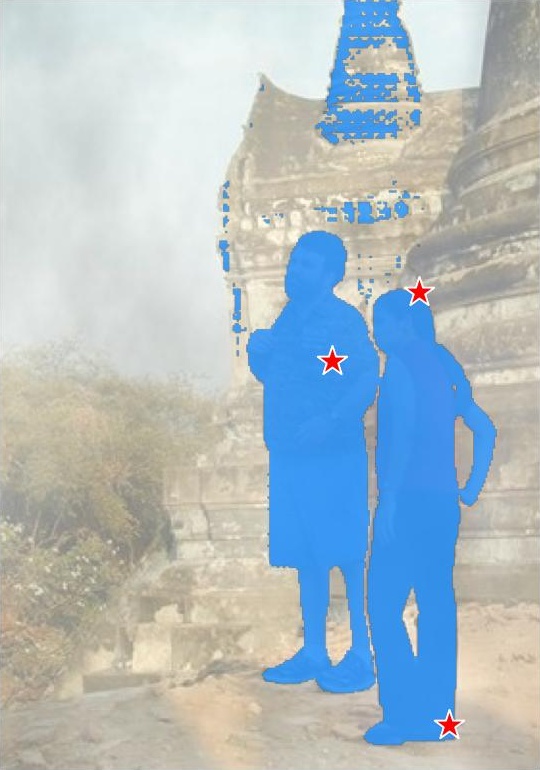} &
    \includegraphics[width=.16\textwidth, height=2.7cm]{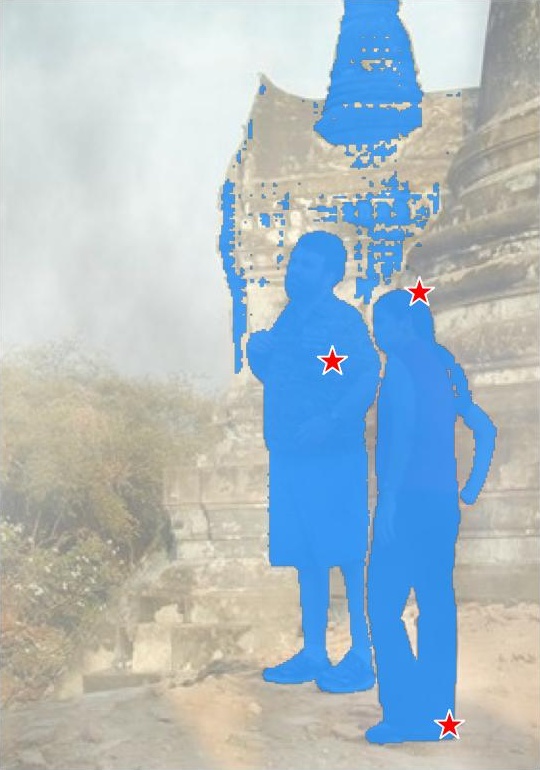} & 
    \includegraphics[width=.16\textwidth, height=2.7cm]{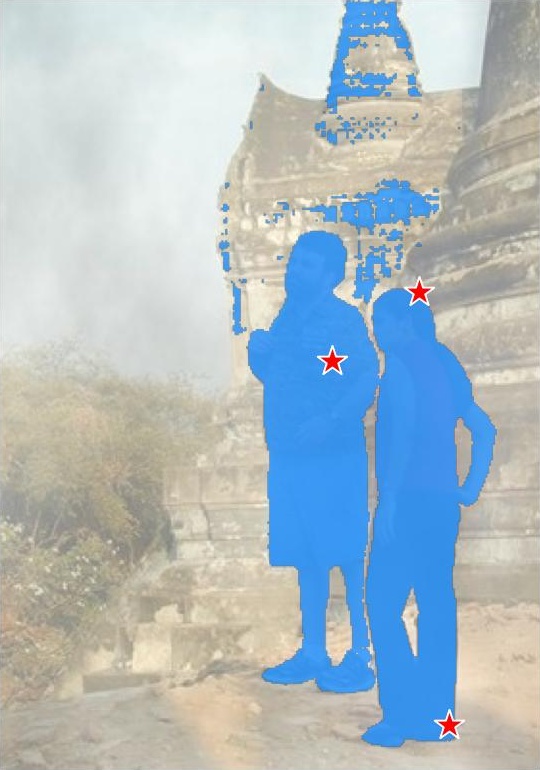} & 
    \includegraphics[width=.16\textwidth, height=2.7cm]{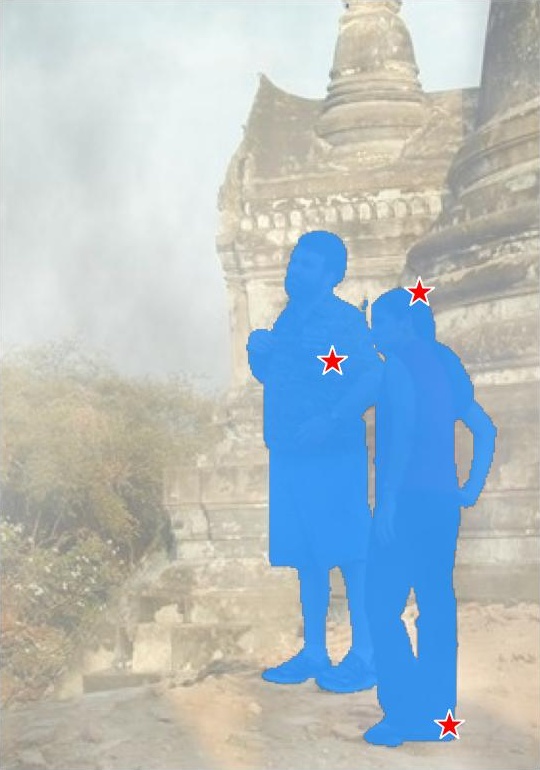} \\
        \multicolumn{6}{c}{\vspace{-14pt}} \\
    \includegraphics[width=.16\textwidth, height=2.cm]{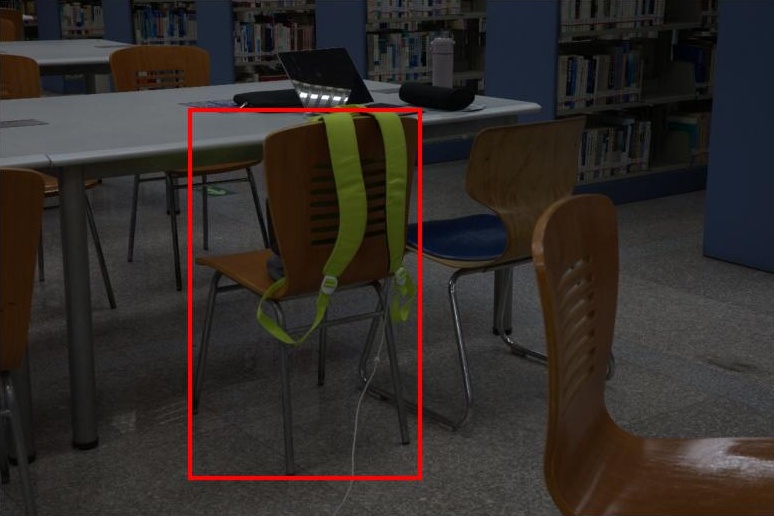} &
    \includegraphics[width=.16\textwidth, height=2.cm]{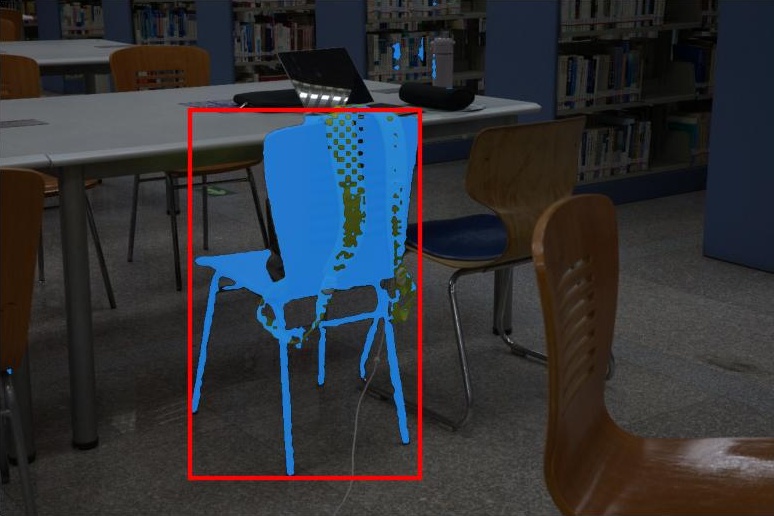} &
    \includegraphics[width=.16\textwidth, height=2.cm]{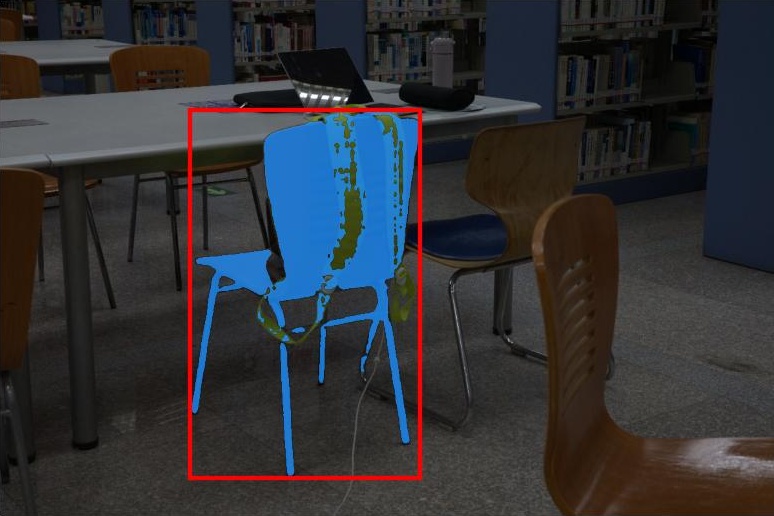} & 
    \includegraphics[width=.16\textwidth, height=2.cm]{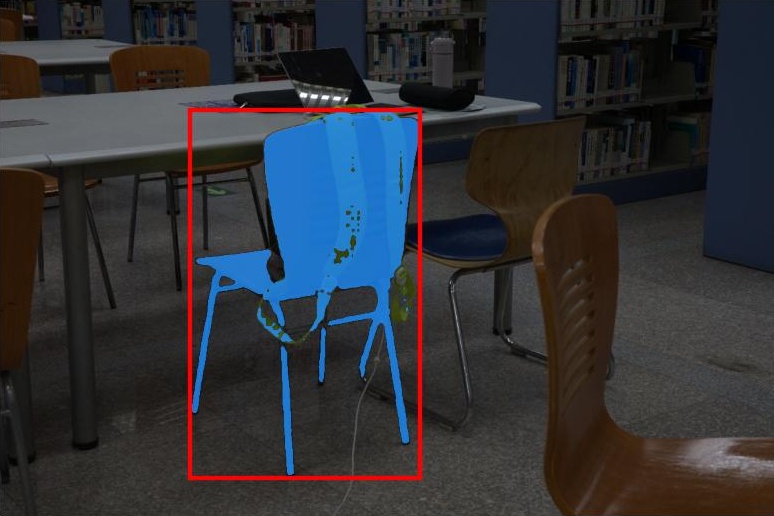} &
    \includegraphics[width=.16\textwidth, height=2.cm]{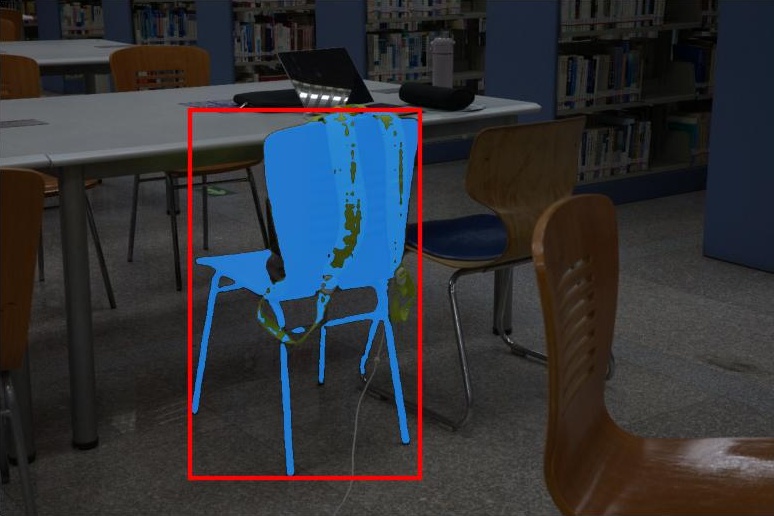} & 
    \includegraphics[width=.16\textwidth, height=2.cm]{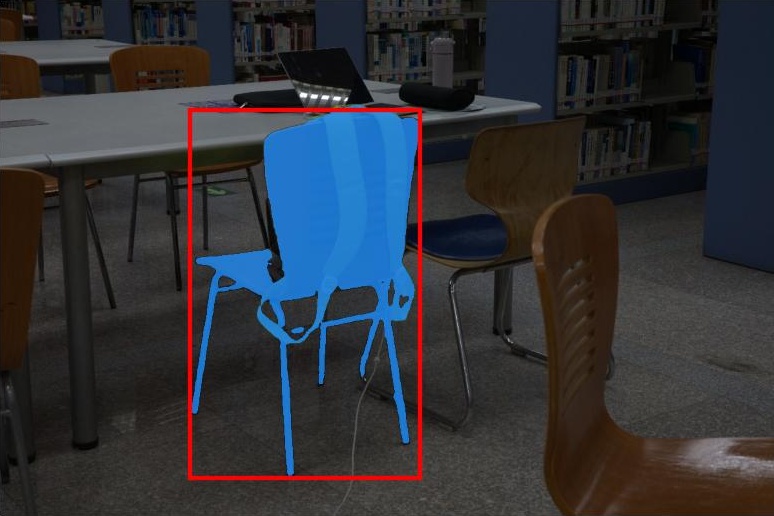}    
    \\    
        \multicolumn{6}{c}{\vspace{-14pt}} \\
    \includegraphics[width=.16\textwidth, height=2.7cm]{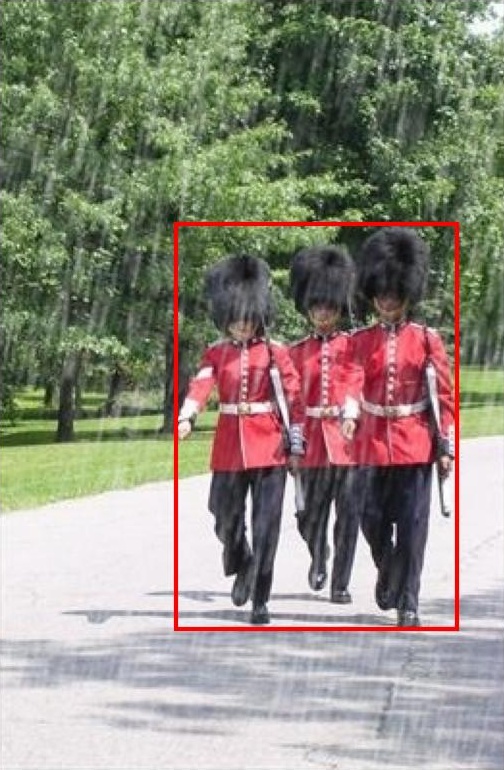} &
    \includegraphics[width=.16\textwidth, height=2.7cm]{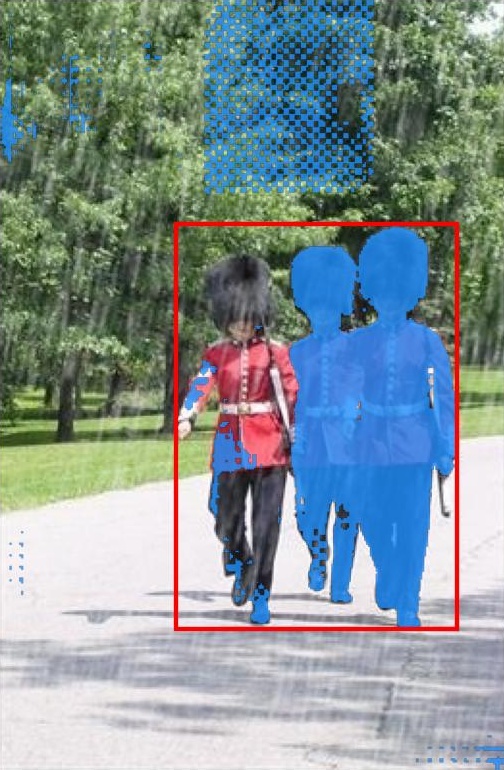} &
    \includegraphics[width=.16\textwidth, height=2.7cm]{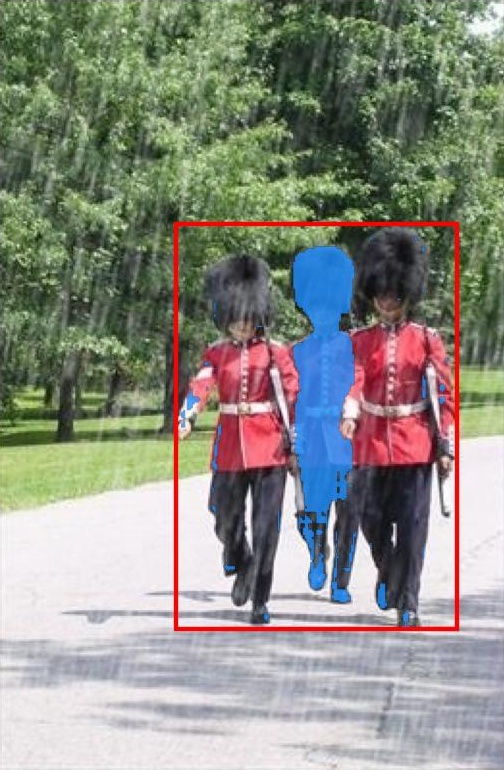} & 
    \includegraphics[width=.16\textwidth, height=2.7cm]{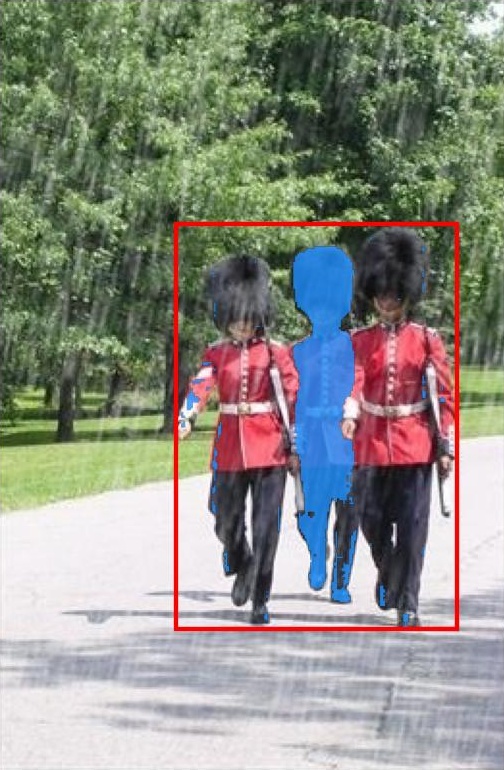} &
    \includegraphics[width=.16\textwidth, height=2.7cm]{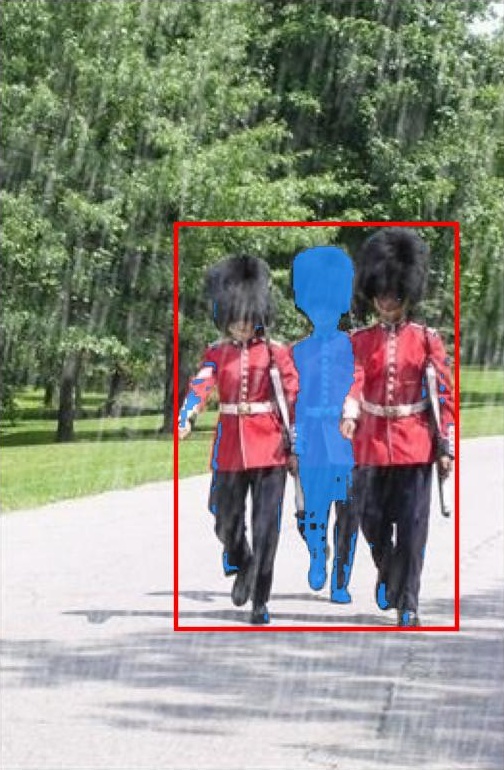} & 
    \includegraphics[width=.16\textwidth, height=2.7cm]{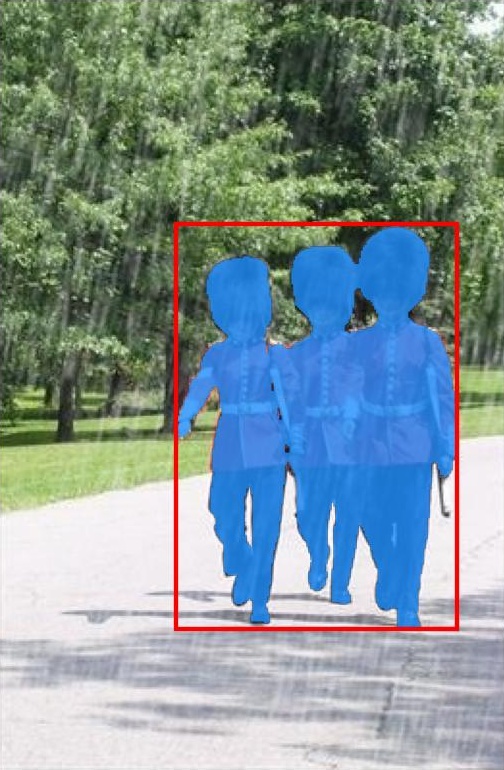}    
    \\

        \multicolumn{6}{c}{\vspace{-14pt}} \\
    \includegraphics[width=.16\textwidth, height=2.cm]{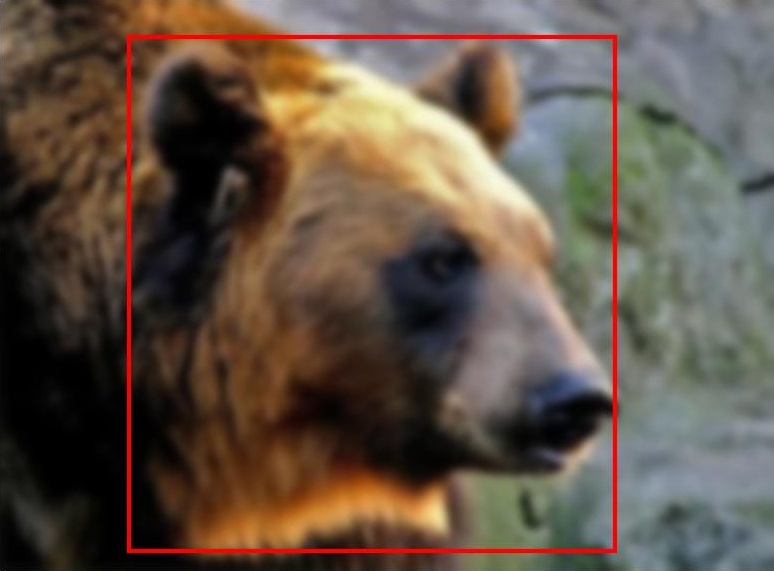} &
    \includegraphics[width=.16\textwidth, height=2.cm]{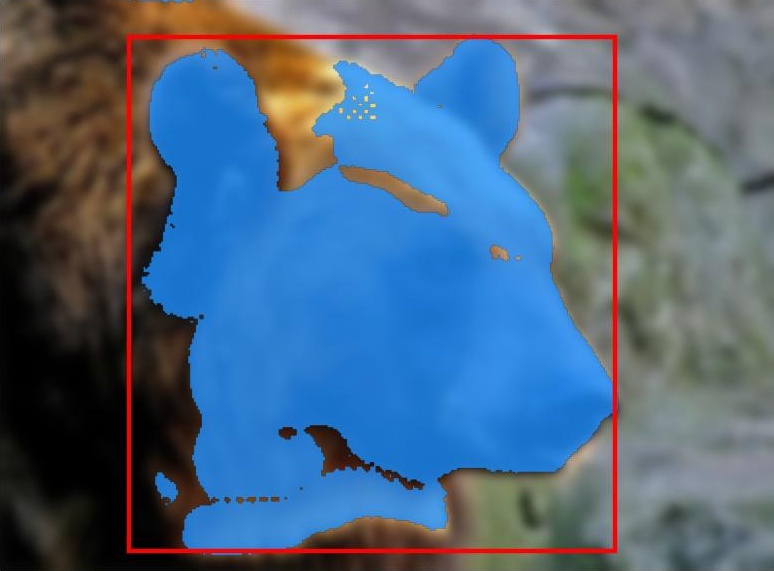} &
    \includegraphics[width=.16\textwidth, height=2.cm]{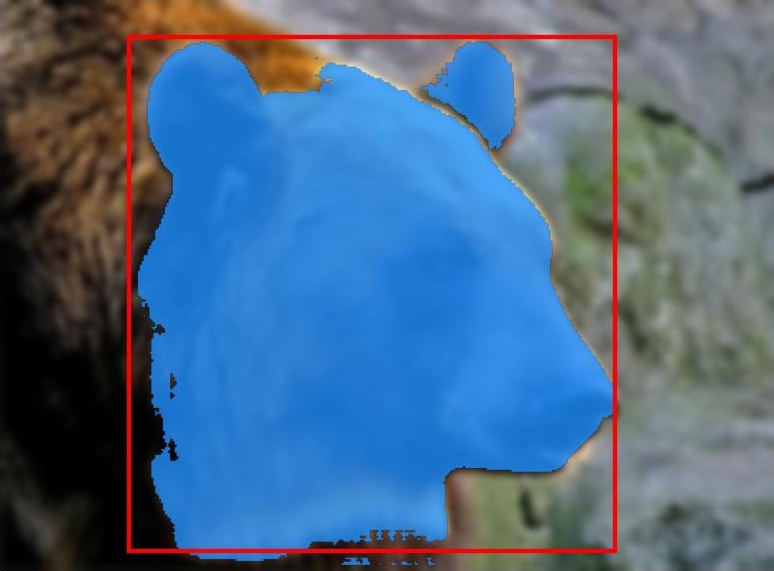} & 
    \includegraphics[width=.16\textwidth, height=2.cm]{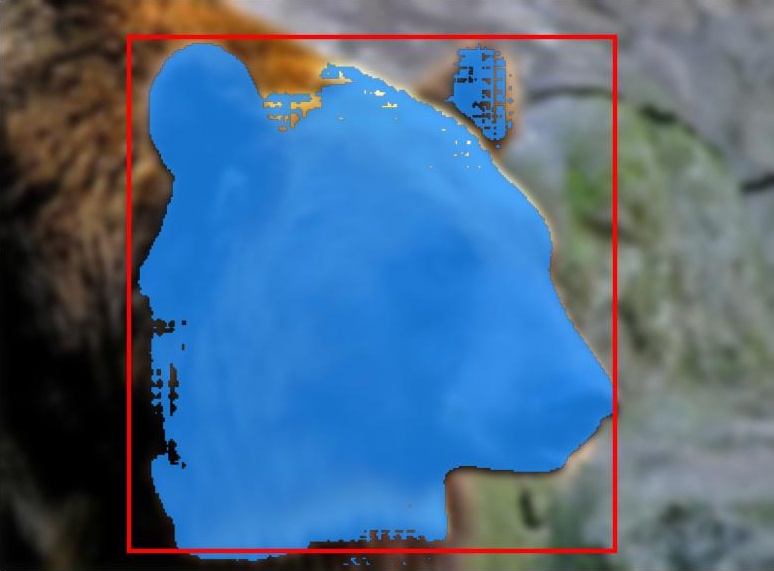} &
    \includegraphics[width=.16\textwidth, height=2.cm]{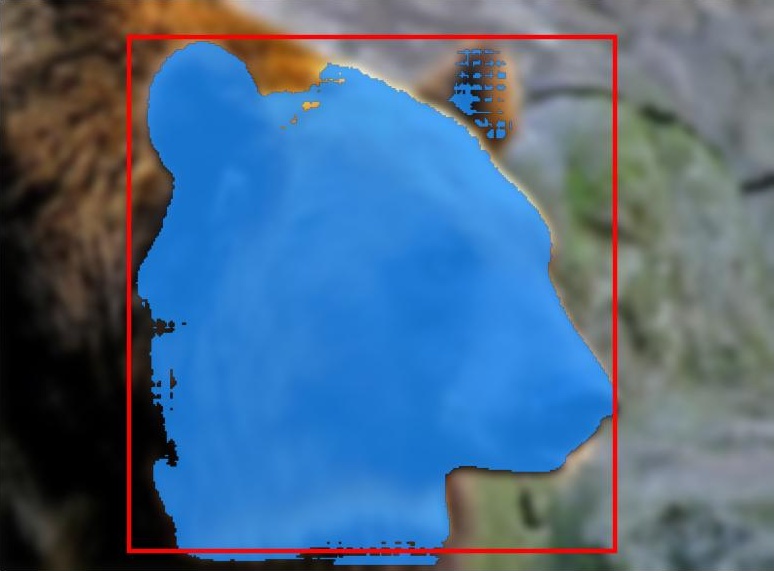} & 
    \includegraphics[width=.16\textwidth, height=2.cm]{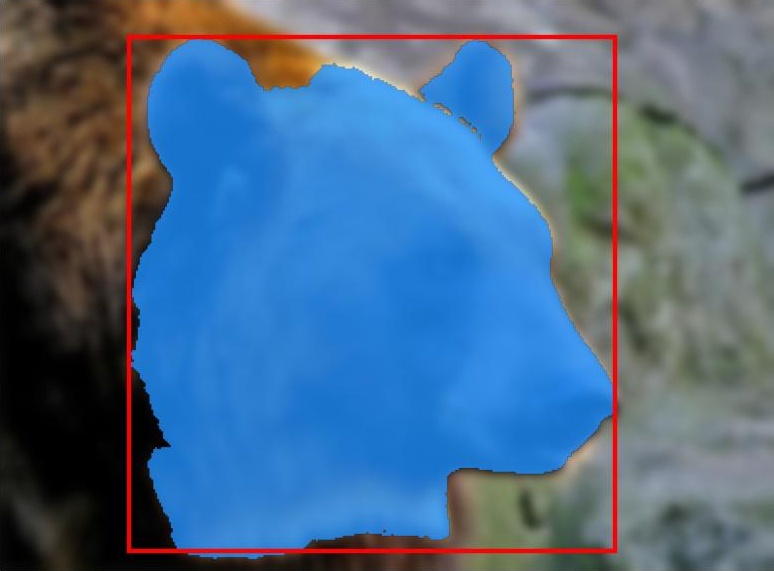} 
\\    
        \multicolumn{6}{c}{\vspace{-14pt}} \\
    \includegraphics[width=.16\textwidth, height=2.cm]{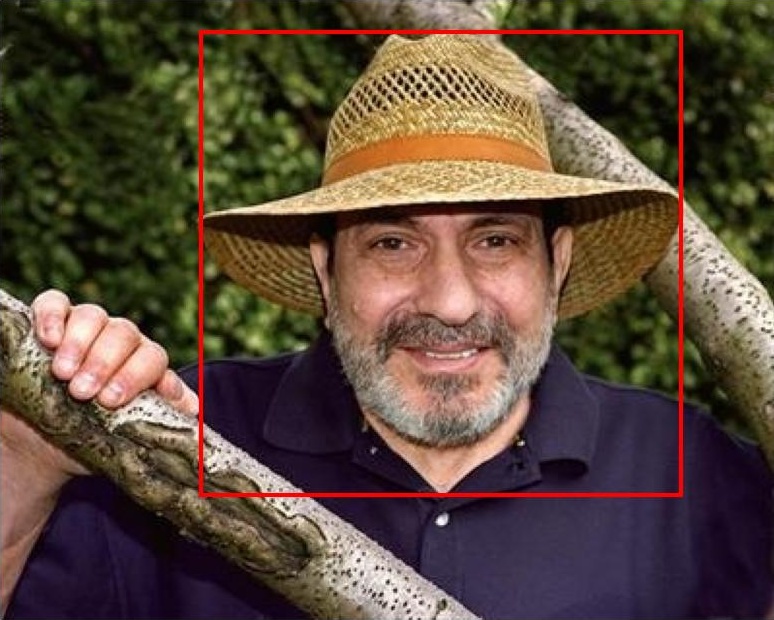} &
    \includegraphics[width=.16\textwidth, height=2.cm]{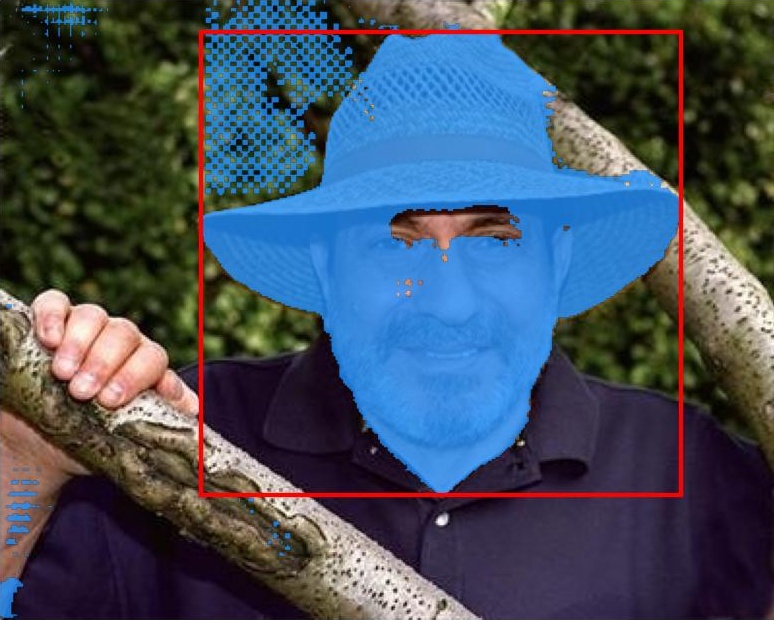} &
    \includegraphics[width=.16\textwidth, height=2.cm]{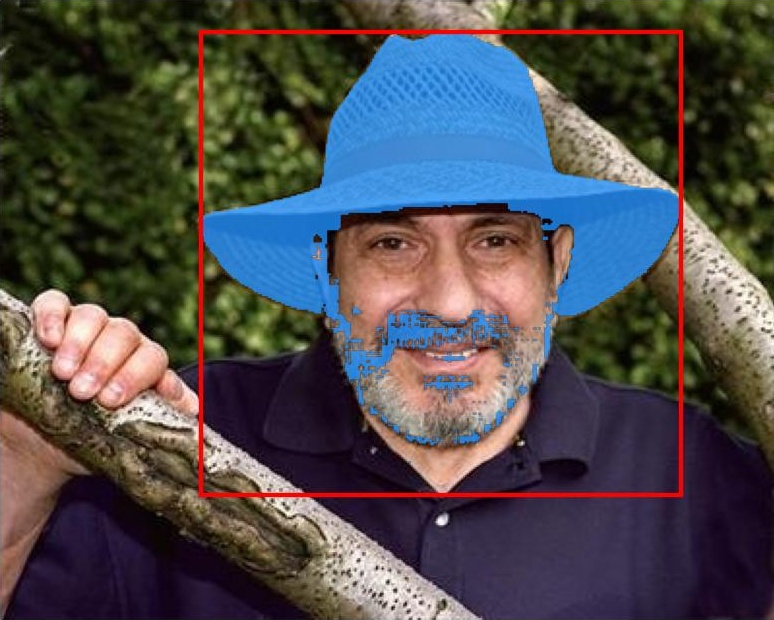} & 
    \includegraphics[width=.16\textwidth, height=2.cm]{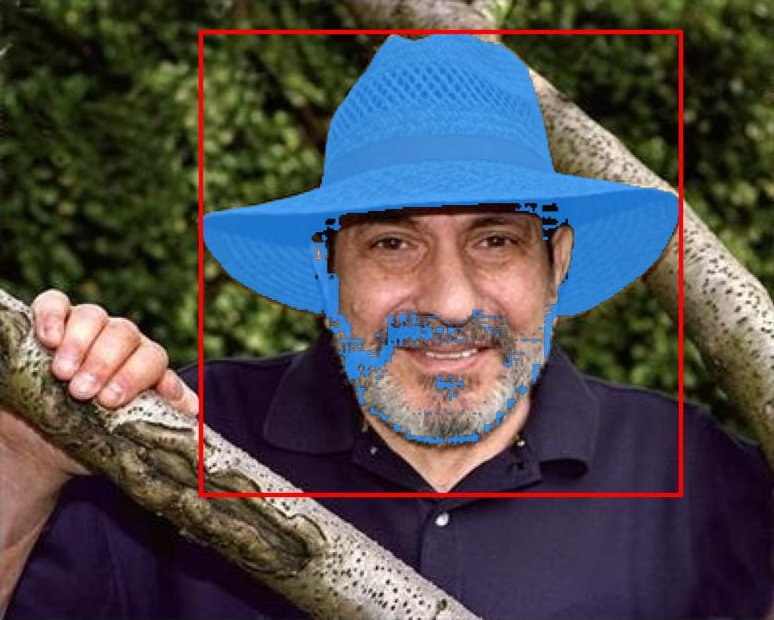} &
    \includegraphics[width=.16\textwidth, height=2.cm]{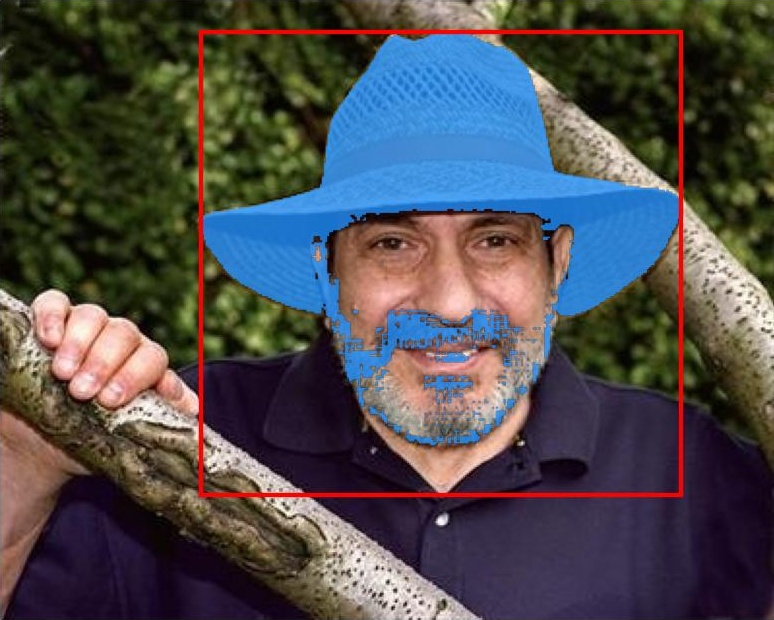} & 
    \includegraphics[width=.16\textwidth, height=2.cm]{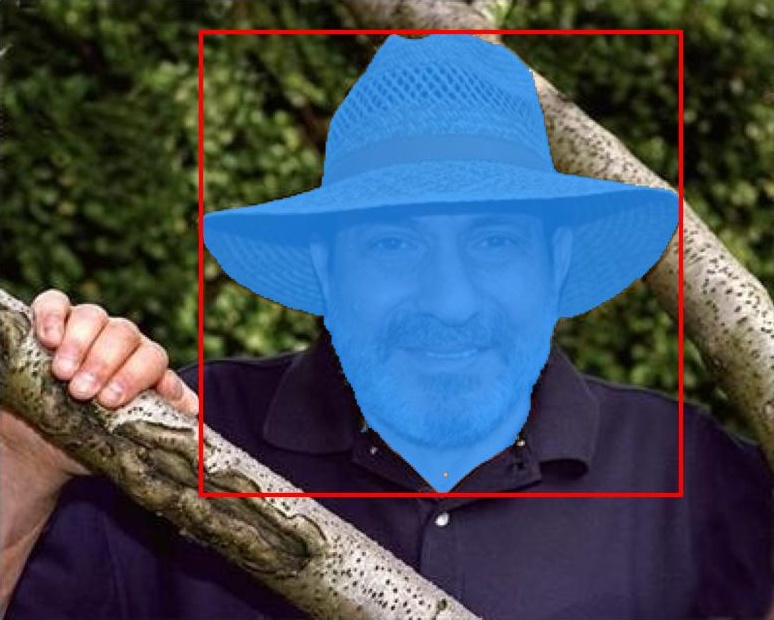}    
    \\   
        \multicolumn{6}{c}{\vspace{-14pt}} \\
    \includegraphics[width=.16\textwidth, height=2.7cm]{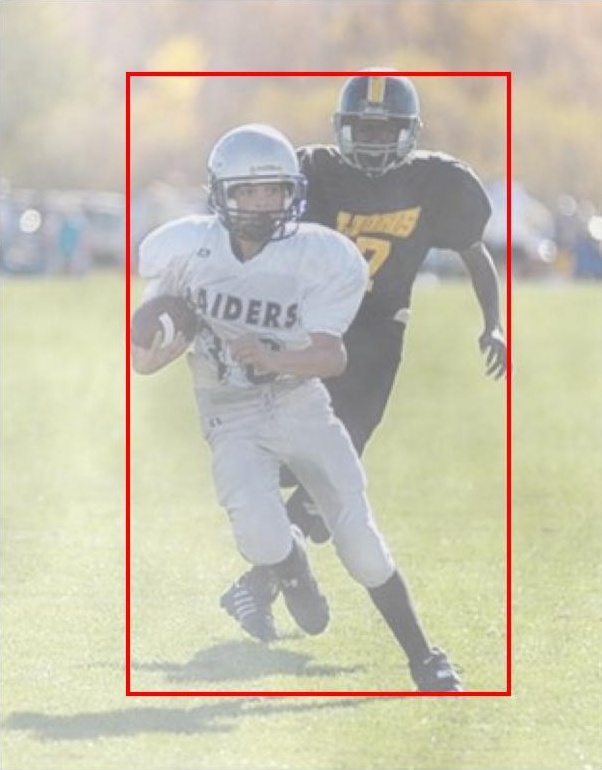} &
    \includegraphics[width=.16\textwidth, height=2.7cm]{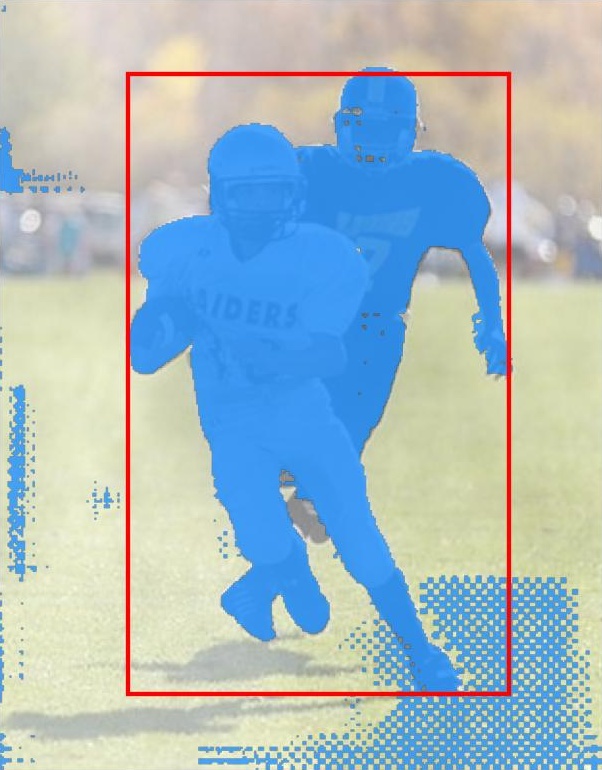} &
    \includegraphics[width=.16\textwidth, height=2.7cm]{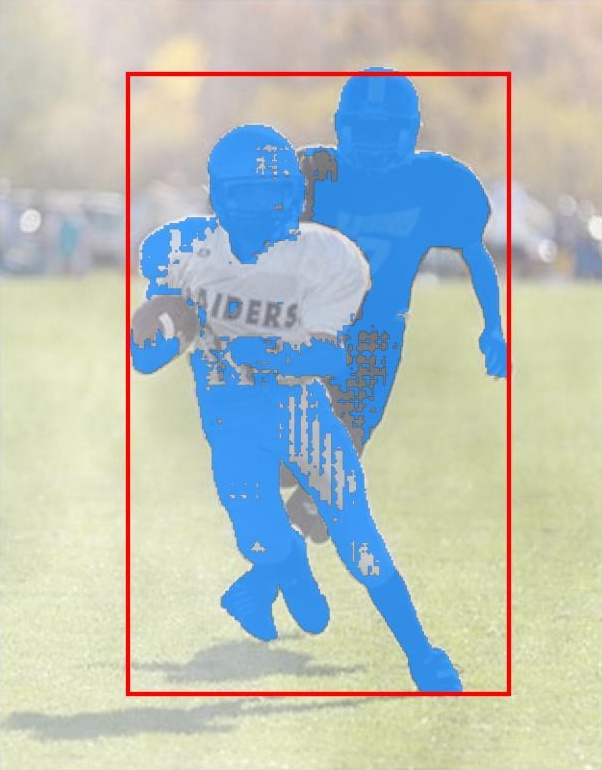} & 
    \includegraphics[width=.16\textwidth, height=2.7cm]{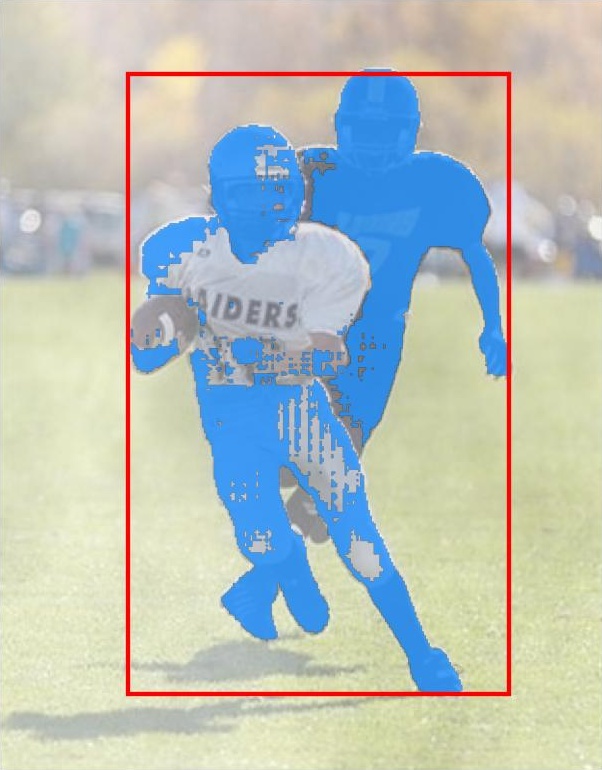} &
    \includegraphics[width=.16\textwidth, height=2.7cm]{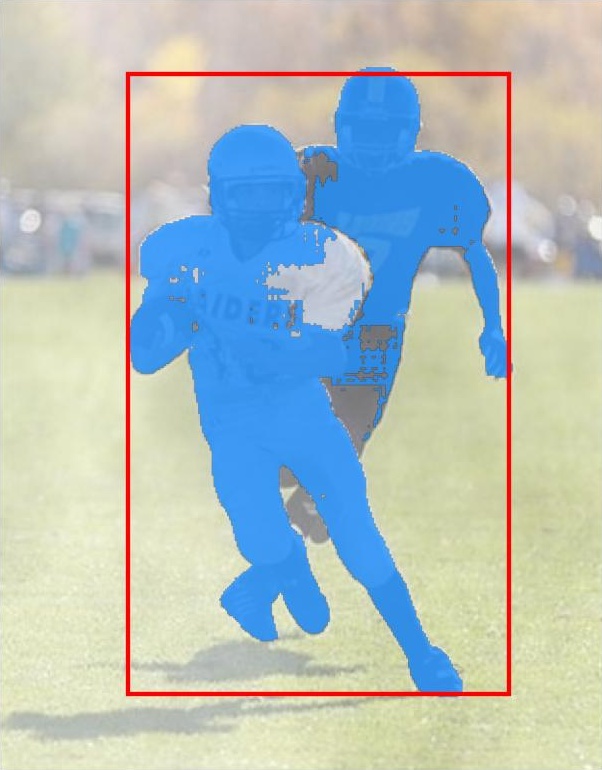} & 
    \includegraphics[width=.16\textwidth, height=2.7cm]{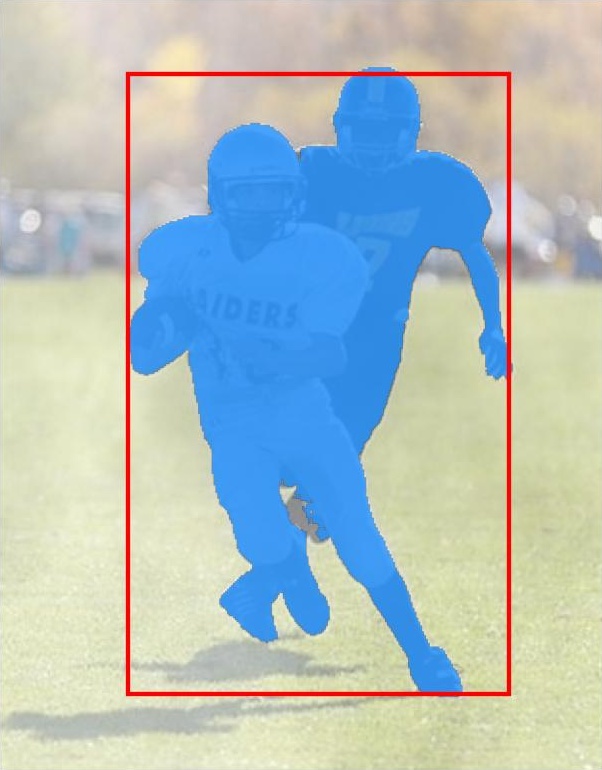}    
    \\    

  \end{tabular}
\caption{\textbf{Qualitative Analysis of Segmentation:} This figure offers a visual comparison to illustrate the enhanced performance of RobustSAM compared to current methods. Notably, Rows 3 and 8 depict scenes without degradations..}
  \label{fig:qualitative}
\end{figure*}

\begin{figure}[ht!]
  \centering
  \setlength{\tabcolsep}{2pt} %
  \begin{tabular}{cc}
    Input & Ground Truth \\
    \includegraphics[width=0.47\linewidth]{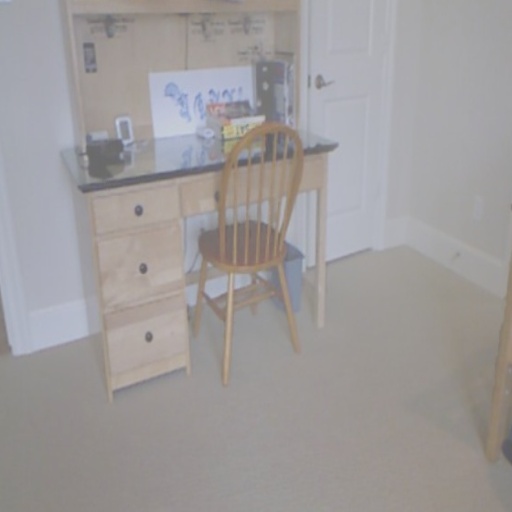} &
    \includegraphics[width=0.47\linewidth]{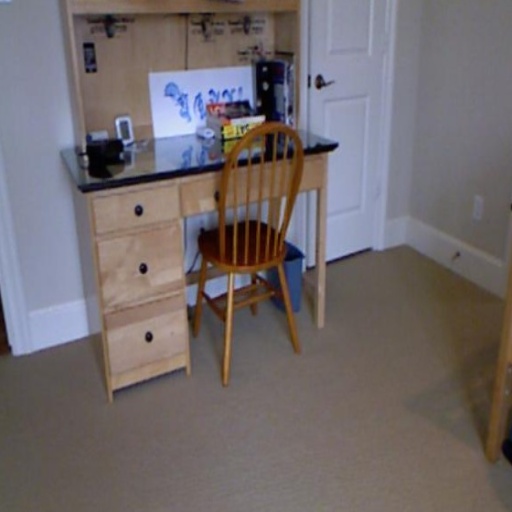} \\
    \multicolumn{2}{c}{\vspace{-8pt}} \\

    Mask-SAM & Dehaze-SAM \\
            \multicolumn{2}{c}{\vspace{-15pt}} \\

        & \scriptsize{(PSNR: 20.838, SSIM: 0.8422)} \\
    \includegraphics[width=0.47\linewidth]{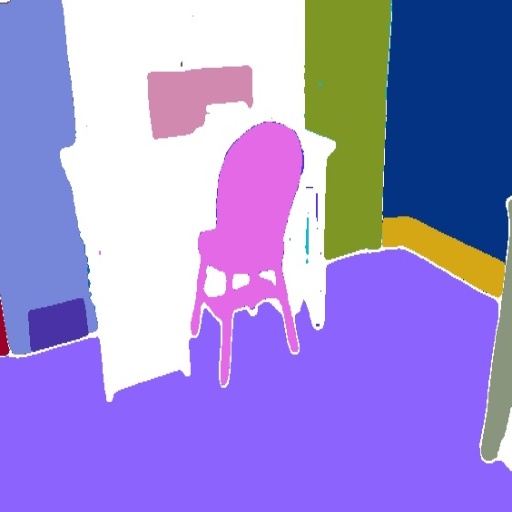} &
    \includegraphics[width=0.47\linewidth]{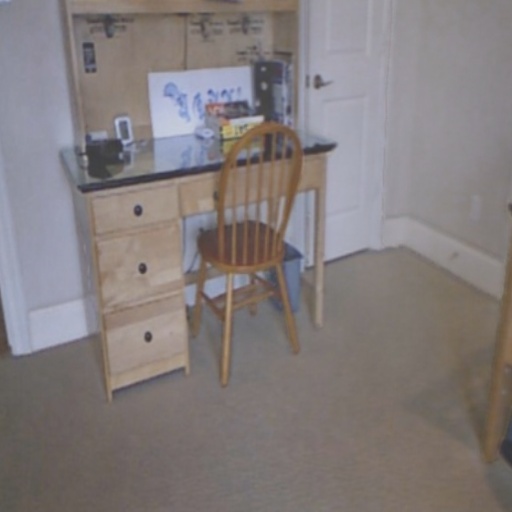} \\
    \multicolumn{2}{c}{\vspace{-8pt}} \\

    Mask-RobustSAM & Dehaze-RobustSAM \\
        \multicolumn{2}{c}{\vspace{-15pt}} \\

        & \scriptsize{(PSNR: 22.029, SSIM: 0.8912)} \\
    \includegraphics[width=0.47\linewidth]{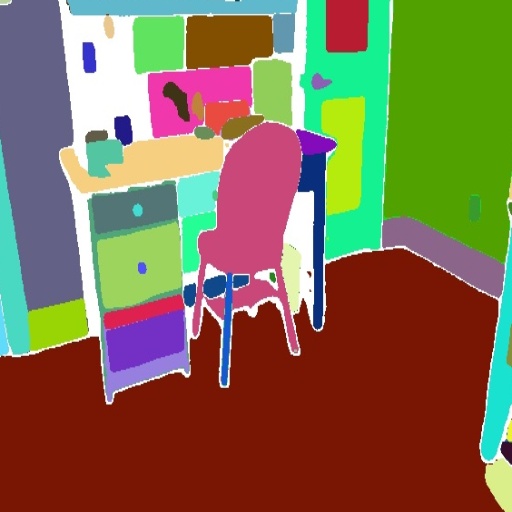} &
    \includegraphics[width=0.47\linewidth]{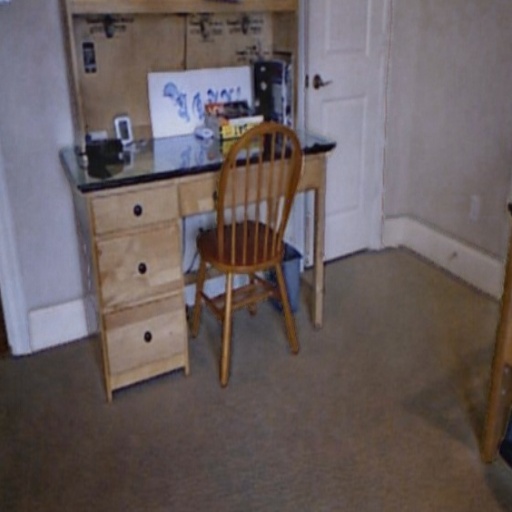} \\
  \end{tabular}
\caption{\textbf{Enhancing SAM-based Dehazing Method:} A qualitative demonstration of RobustSAM's superiority in refining the SAM-based single image dehazing.}
  \label{fig:robust-seg-dehaze}
\end{figure}

\subsection{Qualitative Evaluation}
In our qualitative evaluation, showcased in~\figref{fig:qualitative}, we present a comprehensive set of results illustrating the efficacy of our approach in both degraded and clear (Row 3 and 8) scenarios. We adopt original SAM~\cite{kirillov2023segany}, HQ-SAM~\cite{sam_hq}, Air-Net~\cite{li2022all} + SAM, and URIE~\cite{son2020urie} + SAM in this comparison. These visual comparisons clearly demonstrate the superior segmentation capabilities of our method under various conditions. Notably, our approach maintains high accuracy and detail in degraded scenes, where existing methods often struggle, effectively segmenting intricate patterns and structures. Moreover, our method can maintain the performance in clear conditions.

\subsection{Quantitative Evaluation}
In our quantitative evaluation, we expanded the baseline comparisons, detailed in Tables \ref{tab:eval_real}, \ref{tab:eval_rest}, and \ref{tab:eval_coco}. Our focus was on enhancing segmentation accuracy by preprocessing images with restoration methods like MW-Net~\cite{Patil_2023_ICCV}, SwinIR~\cite{liang2021swinir}, and MPR-Net~\cite{zamir2021multi} before applying the SAM technique. This approach was rigorously tested on the BDD-100k~\cite{yu2018bdd100k} and LIS~\cite{Hong2021Crafting,2023lis} datasets, as well as on subsets of unseen datasets with synthetic degradations, namely COCO~\cite{lin2014microsoft}, NDD20~\cite{trotter2020ndd20}, STREETS~\cite{snyder2019streets}, and FSS-1000~\cite{FSS1000}, all part of the Robust-Seg dataset collection.

Furthermore, we extended our experiments to include fine-tuning of SwinIR, MW-Net, and Air-Net using our degraded-clear image pairs, followed by SAM application (referred to as SwinIR-F, MW-Net-F, and Air-Net-F). We also fine-tuned HQ-SAM with our training data, denoted as HQ-SAM-F. The findings indicate a marginal performance improvement through fine-tuning, but these adaptations still do not match the effectiveness of our proposed method. This underscores the robustness and superiority of our approach, particularly in achieving high segmentation accuracy under various degrees of image degradation.

Moreover, we evaluated the performance of RobustSAM on a specific subset of the SA-1B dataset~\cite{kirillov2023segany}, comprising 11,186 images. This evaluation was conducted in comparison with the standard SAM~\cite{kirillov2023segany} method. The outcomes, presented in Table \ref{tab:sa1b}, clearly indicate that RobustSAM outperforms SAM, demonstrating the efficacy of the proposed RobustSAM approach.

\subsection{Improving SAM-prior Tasks}
We validate the effectiveness of RobustSAM in enhancing downstream tasks (\textit{i.e.,} dehazing~\cite{jin2023let} and deblurring~\cite{li2023sam}) that use SAM as a prior. Our results, showcased in~\figref{fig:robust-seg-dehaze} and~\figref{fig:robust-seg-deblur}, include both the segmentation masks and the reconstructed outcomes for dehazing and deblurring tasks. These results clearly demonstrate that employing RobustSAM as a prior significantly boosts the performance of these tasks. This enhancement is particularly evident in degraded scenarios where RobustSAM maintains its strong segmentation capabilities. By reliably segmenting images even in challenging conditions, RobustSAM provides a robust foundation for subsequent image restoration tasks, leading to improved overall outcomes in both clarity and detail.

\begin{table*}[ht!]
\centering
\scalebox{0.8}{
\begin{tabular}{ccccccccc}
\toprule
             & LVIS~\cite{gupta2019lvis} & ThinObject-5k~\cite{liew2021deep} & MSRA10K~\cite{ChengPAMI} & NDD20~\cite{trotter2020ndd20} & STREETS~\cite{snyder2019streets} & FSS-1000~\cite{FSS1000} & COCO~\cite{lin2014microsoft} & \textbf{Total} \\ \midrule
Image Number & 20252 & 4748 & 10000 & 1000 & 1000 & 1000 & 5000  & 43000 \\ \bottomrule
\end{tabular}}
\caption{\textbf{Data composition of our constructed Robust-Seg dataset.}}
\label{tab:robust-seg}
\end{table*}

\begin{figure*}[ht!]
  \centering
  \setlength{\tabcolsep}{2pt} %
  \begin{tabular}{ccccc}
    Snow & Fog & Rain & Gaussian Noise & ISO Noise \\

    \includegraphics[width=0.18\linewidth]{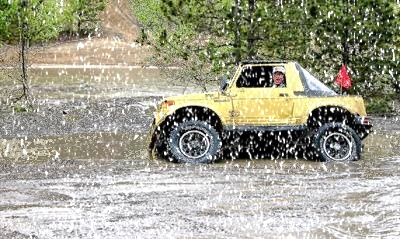} &
    \includegraphics[width=0.18\linewidth]{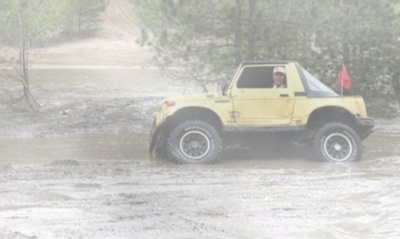} &
    \includegraphics[width=0.18\linewidth]{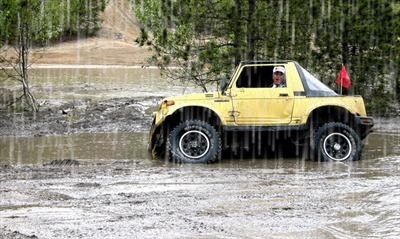} &
    \includegraphics[width=0.18\linewidth]{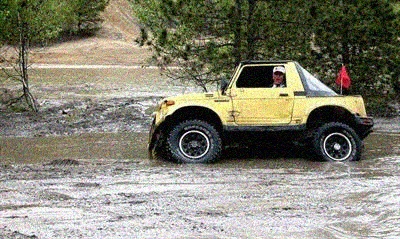} &
    \includegraphics[width=0.18\linewidth]{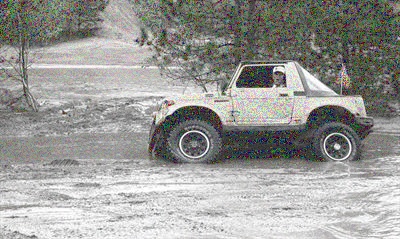} \\

        Impulse Noise & Re-sampling Blur & Motion Blur & Zoom Blur & Color Jitter \\
    \includegraphics[width=0.18\linewidth]{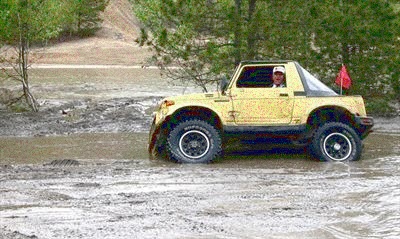} &
    \includegraphics[width=0.18\linewidth]{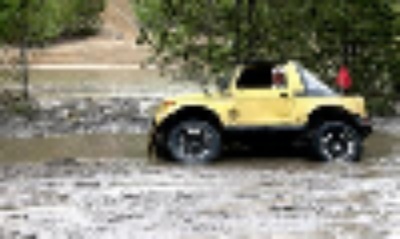} &
    \includegraphics[width=0.18\linewidth]{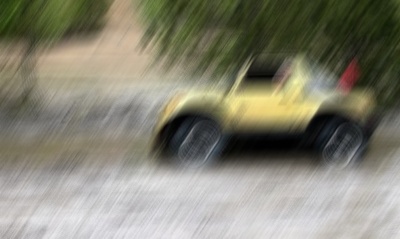} &
    \includegraphics[width=0.18\linewidth]{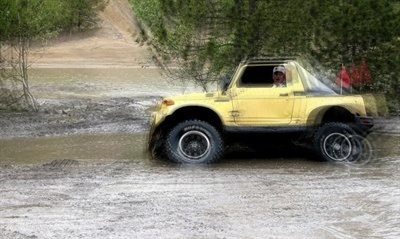} &
    \includegraphics[width=0.18\linewidth]{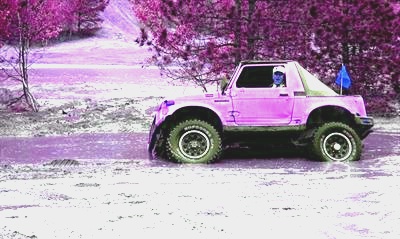} \\

        Compression & Elastic Transform & Frosted Glass Blur & Low Light & Contrast \\
    \includegraphics[width=0.18\linewidth]{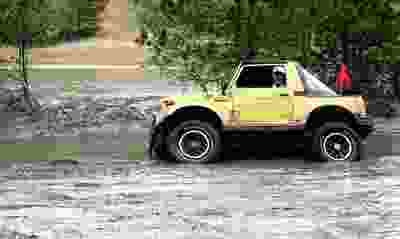} &
    \includegraphics[width=0.18\linewidth]{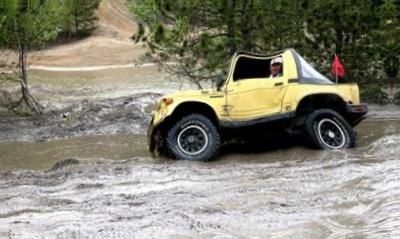} &
    \includegraphics[width=0.18\linewidth]{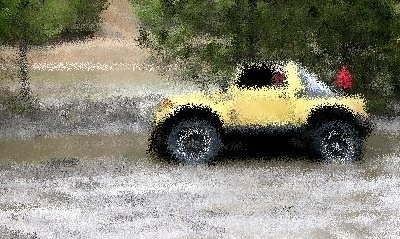} &
    \includegraphics[width=0.18\linewidth]{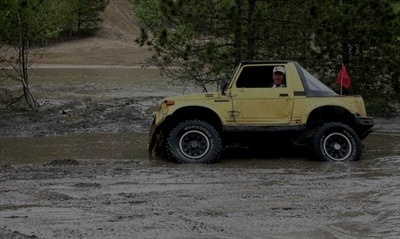} &
    \includegraphics[width=0.18\linewidth]{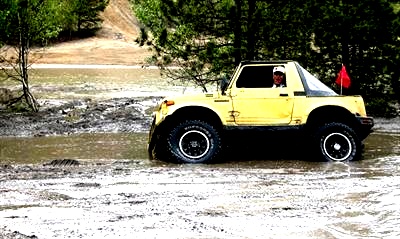} \\

  \end{tabular}
\caption{\textbf{Illustrative samples of images with synthetic degradations from the Robust-Seg dataset.}}
  \label{fig:robust-seg}
\end{figure*}

\section{Robust-Seg Dataset}
The meticulously curated Robust-Seg dataset, designed to train and evaluate the RobustSAM model, encapsulates a rich repository of 43,000 images with corresponding annotated masks. These images are sourced from a suite of renowned datasets, namely LVIS~\cite{gupta2019lvis}, ThinObject-5k~\cite{liew2021deep}, MSRA10K~\cite{ChengPAMI}, NDD20~\cite{trotter2020ndd20}, STREETS~\cite{snyder2019streets}, FSS-1000~\cite{FSS1000}, and COCO~\cite{lin2014microsoft}. We have augmented this collection with 15 types of synthetic alterations using the albumentations~\cite{info11020125} and imgaug~\cite{imgaug} libraries, introducing a diverse range of visual degradations to the dataset. The degradations include snow, fog, rain, Gaussian noise, ISO noise, impulse noise, re-sampling blur, motion blur, zoom blur, color jitter, compression artifacts, elastic transformation, frosted glass blur, low light, and contrast adjustments. Alongside these, one augmentation category is designated for images without any modifications, preserving their original clarity.

These synthetic degradations are meant to simulate a breadth of challenging visual scenarios, thereby extending the robustness of the model against a spectrum of image qualities. This strategic augmentation process yields 688,000 image-mask pairs, significantly expanding the dataset's volume and variety. The specific quantities of images drawn from each contributing dataset are detailed in~\tabref{tab:robust-seg}.

For a visual demonstration of the augmented images and their varied degradations, please refer to~\figref{fig:robust-seg}, where we showcase the synthetic effects introduced to the dataset.

\begin{figure}[ht!]
  \centering
  \setlength{\tabcolsep}{2pt} %
  \begin{tabular}{cc}
    Input & Ground Truth \\
    \includegraphics[width=0.47\linewidth]{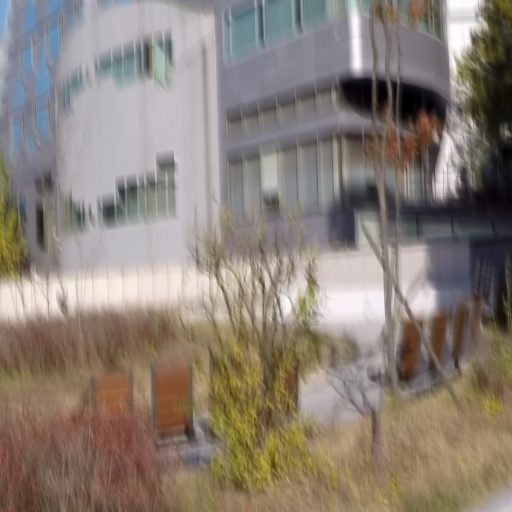} &
    \includegraphics[width=0.47\linewidth]{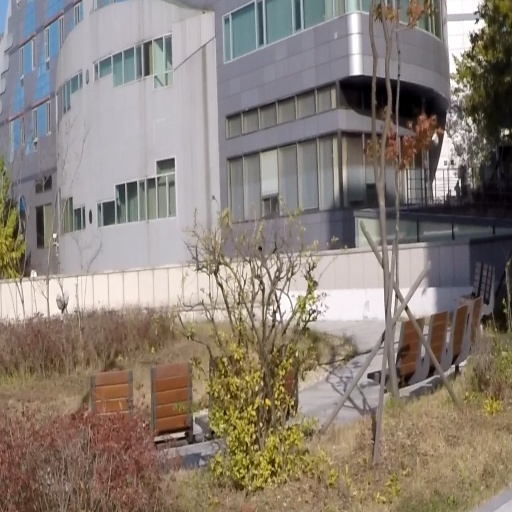} \\
    \multicolumn{2}{c}{\vspace{-8pt}} \\

    Mask-SAM & Deblur-SAM \\
                \multicolumn{2}{c}{\vspace{-15pt}} \\

        & \scriptsize{(PSNR: 26.514, SSIM: 0.7974)} \\

    \includegraphics[width=0.47\linewidth]{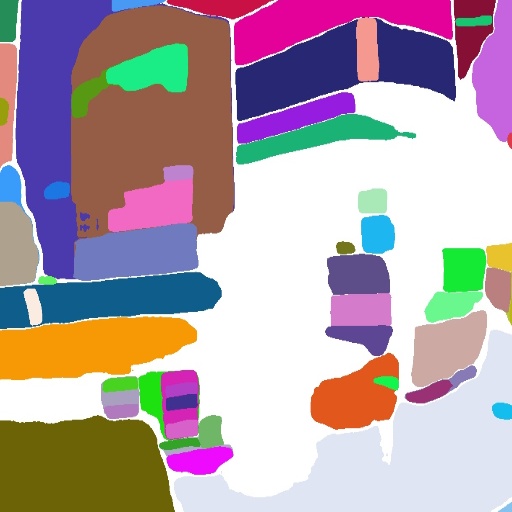} &
    \includegraphics[width=0.47\linewidth]{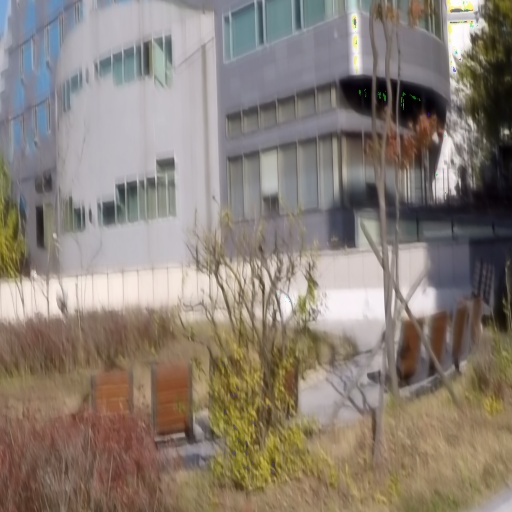} \\
    \multicolumn{2}{c}{\vspace{-8pt}} \\

    Mask-RobustSAM & Deblur-RobustSAM \\
                \multicolumn{2}{c}{\vspace{-15pt}} \\

    & \scriptsize{(PSNR: 27.704, SSIM: 0.8669)} \\

    \includegraphics[width=0.47\linewidth]{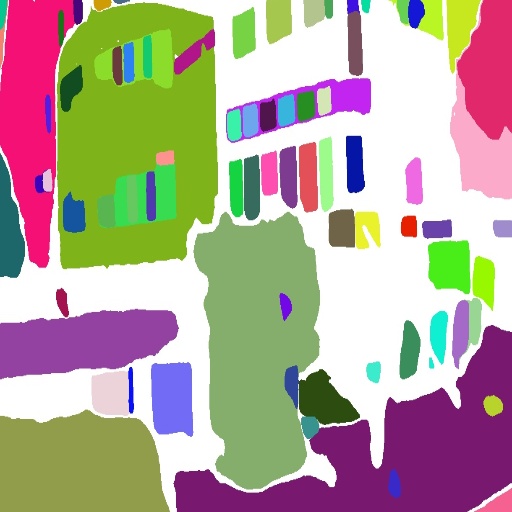} &
    \includegraphics[width=0.47\linewidth]{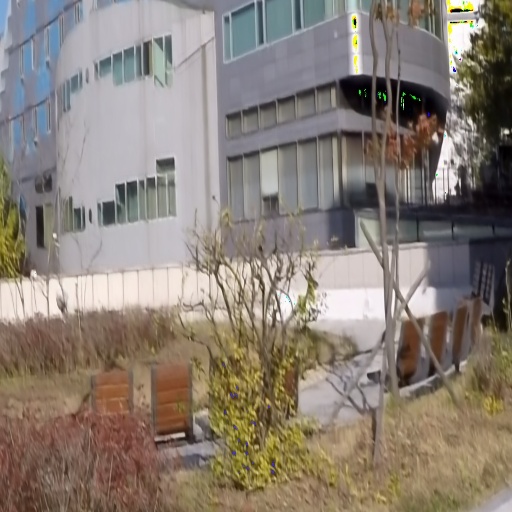} \\
  \end{tabular}
\caption{\textbf{Enhancing SAM-based Deblurring Method:} A qualitative demonstration of RobustSAM's superiority in refining the SAM-based single image deblurring.}
  \label{fig:robust-seg-deblur}
\end{figure}
\section{Implementation Details}
\subsection{Network Architecture}
During training RobustSAM on the composed Robust-Seg dataset, we fix the model parameters of the pre-trained SAM model (gray blocks in Fig. 2 of the main paper) while only making the proposed RobustSAM learnable, including Robust Output Token (ROT), Anti-Degradation Output Token Generation (AOTG) module, Anti-Degradation Mask Feature Generation (AMFG) module and a three-layer MLP which is used to generate the final robust mask.
\noindent \smallskip\\
\textbf{AMFG module.} 
The AMFG module first passes the input feature through the Instance Normalization (IN) and Batch Normalization (BN) layers, respectively. The ReLU activation function is applied to the normalized features. Then, we convolve both output features using a 3×3 convolution layer with padding one and sum them up together. Next, a selector network will be used to generate two attention maps (which have the same shape) based on the summed feature. 

There are several steps in the selector network to generate attention maps. First of all, we utilize an adaptive average pooling layer to reshape the input feature. The reshaped feature is then processed using a fully connected layer, followed by the ReLU activation function. Next, two fully connected layers adjust the dimensions of the input features and generate two unnormalized attention maps. Both attention maps are stacked, and the Softmax function is applied to normalize the attention maps between 0 and 1. Lastly, the normalized attention maps will perform element-wise multiplication with the input features of the selector network.

Following the aforementioned process, we can obtain an enhanced feature. To compensate for any semantic information that may have been lost, this enhanced feature is concatenated with the original input features along the channel dimension. Then, we choose the squeeze-and-excitation~\cite{hu2018squeeze} approach to refine the concatenated feature. 

The output feature of the squeeze-and-excitation module will then be transformed using the Fourier transform to obtain phase components and amplitude components, respectively. After that, we apply a 1×1 convolution with zero padding and stride one on the amplitude components to remove degradation elements. Next, an inverse Fourier transform is performed to restore the refined features to their original spatial representation. 

Finally, a combination of two transposed convolution layers with 2×2 kernels and stride two is applied to align the dimension and generate the final output feature of the AMFG module. 
\noindent \smallskip\\
\textbf{AOTG module.} On the other hand, the AOTG module consists of two IN layers and an MLP network. The original robust output token will first pass through the IN layers to filter out information sensitive to degradation-related details. After that, an MLP is applied to adjust the dimension of the robust output token. There are two fully connected layers inside the MLP, with a ReLU activation function between them.

\subsection{Prompt Generation}
We follow the same inference pipeline of SAM but use the mask prediction from robust output token as the final mask prediction. For box-prompting-based evaluation, we utilize the ground truth mask to generate four corner coordinates of the bounding box. Then, the coordinates are used as the box prompt to feed into our RobustSAM model. For point-prompting-based evaluation, we randomly sample the points from the ground truth masks and use them as the input point prompts.

\subsection{Evaluation Protocol}
We employ several metrics to assess our model's performance:
\noindent \smallskip\\
\textbf{Intersection over Union (IoU)} is a common metric used to measure segmentation accuracy on a particular dataset. It is defined as the area of overlap between the predicted and ground truth segmentation divided by the area of union between the predicted and ground truth segmentation. The IoU metric is given by:
\begin{equation}
    \text{IoU} = \frac{|P \cap G|}{|P \cup G|}
\end{equation}
where \(P\) represents the set of pixels in the predicted segmentation, and \(G\) is the set of pixels in the ground truth segmentation. Higher IoU values indicate better segmentation accuracy.
\noindent \smallskip\\
\textbf{Dice Coefficient (Dice)}~\cite{lin2017focal}, often referred to as the Dice Similarity Coefficient (DSC), is a statistical tool that measures the similarity between two sets of data. In the context of image segmentation, it quantifies the similarity between the predicted segmentation and the ground truth. The Dice Coefficient is calculated as follows:
\begin{equation}
    \text{Dice} = \frac{2|P \cap G|}{|P| + |G|}
\end{equation}
where \(|P \cap G|\) represents the common elements (overlapping pixels) between the predicted and ground truth sets, and \(|P|\) and \(|G|\) are the total elements in each set, respectively. Like IoU, a higher Dice score indicates greater similarity between the predicted and actual segmentations.
\noindent \smallskip\\
\textbf{Pixel Accuracy (PA)} quantifies the proportion of correctly classified pixels in an image. It is calculated as follows:
\begin{equation}
    \text{PA} = \frac{\sum_{i=1}^{N} \mathbb{1}(p_i = g_i)}{N}
\end{equation}
where \(N\) is the total number of pixels in the image, \(p_i\) represents the predicted class of pixel \(i\), and \(g_i\) is the ground truth class of pixel \(i\). The function \(\mathbb{1}(p_i = g_i)\) is an indicator function that equals 1 when the predicted class of a pixel matches its ground truth class and 0 otherwise. The numerator sums up all instances where the predicted and actual classes of pixels are the same, and the denominator is the total number of pixels. A higher PA indicates better accuracy of the model in classifying each pixel.
\noindent \smallskip\\
\textbf{Average Precision (AP)} measures the average of precision scores at different thresholds. It is a way to summarize the precision-recall curve into a single value representing overall segmentation accuracy. AP is calculated as follows:
\begin{equation}
  \text{AP} = \frac{1}{N_{\text{thres}}} \sum_{t=1}^{N_{\text{thres}}} \text{Precision}(t)
\end{equation}
where \(N_{\text{thres}}\) is the total number of threshold levels used, and \(\text{Precision}(t)\) is the precision of the segmentation at a specific threshold \(t\). In practice, AP is computed by taking the average of precision values calculated at several predetermined threshold levels. These levels typically range from 0 to 1, indicating the probability threshold at which a pixel is classified as part of the segmented object. A higher AP value indicates a better model performance across various threshold levels.

\clearpage

{\small
\bibliographystyle{ieee_fullname}
\bibliography{egbib}
}

\end{document}